\documentclass[11pt, letterpaper]{article}

\usepackage[left=1in, right=1in, top=1in, bottom=1in]{geometry}
\usepackage{lineno}

\usepackage[utf8]{inputenc}
\usepackage[T1]{fontenc}
\usepackage{bm}
\usepackage{type1cm}
\usepackage{lettrine}
\usepackage{amsmath,amssymb,amsthm}
\usepackage{moreverb}
\usepackage{mathtools}
\usepackage{amsmath}
\usepackage{amssymb}
\usepackage{algorithmic}
\usepackage{array}
\usepackage{graphics}
\usepackage{graphicx}
\usepackage{caption}
\usepackage{extarrows}
\usepackage{color}
\usepackage{framed}
\usepackage{wrapfig}
\usepackage{bm}
\usepackage{mathrsfs}
\usepackage{mathabx}
\usepackage{multirow}
\usepackage{longtable}
\usepackage{xr-hyper}
\usepackage{hyperref}
\usepackage{paralist}
\usepackage{indentfirst}
\usepackage{relsize}
\usepackage{extarrows}
\usepackage{lineno,xcolor}
\usepackage{upgreek}
\usepackage{bm}
\usepackage{mwe}
\usepackage{booktabs}
\usepackage{authblk}
\usepackage{lettrine}
\usepackage{type1cm}

\usepackage[figurename=Figure]{caption}
\usepackage[normalem]{ulem}
\usepackage{textcomp}

\let\olddegree\degree  
\let\degree\relax 
\usepackage{gensymb}
\let\degree\olddegree  

\usepackage[para,online,flushleft]{threeparttable}
\usepackage{subcaption}
\usepackage{tikz}
\usepackage{tikzscale}

\usepackage{nameref}

\usepackage[final, commandnameprefix=always]{changes}

\usepackage[font=footnotesize,labelfont=bf]{caption}

\definecolor{myred}{RGB}{214,39,40}
\definecolor{mygray}{RGB}{176,176,176}
\definecolor{myorange}{RGB}{255,127,14}
\definecolor{mygreen}{RGB}{44,160,44}
\definecolor{mylightgray}{RGB}{204,204,204}
\definecolor{mypurple}{RGB}{148,103,189}
\definecolor{mybrown}{RGB}{140,86,75}
\definecolor{steelblue}{RGB}{31,119,180}

\DeclareRobustCommand\blueline{\raisebox{2pt}{\tikz{\draw[steelblue,line width = 0.8pt] (0,0)--(0.5,0);}}}
\DeclareRobustCommand\orangeline{\raisebox{2pt}{\tikz{\draw[myorange,line width = 0.8pt] (0,0)--(0.5,0);}}}
\DeclareRobustCommand\greenline{\raisebox{2pt}{\tikz{\draw[mygreen,line width = 0.8pt] (0,0)--(0.5,0);}}}
\DeclareRobustCommand\redline{\raisebox{2pt}{\tikz{\draw[myred,line width = 0.8pt] (0,0)--(0.5,0);}}}
\DeclareRobustCommand\purpleline{\raisebox{2pt}{\tikz{\draw[mypurple,line width = 0.8pt] (0,0)--(0.5,0);}}}

\hypersetup{
bookmarksopen=true,
bookmarksnumbered=true,
unicode=false,
pdftoolbar=true,
pdfmenubar=true,
pdffitwindow=false,
pdfstartview={FitH},
pdftitle={My title},
pdfauthor={Author},
pdfsubject={Subject},
pdfcreator={Creator},
pdfproducer={Producer},
pdfkeywords={keywords},
pdfnewwindow=true,
colorlinks=true,
linkcolor=blue,
citecolor=blue,
filecolor=blue,
urlcolor=blue
}

\makeatletter
\newcommand*{\addFileDependency}[1]{
  \typeout{(#1)}
  \@addtofilelist{#1}
  \IfFileExists{#1}{}{\typeout{No file #1.}}
}
\makeatother

\newcommand{\beginsupplement}{%
        \setcounter{table}{0}
        \renewcommand{\thetable}{S\arabic{table}}%
        \setcounter{figure}{0}
        \renewcommand{\thefigure}{S\arabic{figure}}%
}
\begin{document}

\title{\textbf{Multi-resolution partial differential equations preserved learning framework for spatiotemporal dynamics}}

\author[1]{Xin-Yang Liu}
\author[2]{Min Zhu}
\author[2]{Lu Lu}
\author[3,4]{Hao Sun}
\author[1,5,6,*]{Jian-Xun Wang}

\affil[1]{\small Department of Aerospace and Mechanical Engineering, University of Notre Dame, Notre Dame, IN, USA}
\affil[2]{\small Department of Statistics and Data Science, Yale University, New Haven, CT, USA}
\affil[3]{\small Gaoling School of Artificial Intelligence, Renmin University of China, Beijing, China}
\affil[4]{\small Beijing Key Laboratory of Big Data Management and Analysis Methods, Beijing, China}
\affil[5]{Lucy Family Institute for Data \& Society, University of Notre Dame, Notre Dame, IN, USA} 
\affil[6]{ Center for Sustainable Energy (ND Energy), University of Notre Dame, Notre Dame, IN, USA \vspace{12pt}}

\affil[*]{Corresponding author. E-mail: jwang33@nd.edu}

\date{}

\maketitle

\normalsize

\vspace{-28pt} 
\begin{abstract}
\small


    Traditional data-driven deep learning models often struggle with high training costs, error accumulation, and poor generalizability in complex physical processes. Physics-informed deep learning (PiDL) addresses these challenges by incorporating physical principles into the model. Most PiDL approaches regularize training by embedding governing equations into the loss function, yet this depends heavily on extensive hyperparameter tuning to weigh each loss term. To this end, we propose to leverage physics prior knowledge by ``baking'' the discretized governing equations into the neural network architecture via the connection between the partial differential equations (PDE) operators and network structures, resulting in a PDE-preserved neural network (PPNN). This method, embedding discretized PDEs through convolutional residual networks in a multi-resolution setting, largely improves the generalizability and long-term prediction accuracy, outperforming conventional black-box models. The effectiveness and merit of the proposed methods have been demonstrated across various spatiotemporal dynamical systems governed by spatiotemporal PDEs, including reaction-diffusion, Burgers', and Navier-Stokes equations. 
\end{abstract}

\vspace{12pt} 
\section*{INTRODUCTION}
Computational modeling and simulation capabilities play an essential role in understanding, predicting, and controlling various physical processes (e.g., turbulence, heat-flow coupling, and fluid-structure interaction), which often exhibit complex spatiotemporal dynamics. These physical phenomena are usually governed by partial differential equations (PDEs) and can be simulated by solving these PDEs numerically based on, e.g., finite difference (FD), finite volume (FV), finite element (FE), or spectral methods. However, predictive modeling of complex spatiotemporal dynamics using traditional numerical methods can be significantly challenging in many practical scenarios: (1) governing equations for complex systems might not be fully known due to a lack of complete understanding of the underlying physics, for which a first-principled numerical solver cannot be built; (2) conventional numerical simulations are usually time-consuming, making it infeasible for many applications that require many repeated model queries, e.g., optimization design, inverse problems, and uncertainty quantification (UQ), attracting increasing attention in scientific discovery and engineering practice.

Recent advances in scientific machine learning (SciML) and ever-growing data availability open up new possibilities to tackle these challenges. In the past few years, various deep neural networks (DNNs) have been designed to learn the spatiotemporal dynamics in latent spaces enabled by proper orthogonal decomposition (POD)~\cite{lui2019construction,san2019artificial,gao2020non,fresca2022pod} or convolutional encoding-decoding operations~\cite{murata2020nonlinear,mohan2020spatio,maulik2021reduced,fukami2021model}. In particular, fast neural simulators based on graph neural networks (GNN) have been proposed and demonstrated to predict spatiotemporal physics on irregular domains with unstructured meshes~\cite{pfaff2020learning,han2022predicting}. Although showing good promise, most of these works are purely data-driven and black-box in nature, which rely on ``big data'' and may have poor generalizability, particularly in out-of-sample regimes in the parameter space. As a more promising strategy, baking physics prior knowledge (e.g., conservation laws, governing equations, and constraints) into deep learning is believed to be very effective to improve its sample efficiency and generalizability~\cite{baker2019workshop}, here referred to as physics-informed deep learning (PiDL). An impressive contribution in this direction is physics-informed neural networks (PINNs)~\cite{raissi2019physics}, where well-posed PDE information is leveraged to enable deep learning in data-sparse regimes. The general idea of PINNs is to learn (or solve) the PDE solutions with DNNs, where the loss functions are formulated as a combination of the data mismatch and residuals of known PDEs, unifying forward and inverse problems within the same DNN optimization framework. The merits of PINNs have been demonstrated over various scientific applications, including fast surrogate/meta modeling~\cite{sun2020surrogate,zhang2020physics,haghighat2021physics}, parameter/field inversion~\cite{sun2020physics,arzani2021uncovering,lu2021physics,zhang2022analyses}, and solving high-dimensional PDEs~\cite{han2018solving,zhang2019quantifying}, to name a few. Due to the scalability challenges of the pointwise fully-connected PINN formulation to learn continuous functions~\cite{yang2019adversarial,kharazmi2021hp,jagtap2020conservative} or operators~\cite{lu2021learning,li2020fourier,wang2021learning,goswami2022physics}, many remedies and improvements in terms of training and convergence have been proposed~\cite{jagtap2020adaptive,wang2022and,wang2022respecting}. In particular, there is a growing trend in developing field-to-field discrete PINNs by leveraging convolution operations and numerical discretizations, which have been demonstrated to be more efficient in spatiotemporal learning~\cite{gao2021phygeonet,ren2022phycrnet}. For example, convolution neural networks (CNN) or graph convolution networks (GCN) were built to approximate the discrete PDE solutions, where the PDE residuals can be formulated in either strong or weak forms by finite-difference~\cite{geneva2020modeling,gao2021super,wandel2021teaching}, finite volume~\cite{ranade2021discretizationnet}, or finite element methods~\cite{yao2020fea,mitusch2021hybrid,wang2021variational,yin2022interfacing,gao2022physics}. Moreover, recurrent network formulation informed by discretized PDEs have been developed for spatiotemporal dynamic control using model-based reinforcement learning~\cite{liu2021physics}. 

In the realm of PINN framework, the term "physics-informed" generally denotes the incorporation of PDE residuals into the loss or likelihood functions to guide or constrain DNN training. Despite this development, the question of how to effectively use physics-inductive bias—i.e., (partially) known governing equations—to inform the learning architecture design remains an intriguing, relatively unexplored area. The primary focus of this paper is to address this issue. Recent studies have revealed the deep-rooted relationship between neural network structures and ordinal/partial differential equations (ODEs/PDEs)~\cite{haber2017stable,lu2018beyond,rousseau2020residual,ruthotto2020deep,chamberlain2021grand,eliasof2021pde}. For example, Lu et al.~\cite{lu2018beyond} bridged deep convolutional network architectures and numerical differential equations. Chen et al.~\cite{chen2018neural} showed that the residual networks (ResNets)~\cite{he2016deep} can be interpreted as the explicit Euler discretization of an ODE, and ODEs can be used to formulate the continuous residual connection with infinite depths, known as the NeuralODE~\cite{gholami2019anode}. Motivated by differential equations, novel deep learning architectures have been recently developed in the computer science community, e.g., new convolutional ResNets guided by parabolic and hyperbolic PDEs~\cite{ruthotto2020deep}, GRAND as a graph network motivated by diffusion equations~\cite{chamberlain2021grand}, and PDE-GCN motivated by hyperbolic PDEs to improve over-smooth issues in deep graph learning~\cite{eliasof2021pde}. However, these studies mainly aimed to develop generic DNN architectures with some desired features by utilizing specific properties of certain PDEs (e.g., diffusion, dispersion, etc.), and the designed neural networks are not necessarily used to learn the physical processes governed by those PDEs. An attempt was made by Shi et al.~\cite{shi2020finite} to learn PDE-governed dynamics by limiting trainable parameters of CNN using finite difference operators. Despite being a novel attempt, the approach is still purely data-driven without effectively utilizing governing PDEs.

Therefore, this work explores PiDL through learning architecture design, inspired by the broader concept of differentiable programming ($\partial$P) - extending DNNs to more general computer programs that can be trained in a similar fashion to deep learning models~\cite{innes2019differentiable}. In general, a $\partial$P model is formulated by marrying DNNs with a fully differentiable physics-based solver, and thus the gradients can be back-propagated through the entire hybrid neural solver based on automatic differentiation (AD) or discrete adjoint methods. Relevant works include universal differential equations (UDE)~\cite{rackauckas2020universal}, NeuralPDE~\cite{sun2020neupde}, and others, where DNNs are formulated within a differentiable PDE solver for physics-based modeling. In particular, this idea has been recently explored in predictive modeling of rigid body dynamics~\cite{hochlehnert2021learning,heiden2021neuralsim}, epidemic dynamics~\cite{hackenberg2021using}, and fluid dynamics~\cite{kochkov2021machine,belbute2020combining,um2020solver}. These studies imply great promise of incorporating physics-induced prior (i.e., PDE) into DNN architectures.

In this paper, we present a creative approach to designing distinctive learning architectures for predicting spatiotemporal dynamics, where the governing PDEs are preserved as convolution operations and residual connections within the network architecture. This is in sharp contrast to prior PiDL work where the physical laws were enforced as soft constraints within the loss functions, supported by an comprehensive comparision between the proposed method and physics-informed variants of multiple state-of-the-art neural operators. Specifically, we develop an auto-regressive neural solver based on a convolutional ResNet framework, where the residual connections are constructed by preserving the PDE operators in governing equations, which are (partially) known a priori, discretized on low-resolution grids. Meanwhile, encoding-decoding convolution operations with trainable filters enable high-resolution state predictions on fine grids. Compared to classic ResNets with black-box residual connections, the proposed PPNN is expected to be superior in terms of both training efficiency and out-of-sample generalizability for, e.g., unseen boundary conditions and parameters, and extrapolating in time. Conceptually, the proposed framework is similar to using neural networks for closure modeling of classic numerical solvers, which has been explored previously. However, several distinct features make our methodology more general that extends substantially beyond prior studies on merging machine learning with numerical solvers~\cite{bar2019learning, san2018neural, beck2019deep}. Our work is not focused on simply coupling a neural network with a numerical solver or training it to learn specific closures. Instead, the proposed framework integrates (partially or wholly known) physical laws, expressed as PDE operators, directly into the neural networks. This leads to a creative neural architecture design, reflecting a unique design strategy that leverages the profound connection between neural network architecture components and ODEs/PDEs. The differentiability brought by representing numerical operators with neural network components makes an end-to-end time sequence training possible, which distincts the proposed method from closure model learning. This strategy offers a fresh perspective on incorporating physical knowledge into neural network design, underscoring that such integration can enhance the model's performance in predicting complex spatiotemporal dynamics. When compared with the other approach of leveraging physics priors into neural network training: the "physics-informed" methods, our proposed PPNN does show significant merit in terms of cost, generalizability and long-term prediction accuracy. The contributions of this work are summarized as follows: (i) a framework for physics-inspired learning architecture design is presented, where the PDE structures are preserved by the convolution filters and residual connection; (ii) multi-resolution information passing through network layers is proposed to improve long-term model rollout predictions over large time steps; (iii) the superiority of the proposed PPNN is demonstrated for PDE operator learning in terms of training complexity, extrapolability, and generalizability in comparison with the baseline black-box models, using a series of comprehensive numerical experiments on spatiotemporal dynamics governed by various parametric unsteady PDEs, including reaction-diffusion equations, Burgers' equations, and unsteady Navier-Stokes equations.

\section*{RESULTS AND DISCUSSION}
\label{sec:result}
\subsection*{Learning spatiotemporal dynamics governed by PDEs}
We consider a multi-dimensional spatiotemporal system of $\bm{u}(\bm{x}, t; \bm{\lambda})$ governed by a set of nonlinear coupled PDEs parameterized by $\bm{\lambda} \in \mathbb{R}^d$, which is a $d-$dimensional parameter vector, while $\bm{x}$ and $t$ are spatial and temporal coordinates, respectively. Our goal is to develop a data-driven neural solver for rapid predictions of spatiotemporal dynamics given different parameters $\bm{\lambda}$. The neural solver is formulated as a next-step DNN model by learning the dynamic transitions from the current step $t$ to the next time step $t+\Delta t$ ($\Delta t$ is the time step). 

\begin{figure}[t!]
    \centering
    \includegraphics[width=0.8\textwidth]{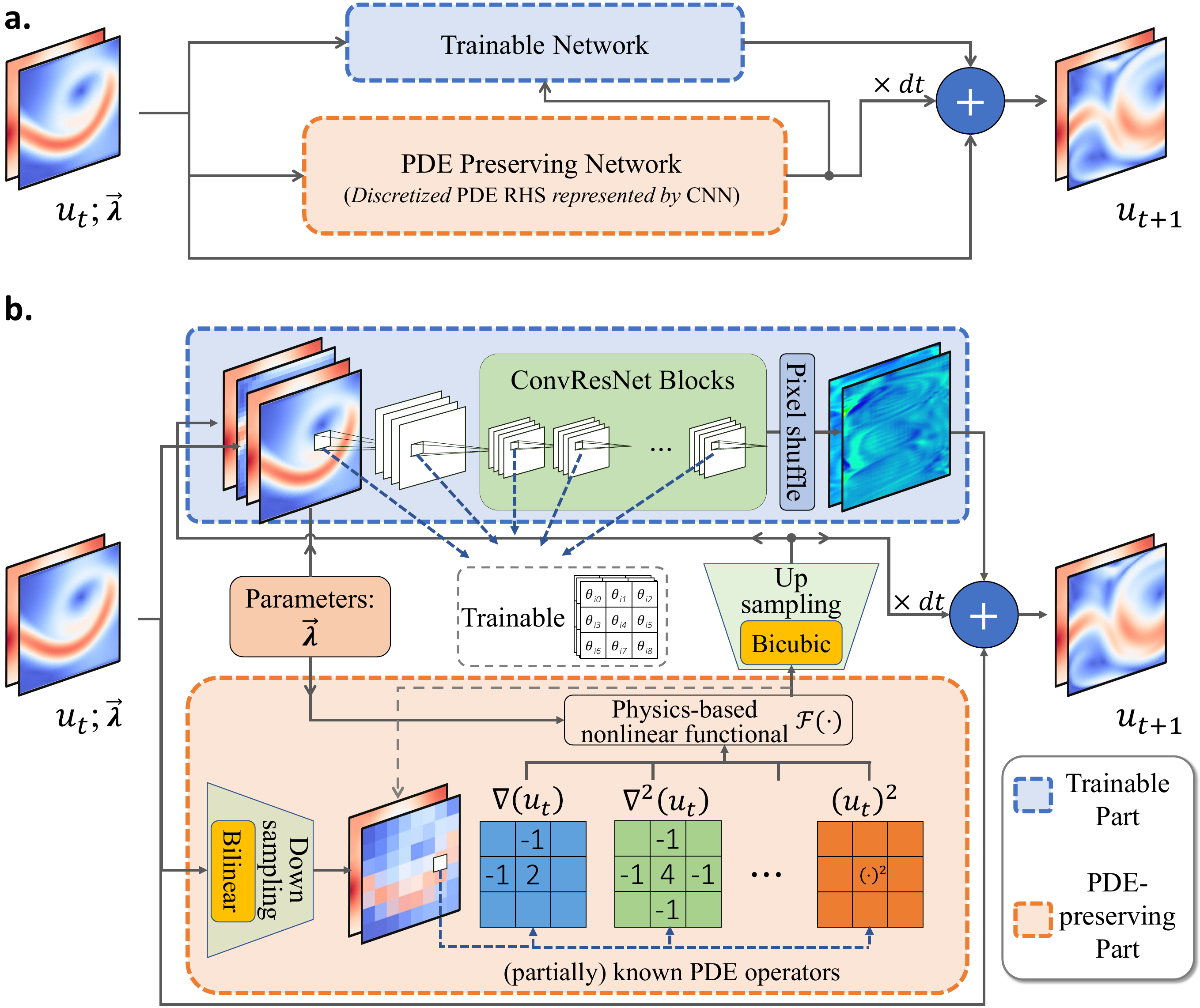}
    \caption{Schematic diagram of the proposed partial differential equation (PDE)-preserved neural network (PPNN). \textbf{a}. A schematic representation illustrating the concept of the PPNN framework. \textbf{b}. A detailed schematic of the ConvResNet-based PPNN, which consists of the trainable part and the PDE-preserving part. The two portions of PPNN are combined together in a multi-resolution setting. The discretized form of the governing PDEs are embedded into the network structure via prescribed convolutions filters and the residual connection.}
    \label{fig:PPNN}
\end{figure}

This study focuses on the learning architecture design for improving the robustness, stability, and generalizability of data-driven next-step predicting models, which commonly suffer from considerable error accumulations due to the auto-regressive formulation and fails to operate in a long-span model rollout. 
In contrast to existing models which are black-box, we propose a PDE-preserved neural network (PPNN) architecture inspired by the relationship between network structures and PDEs, by hypothesizing that the predictive performance can be significantly improved if the network is constructed by preserving (partially) known governing PDEs of the spatiotemporal dynamics to be learned. Specifically, the known portion of the governing PDEs in discrete forms are preserved in residual connection blocks. As shown in Fig.~\ref{fig:PPNN}\textbf{a}, the PPNN architecture features a residual connection which consists of two parts: a trainable network and a PDE preserving network, where the right hand side (RHS) of the governing PDE, discretized on finite difference grid, is represented by a convolution neural network. The weights of the PDE preserved convolutional residual component are determined by the discretization scheme and remain constant during training. 

However, in practice, neural solvers are expected to roll out much faster than numerical solvers, and the time step $\Delta t$ would be orders of magnitude larger than that used in conventional numerical solvers, which may lead to catastrophic stability issues if naively embedding the discretized PDE into the neural network. To this end, we implement a multi-resolution PPNN based on the convolutional (conv) ResNet backbone (shown in Fig. \ref{fig:PPNN}\textbf{b}), where PDE-preserving blocks work on a coarse grid to enable stable model rollout with large evolving steps. This is achieved by using the bilinear down-sampling and bicubic up-sampling algorithms to auto-encode the PDE-preserved hidden feature in a low-resolution space, which is then fed into the main residual connection in the original high-resolution space. 
 
Together with the trainable block, which consists of decoding-encoding convResNet blocks defined on the fine mesh, PPNN enables predictions at a high resolution. Moreover, the network is conditioned on physical parameters $\bm{\lambda}$, enabling fast parametric inference and generalizing over the high-dimensional parameter space. (More details are discussed in the \nameref{sec:methodology} section.)

In this section, we evaluate the proposed PDE structure-preserved neural network (PPNN) on three nonlinear systems with spatiotemporal dynamics, where the governing PDEs are known or partially-known a priori. Specifically, the spatiotemporal dynamics governed by FitzHugh-Nagumo reaction diffusion (RD) equations, Burgers' equations, and incompressible Navier-Stokes (NS) equations with varying parameters $\bm{\lambda}$ (e.g., IC, diffusion coefficients, Reynolds number, etc.) in 2D domains are studied. In particular, we will study the scenarios where either fully-known or incomplete/inaccurate governing PDEs are preserved. To demonstrate the merit of preserving the discrete PDE structure in ConvResNet, the proposed PPNN is compared with the corresponding black-box ConvResNet next-step model as a baseline, which is a CNN variant of the MeshGraphNet~\cite{pfaff2020learning} (see section \nameref{sec:methd2}). For a fair comparison, the network architecture of the trainable portion of the PPNN is the same as the black-box baseline model. Moreover, all models are compared on the same sets of training data in each test case. The generalizability, robustness, training and testing efficiency of the PPNN are investigated in comparison with its corresponding blackbox baseline. It is noted that the novelty of this work lies not in exploring varied methods for learning closures for traditional PDE solvers but in the inventive integration of known physical laws into the architecture of convolutional residual neural networks. We, therefore, consider it critical to compare the PPNN with its black-box counterpart, which learn from data without explicit integration of the underlying physics. This comparison enables us to highlight the unique benefits of integrating known physics into deep learning models, an area that has, to date, received limited attention. Given the prevalence of black-box neural networks in data-driven surrogate modeling where the governing PDEs are often known or partially known, this comparison is both relevant and fair. We believe that this provides a valuable perspective and a substantial contribution to the field. Moreover, it is also worth noting that, PPNN is not constrained to any specific DNN architectures. Rather, we demonstrate that it serves as a versatile framework that can be synergistically combined with a variety of DNN architectures such as U-Net~\cite{ronneberger2015u} -- widely recognized for its multi-scale structure, 
and Vision Transformer (ViT)~\cite{dosovitskiy2020image}, which has become the backbone for most computer vision tasks. (see section \nameref{sec:unetViT}).  Moreover, the relationship between the PDE-preserving portion of PPNN and numerical solvers is discussed. Note that we use the same network setting, i.e., same network structure, hyperparameters and training epochs, for all the test cases (except for the NS system, which has slight modifications adapting to three state variables). More details about the neural network settings can be found in Section \ref{sec:nn} in supplementary information.

All the DNN predictions are evaluated against the high-resolution fully-converged numerical solutions as the reference using a full-field error metric $\epsilon_t$ defined at time step $t$ as,
\begin{linenomath*}
\begin{equation}
    \epsilon_t = \frac{1}{N}\sum_{i=1}^{N}\frac{\left\lVert f_{\theta}(\hat{\bm{u}}_{t-1},\, \bm{\lambda}_i|\Tilde{\bm{\theta}}) + \hat{\bm{u}}_{t-1} - \bm{u}_{t}(\bm{\lambda}_i)\right \rVert_2}{\left	\lVert\bm{u}_{t}(\bm{\lambda}_i)\right\rVert_2},
\end{equation}
\end{linenomath*}
where $N$ indicates the number of the testing physical parameters $\bm{\lambda}_i$, $\bm{u}_t(\bm{\lambda}_i)$ is the reference solution at time step $t$ corresponding to the physical parameter $\bm{\lambda}_i$, $f_{\theta}$ represents the trained neural network function with optimized weights $\Tilde{\bm{\theta}}$, and $\hat{\bm{u}}_{t-1}$ represents the state predicted by the model at previous time step $t-1$,
\begin{linenomath*}
\begin{equation}
\begin{split}
    \hat{\bm{u}}_{t} &= f_{\theta}(\hat{\bm{u}}_{t-1},\, \bm{\lambda}_i|\bm{\theta}) + \hat{\bm{u}}_{t-1},\, t\in[2,n]\\
    \hat{\bm{u}}_{1} &= f_{\theta}(\bm{u}_{0}(\bm{\lambda}_i),\, \bm{\lambda}_i|\theta) + \bm{u}_{0}(\bm{\lambda}_i)
\end{split}
\end{equation}
\end{linenomath*}
where $n$ is the number of testing steps, $\bm{u}_{0}(\bm{\lambda}_i)$ represents the initial condition given $\bm{\lambda}_i$. For brevity, numerical details for each case are given in Section \ref{sec:detail} of the supplementary information.

\subsection*{When the governing PDEs are fully known} \label{sec:fully_known}
We herein consider three well-known spatiotemporal PDEs (e.g., the FitzHugh–Nagumo reaction diffusion equations, the Viscous Burgers’ equation and the Naiver-Stokes equations) when the closed-form equations are fully known.

\paragraph{FitzHugh–Nagumo reaction diffusion equations}
We first consider a spatiotemporal dynamic system governed by the FitzHugh–Nagumo equations with periodic BCs, which is a generic model for excitable media. The main part of the FitzHugh–Nagumo model is reaction-diffusion (RD) equations,
\begin{linenomath*}
\begin{equation}
    \frac{\partial \bm{u}}{\partial t} = \gamma \nabla^2 \bm{u} + \bm{R}(\bm{u})\,,\quad t\in[0,T],
\end{equation}
\end{linenomath*}
where $\bm{u} = \left[u(x,y,t),\, v(x,y,t)\right]^T \in \mathbb{R}^2$ are two interactive components, $\gamma$ is the diffusion coefficient, $T=0.2$s is the time length we simulated, and $\bm{R}(\bm{u}) = \left[R_u(u,v),\, R_v(u, v)\right]^T$ are source terms for the reaction, 
\begin{linenomath*}
\begin{equation}
    \begin{split}
        R_u(u, v) &= u - u^3 - v + \alpha,\\
        R_v(u, v) &= \beta(u-v),
    \end{split}
    \label{eq:reaction}
\end{equation}
\end{linenomath*}
where $\alpha = 0.01$ represents the external stimulus and $\beta = 0.25$ is the reaction coefficient. The initial condition (IC) $\bm{u}_0$ is a random field and generated by randomly sampling from a normal distribution,
\begin{linenomath*}
\begin{equation}
    u(x,y,0), v(x,y,0) \sim \mathcal{N}(0,1),
    \label{eq:rd_ic}
\end{equation}
\end{linenomath*}
which is then linearly scaled to $[0.1, 1.1]$. Given different ICs and diffusion coefficients $\gamma$, varying dynamic spatial patterns of neuron activities can be simulated. Here, the next-step neural solvers are trained to learn and used to predict the spatiotemporal dynamics of varying modeling parameters (i.e., ICs and diffusion coefficients). Namely, we attempt to build a surrogate model in a very high-dimensional parameter space $\bm{\lambda}\in\mathbb{R}^d$, where $d = 65,537$, since the dimensions for IC and diffusion coefficient are $256^2$ and $1$, respectively. The reference solutions are obtained on the simulation domain $(x,y)\in[0,\,6.4]\times [0, 6.4]$, discretized with a fine mesh of $256\times256$ grids, based on the finite difference method.

\begin{figure}[t!]
    \centering
    \includegraphics[width=\textwidth]{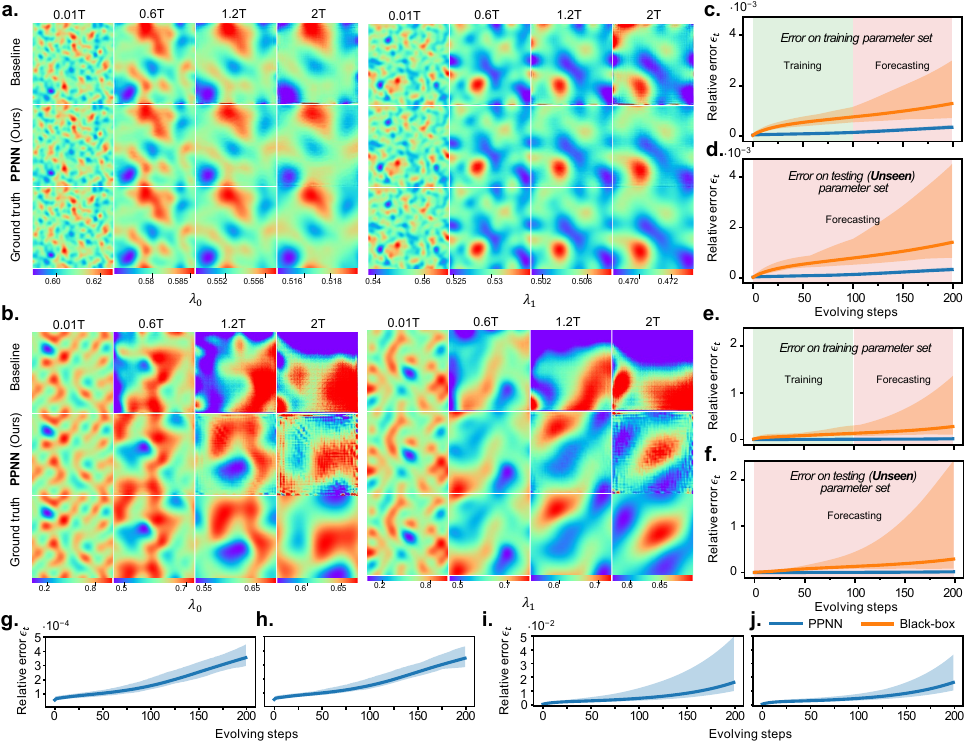}
    \caption{Prediction comparison in the reaction-diffusion (RD) case and viscous Burgers' case. \textbf{a} and \textbf{b}, Predicted solution snapshots of $u$ for the RD equations (\textbf{a}), and the velocity magnitude $\lVert\bm{u}\rVert_2$ for the Burgers' equations (\textbf{b}) at different time steps and unseen parameters, obtained by black-box ConvResNet (baseline model, first rows), and  partial differential equation preserved neural network (PPNN, our method, second rows), compared against ground truth (high-resolution numerical simulation, third rows). ${\lambda}_0$, ${\lambda}_1$ are randomly selected testing (unseen) parameters in each system. \textbf{c} - \textbf{f}, Relative prediction error $\epsilon_t$ of PPNN (blue lines \blueline) and black-box ConvResNet baseline (orange lines \orangeline) for the RD dynamics (\textbf{c}-\textbf{d}) and Burgers' equations (\textbf{e} - \textbf{f}), averaged on 100 randomly sampled training parameters $\bm{\lambda}$ (\textbf{c}, \textbf{e}) and testing (unseen) parameters (\textbf{d}, \textbf{f}). The shaded area shows the maximum and minimum relative errors of all testing trajectories. \textbf{g}, \textbf{h}, Zoom in views of the relative error curve of PPNN shown in \textbf{c} (\textbf{g}) and \textbf{d} (\textbf{h}), respectively. \textbf{i}, \textbf{j}, Zoom in views of the relative error curve of PPNN shown in \textbf{e} (\textbf{i}) and \textbf{f} (\textbf{j}), respectively. 
    }
    \label{fig:rd_bg}
\end{figure}

Figure~\ref{fig:rd_bg}\textbf{a} shows the PPNN-predicted solution snapshots of the RD equations at four randomly selected test parameters (i.e., randomly generated ICs and unseen diffusion coefficients). The prediction results of baseline black-box ConvResNet (first row) and the proposed PPNN (second row) are compared against the ground truth reference (third row).  It can be seen that both models agree with the reference solutions for $t < 0.6T$, showing good generalizability on testing ICs and $\gamma$ for a short-term model rollout. However, the error accumulation becomes noticeable for the black-box baseline when $t > T$, and the spatial patterns of the baseline predictions significantly differ from the reference at $t = 2T$, which is an expected issue for the next-step predictors. In contrast, the results of our PPNN have an good agreement with the reference solutions over the entire time span $[0, 2T]$ on all testing parameters, showing great robustness, predictability, and generalizability in both the spatiotemporal domain and parameter space. Predicted solutions on more testing parameters are presented in Fig.~\ref{fig:rdmore}.

To further examine the error propagation in time for both models, the relative testing errors $\epsilon_t$ averaged over 100 randomly selected parameters in training and testing sets are computed and plotted in Fig.~\ref{fig:rd_bg}, where Fig.~\ref{fig:rd_bg}\textbf{c} shows the averaged model-rollout error evaluated on 100 training parameters and the Fig.~\ref{fig:rd_bg}\textbf{d} shows the error averaged on 100 randomly generated testing parameters. (Zoom in views of Fig.~\ref{fig:rd_bg}\textbf{c} and Fig.~\ref{fig:rd_bg}\textbf{d} can be found in Fig.~\ref{fig:rd_bg}\textbf{g} and Fig.~\ref{fig:rd_bg}\textbf{h}, respectively.) The model is only trained within the range of $1T$ ($100\Delta t$), and it is clearly seen that the rollout error of the black-box model significantly grows in the extrapolation range $[T, 2T]$ (from $100\Delta t$ to $200\Delta t$), where $\Delta t = 200\delta t$ is the learning step size which is 200 numerical timesteps $\delta t$. The error accumulation becomes more severe for the unseen testing parameters. However, our PPNN predictions maintain an impressively low error, even when extrapolating twice the length of the training range. Besides, the scattering of the error ensemble is significantly reduced compared to the black-box baseline, indicating great robustness of the PPNN for various testing parameters. 

\paragraph{Viscous Burgers' equation}
\label{sec:burgers}
For the second case, we study the spatiotemporal dynamics governed by the viscous Burgers' equations on a 2D domain with periodic boundary conditions, 
\begin{linenomath*}
\begin{equation}
\frac{\partial \bm{u}}{\partial t} + \bm{u} \cdot \nabla \bm{u} = \nu \nabla^2 \bm{u}\,,\quad t\in[0,T], 
\label{eq:burgers}
\end{equation}
\end{linenomath*}
where $\bm{u} = \left[(u(x,y,t),\, v(x,y,t)\right]^T\in \mathbb{R}^2$ is the velocity vector, $T=2$s is the time length we simulated, and $\nu$ represents the viscosity. The initial condition (IC) $\bm{u}_0$ is generated according to,
\vspace{-5pt}
\begin{linenomath*}
\begin{equation}
\bm{u}_0 = 
\begin{cases}\displaystyle
    u_0 = \sum^4_{i = -4 } \sum^4_{j = -4 } r^{(1)}_{i,j} \sin{\left(i x + j y\right)} + r^{(2)}_{i,j} \cos{\left(i x + j y\right)}\\ \displaystyle
    v_0 = \sum^4_{i = -4 } \sum^4_{j = -4 } r^{(3)}_{i,j} \sin{\left(i x + j y\right)} + r^{(4)}_{i,j} \cos{\left(i x + j y\right)}
\end{cases} \quad r^{(k)}_{i,j} \sim \mathcal{N}(0,1); k=1,2,3,4,
\end{equation}
\end{linenomath*}
where $x, y$ are spatial coordinates of grid points, and $r^{(k)}_{i,j}; k\in{1,2,3,4}$ are random variables sampled independently from a normal distribution. The IC is normalized in the same way as mentioned in the RD case. We attempt to learn the dynamics given different ICs and viscosities. Similar to the RD cases, the parameter space $\mathbb{R}^{d}$ is also high-dimensional ($d = 324$), as the IC is parameterized by $4\times9^2$ independent random variables and the scalar viscosity can also vary in range $[0.02,0.07]$. The reference solution is generated by solving the Burgers' equations on the domain of $(x,y)\in [0,\,3.2]^2$, discretized by a fine mesh of $256 \times 256$ grids using finite difference method. 

The velocity magnitude contours of the 2D Burgers' equation with different testing parameters are shown in Fig.~\ref{fig:rd_bg}\textbf{b}, obtained by the black-box baseline, PPNN, and reference numerical solver, respectively. Note that all the testing parameters are not seen during training. (More predicted solutions on different testing parameters are presented in  Fig.~\ref{fig:bgmore}.) Similar to the RD case, PPNN shows a significant improvement over the black-box baseline in terms of long-term rollout error accumulation and generalizability on unseen ICs and viscosity $\nu$. Due to the strong convection effect, black-box baseline predictions deviate from the reference very quickly, and significant discrepancies in spatial patterns can be observed as early as $t < 0.6T$. In general, the black-box baseline suffers from the poor out-of-sample generalizability for unseen parameters, making the predictions useless. Our PPNN significantly outperforms the black-box baseline, and its prediction results agree with the reference for all testing samples. Although slight prediction noises are present after a long-term model rollout ($t > 1.2T$), the overall spatial patterns can be accurately captured by the PPNN even at the last learning step ($t = 2T$). 
The error propagation of both models is given in Fig.~\ref{fig:rd_bg}, where the rollout errors $\epsilon_t$ at each time step, averaged over 100 randomly selected parameters from training and testing sets, are plotted.
Fig.~\ref{fig:rd_bg}\textbf{e} shows the averaged model rollout error evaluated on 100 training parameters, while Fig.~\ref{fig:rd_bg}\textbf{f} shows the error averaged on $100$ randomly generated parameters, which are not used for training.  Zoom in views of Fig.~\ref{fig:rd_bg}\textbf{e} and Fig.~\ref{fig:rd_bg}\textbf{f} can be found in Fig.~\ref{fig:rd_bg}\textbf{i} and Fig.~\ref{fig:rd_bg}\textbf{j}, respectively. As both models are only trained with the $1T$ ($100\Delta t$) time steps for each parameter in the training set, it is clear that the error of the black-box model grows rapidly once stepping into the extrapolation range $[T,\,2T]$. The error accumulation effect of the black-box model becomes more obvious for those parameters which are not in the training set due to the poor generalizability. In contrast, the error of PPNN predictions remains surprisingly low even in the extrapolation range for both training and testing regimes, and there is nearly no error accumulation. In addition, the error scattering significantly shrinks compared to that of the black-box model, indicating significantly better accuracy, generalizability and robustness of the PPNN compared to the black-box baseline. 

\begin{figure}[t!]
    \centering
    \includegraphics[width=\textwidth]{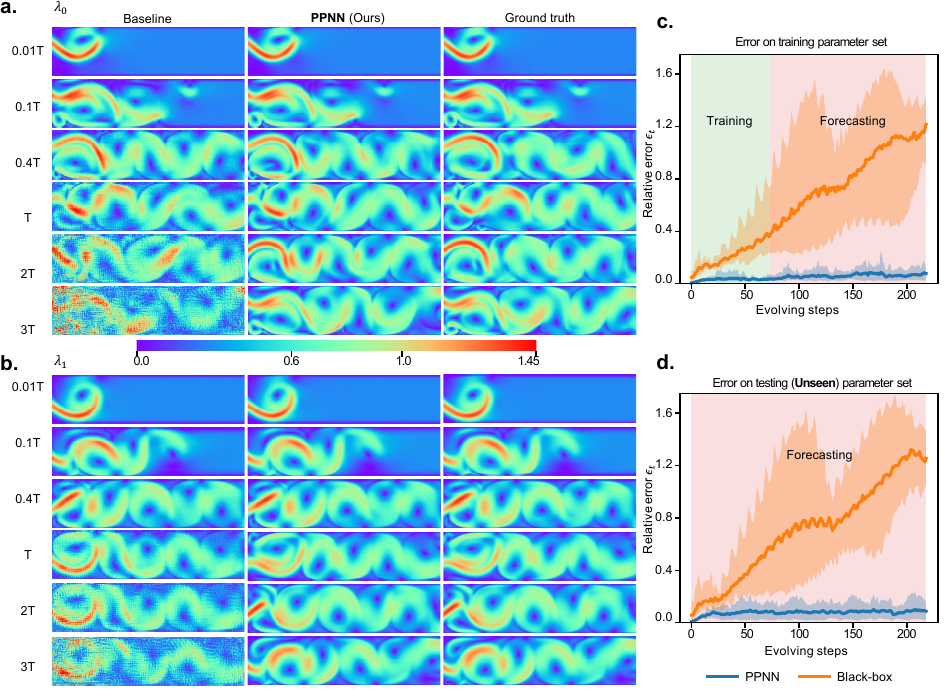}
    \caption{Prediction comparison in the case governed by Naiver-Stocks (NS) equations. \textbf{a}-\textbf{b}, Predicted solution snapshots of velocity magnitude $\lVert \bm{u}\rVert_2$ for the NS equations obtained by black-box ConvResNet (baseline);  partial differential equation preserved neural network (PPNN, Ours), compared against the ground truth (high-resolution numerical simulation), where $\bm{\lambda}_0$ is ($\mathrm{Re} = 2500, y_0 = 0.325$, shown in \textbf{a}), and $\bm{\lambda}_1$ is high Reynolds number ($\mathrm{Re} = 8500, y_0 = 0.575$, shown in \textbf{b}). \textbf{c}-\textbf{d}, Relative prediction error $\epsilon_t$ of PPNN (blue lines \blueline) and black-box ConvResNet baseline (orange lines \orangeline) at different timesteps for the NS equation, averaged on 5 randomly sampled (\textbf{c}) training parameters and (\textbf{d}) testing (unseen) parameters. The shaded areas show the scattering of the relative errors over all testing trajectories. }
    \label{fig:ns}
\end{figure}

\paragraph{Naiver-Stokes equations}
The last case investigates the performance of PPNN to learn an unsteady fluid system exhibiting complex vortex dynamics, which is governed by the 2D parametric unsteady Naiver-Stokes (NS) equations:
\begin{linenomath*}
\begin{equation}
    \begin{split}
        \frac{\partial \bm{u}}{\partial t} + \bm{u} \cdot \nabla \bm{u} &= -\nabla p + \nu \nabla^2 \bm{u}\,,\quad t\in[0,T], \\
        \nabla \cdot \bm{u} &= 0,
    \end{split}
\end{equation}
\end{linenomath*}
where $\bm{u} = \left[u(x,y,t), v(x,y,t)\right]^T \in \mathbb{R}^2$ is the velocity vector, $p(x,y,t) \in \mathbb{R}$ is the pressure, and $\nu = 1/\mathrm{Re}$ represents the kinematic viscosity ($\mathrm{Re}$ is the Reynolds number). The NS equations are solved in a 2D rectangular domain $(x,y)\in [0,4]\times[0,1]$, where a jet with dynamically-changed jet angle is placed at the inlet. Namely, the inflow boundary is defined by a prescribed velocity profile $\bm{u}(0,y,t)$,
\begin{linenomath*}
\begin{equation}
\bm{u}(0,y,t) = 
\begin{bmatrix}
    u(0,y,t)\\
    v(0,y,t)
\end{bmatrix} = 
\begin{bmatrix}
    \exp{\left(-50 \left(y - y_0\right)^2\right)}\\
    \sin{(t)}\cdot\exp{\left(-50 \left(y - y_0\right)^2\right)}
\end{bmatrix}
\end{equation}
\end{linenomath*}
where $y_0$ represents the vertical position of the center of the inlet jet. The outflow boundary condition is set as pressure outlet with a reference pressure of $p(4,y,t)=0$. No-slip boundary conditions are applied on the upper and lower walls. In this case, the neural network models are expected to learn the fluid dynamics with varying Reynolds number $\mathrm{Re}$ and jet locations $y_0$. Namely, a two-dimensional physical parameter vector $\bm{\lambda} = [\mathrm{Re}, y_0]^T$ is considered. In training set, we use five different $\mathrm{Re}$ evenly distributed in the range $\left[2\times10^3, 1\times10^4\right]$ and $9$ different jet locations $y_0$ uniformly selected from $0.3$ to $0.7$.
Figure~\ref{fig:ns}\textbf{a}-\textbf{b} shows the snapshots of velocity magnitude of the NS equations at two representative testing parameters, which are not seen in the training set. To be specific, $\bm{\lambda}_0 = [2500, 0.325]^T$ represents a relatively low Reynolds number $Re = 2500$ with the jet located at $y_0 = 0.325$, while $\bm{\lambda}_1 = [8500,0.575]^T$ is a higher Reynolds number case ($\mathrm{Re} = 8500$) with the jet located at $y_0 = 0.325$. The rollout prediction results of the PPNN and baseline black-box ConvResNet are compared with the ground truth reference. Although both models can accurately capture the spatiotemporal dynamics at the beginning stage (when $t\leq0.4T$), showing good predictive performance for the unseen parameters for a short-term rollout, the predictions by the black-box model are soon overwhelmed by the noises due to the rapid error accumulation ($t>T$). However, the proposed PPNN significantly outperforms the black-box baseline as it managed to provide accurate rollout predictions even at the last testing steps ($t = 3T$), which extrapolate as three times long as the training range, indicating preserving the PDE structure can effectively suppress the error accumulation, which is unavoidable in most auto-regressive neural predictors. 
To further investigate the error propagation in time for both models, we plot the relative testing errors $\epsilon_t$ against time in Fig.~\ref{fig:ns}\textbf{c}-\textbf{d}, which are averaged over $5$ randomly selected parameters in both training (Fig.~\ref{fig:ns}\textbf{c}) and testing sets (Fig.~\ref{fig:ns}\textbf{d}). We can clearly see that PPNN managed to maintain  low rollout error in both training and extrapolation ranges, in contrast to the significantly higher error accumulation in the black-box baseline results. In particular, the black-box model relative error visibly grows only after a short-term model rollout and increases rapidly once it enters the extrapolation range even for testing on the training parameter set (Fig.~\ref{fig:ns}c), and the errors are accumulated even faster for the testing on unseen parameters (Fig.~\ref{fig:ns}d). On the contrary, our PPNN has almost no error accumulation and performs much more consistently between the training and extrapolation ranges, with significantly lower rollout errors. The results again demonstrate outstanding predictive accuracy and generalizability of the proposed method. Besides, PPNN also shows a significantly smaller uncertainty range, indicating great robustness among different testing parameters. 

\subsection*{When the governing PDEs are partially known}
In real-world applications, the underlying physics behind complex spatiotemporal phenomena might not be fully understood, and thus the governing equations can be incomplete, e.g., with unknown source terms, inaccurate physical parameters, or uncertain I/BCs. Such partially-known physics poses great challenges to the traditional simulation paradigm since the governing equations are partially known. Nonetheless, the incomplete prior knowledge can be well utilized in our proposed PPNN framework, where preserving partially-known governing PDE structures can still bring significant merits to data-driven spatiotemporal learning and prediction, which will be discussed in this subsection.   

\begin{figure}[t!]
    \centering
    \includegraphics[width=\textwidth]{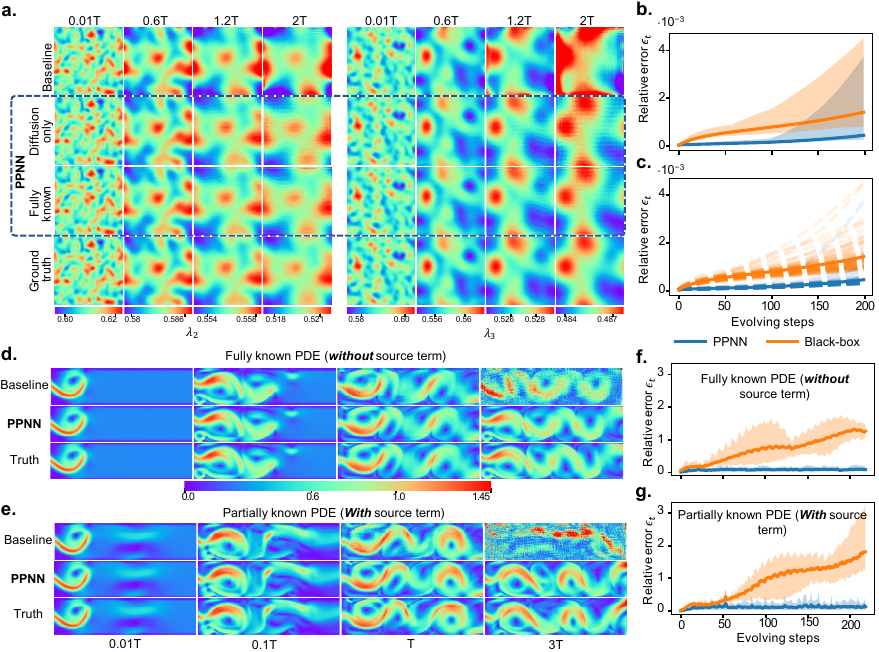}
    \caption{Prediction comparison in the cases where the governing equations are partially known. \textbf{a}, Predicted solution snapshots of $u$ for the reaction-diffusion (RD) equations at different time steps and unseen parameters, obtained by black-box ConvResNet (baseline model), and  partial differential equation (PDE)-preserved neural network (PPNN, preserving diffusion terms only), and PPNN (preserving complete FitzHugh–Nagumo RD equations), compared against ground truth. ${\lambda}_2$ and ${\lambda}_3$ are two randomly selected testing (unseen) parameters. 
    \textbf{b}-\textbf{c}, Averaged relative testing error $\epsilon_t$ of the PPNN with incomplete PDE (blue lines \blueline) and Black-Box ConvResNet baseline (orange lines \orangeline) for the RD dynamics evaluated on 100 randomly generated testing (unseen) parameters (same parameters as shown in Fig.~\ref{fig:rd_bg}\textbf{c}). Shaded areas in \textbf{b} indicate envelopes of the maximum and minimum relative errors of all testing trajectories, while the dash lines in \textbf{c} indicate the relative error of each test trajectory. \textbf{d}-\textbf{e}, Predicted solution snapshots of flow velocity magnitude $\lVert \bm{u}\rVert_2$ obtained by black-box ConvResNet (baseline), PPNN (ours), compared against ground truth (high-resolution numerical simulation) of the NS equations without (\textbf{d}) and with (\textbf{e}) an unknown magnetic source term, respectively. The PPNN only preserves a NS equation portion for both scenarios, which are at the same testing (unseen) parameter $\bm{\lambda} = [9000, 0.475]^T$, which is not in the training set. \textbf{f}-\textbf{g}, Relative prediction errors $\epsilon_t$ of the PPNN (blue line \blueline) and black-box ConvResNet baseline (orange line \orangeline) for the NS equation with (\textbf{e}) and without (\textbf{f}) a unknown magnetic body force, averaged on five randomly sampled unseen parameters. The shaded area shows the scattering of relative errors for all testing trajectories.}
    \label{fig:partially}
\end{figure}

\paragraph{Reaction diffusion equations with unknown reaction term}
We first revisit the aforementioned FitzHugh–Nagumo RD equations. Here, we consider the scenario where only the diffusion phenomenon is known in the FitzHugh–Nagumo RD dynamics. Namely, the reaction source terms remain unknown and PPNN only preserves the incomplete RD equations, i.e., 2D diffusion equations, 
\begin{linenomath*}
\begin{equation}
    \frac{\partial\bm{u}}{\partial t} = \gamma \nabla^2 \bm{u}.
\end{equation}
\end{linenomath*}
All the case settings remain the same as those discussed previously. Although incomplete/inaccurate prior knowledge about the RD system is preserved, our PPNN still shows a significant advantage over the black-box baseline. Figure~\ref{fig:partially}\textbf{a} compares the snapshots of reactant $u$ at two randomly selected unseen parameters $\bm{\lambda}_2$ and $\bm{\lambda}_3$ predicted by black-box baseline model (first rows), PPNN with the diffusion terms preserved only (second rows), PPNN with the complete RD equation preserved (third rows), against the ground truth (fourth rows). The PPNNs preserving either complete or incomplete RD equations accurately capture the overall patterns and well agree with the reference solutions, while the black-box baseline shows notable discrepancy and large errors, particularly at $t = 2T$, which is the twice of the training phase length. At the last extrapolation step, the prediction results of black-box baseline show some visible noise and are less smooth compared to the results by preserving the complete RD equation, indicating that lack of the prior information on the reactive terms could slightly reduce the improvement by PPNN. Figure~\ref{fig:partially}\textbf{b}-\textbf{c} shows the relative model rollout errors averaged over 100 test trajectories, which are not seen in the training set. The shaded area in the upper panel shows the error distribution range of these 100 test trajectories. Even the preserved PDEs are not complete/accurate, the mean relative error (blue line \blueline) remains almost the same as the PPNN with fully-known PDEs (see Fig.~\ref{fig:rd_bg}\textbf{a}), which is significantly lower than that of the black-box baseline (orange line \orangeline), showing a great advantage of preserving governing equation structures even if the prior physics knowledge is imperfect. Compared to the PPNN with fully-known PDEs, the error distribution range by preserving partially-known PDEs is increased and error ensemble is more scattered, implying slightly decreased robustness. Although the envelope of the error scattering for incomplete PDEs is much larger than that of the case with fully-known PDEs, this is due to a single outlier trajectory, which can be seen in Fig.~\ref{fig:partially}\textbf{c}. This indicates embedding a incomplete PDE terms will leads to restricted performance of PPNN when the disregarded term plays an important role in the dynamic system. In general, the standard deviation of the error ensemble from the PPNN with partially-known PDE ($\sigma = 1.123\times 10^{-4}$) is still significantly lower than that of the black-box baseline ($\sigma = 3.412\times 10^{-4}$). In comparison, the standard deviation of errors in PPNN with fully-known PDEs over the 100 test trajectories is $0.854\times 10^{-4}$.   

\paragraph{Naiver-Stokes equations with an unknown magnetic field}
In the second case, we consider the the complex magnetic fluid dynamic system governed by Naiver-{Stokes} equations with an unknown magnetic field:
\begin{linenomath*}
\begin{equation}
    \begin{split}
        \frac{\partial \bm{u}}{\partial t} + \bm{u} \cdot \nabla \bm{u} &= -\nabla p + \nu \nabla^2 \bm{u} + \bm{F},\\
        \nabla \cdot \bm{u} &= 0,
    \end{split}
\end{equation}
\end{linenomath*}
where $\bm{u} = [u, v]^T$ is the velocity vector; $p$ is the pressure; while $\nu$ represents the kinematic viscosity. Here $\bm{F} = [F_x, F_y]^T$ represents the body force introduced by a magnetic field:
\begin{linenomath*}
\begin{equation}
\begin{split}
    F_x &= m H \frac{\partial H}{\partial x}\, ,\qquad F_y = m H \frac{\partial H}{\partial y}\\
    H(x,y) &= \exp{\left[-8 \left(\left(x-L/2\right)^2 + \left(y-W/2\right)^2\right) \right]}
\end{split}
\end{equation}
\end{linenomath*}
where $m=0.16$ is the magnetic susceptibility, and $H$ is a time-invariant magnetic intensity.
The contour of the magnitude of the body force source term is shown in the supplementary information (see Fig.~\ref{fig:magnetic}). In this case, the magnetic field remains unknown and PPNN only preserves the NS equation without the magnetic source term. All the other case settings remain unchanged as described in the Naiver-Stokes equation case.

Similar to what we observed in the example of RD equations with the unknown reaction term, the PPNN still remains a significant advantage over the black-box baseline even by preserving an incomplete physics of the flow in a magnetic field. Fig.~\ref{fig:partially}\textbf{d}-\textbf{e}, shows the velocity magnitude $\lVert\bm{u}\rVert_2$ results of the flow with (Fig.~\ref{fig:partially}\textbf{d}) or without (Fig.~\ref{fig:partially}\textbf{e}) a magnetic field at the same testing parameter ($\bm{\lambda} = [\mathrm{Re}=9000$, $y_0 = 0.475]^T$), predicted by the PPNN and black-box ConvResNet, compared against the reference solution. For both scenarios, only the NS equation portion is preserved in the PPNN, i.e., the magnetic field remains unknown. Figure~\ref{fig:partially}\textbf{d} shows the solution snapshots for the flow without the magnetic field (i.e., PPNN preserving the complete physics), while Fig.~\ref{fig:partially}\textbf{e} shows the predictions of the flow with the magnetic field (i.e., PPNN preserving an incomplete physics). Comparing the reference solutions at upper and lower panels, the spatiotemporal patterns of the flow fields exhibit notable differences for the cases with and without magnetic fields. In both scenarios, the black-box baseline model suffers from the long-term model rollout, particularly for the flow within the magnetic field, the black-box baseline completely fails to capture the physics when $t > 2T$. In both scenarios, the PPNN outperforms the black-box baseline. In particular at the last time step $t=3T$, which is three times the training phase length, the black-box predictions are totally overwhelmed by noise, while our PPNN predictions still agree with the reference very well. Compared to the case preserving the complete physics (Fig.~\ref{fig:partially}\textbf{d}), a slight deviation from the reference solution can be observed in the PPNN predictions of the flow with an unknown magnetic field (Fig.~\ref{fig:partially} \textbf{e}), indicating that incomplete prior knowledge could slightly affect the PPNN performance negatively. Nonetheless, preserving the partially-known PDE structure still brings significant merit.
The error propagation is shown in Fig.~\ref{fig:partially}\textbf{f}-\textbf{g}. The relative model rollout errors are averaged over 5 randomly selected unseen parameters for the systems with (Fig.~\ref{fig:partially}\textbf{f}) and without (Fig.~\ref{fig:partially}\textbf{g}) the magnetic field. Comparing to the PPNN with completely-known PDEs, the PPNN preserving incomplete/inaccurate prior knowledge does show a slight increment in the mean relative error $\epsilon_t$ as well as the error scattering, which implies a slight decrease in the robustness. However, the significant advantage over the black-box baseline remains, and almost no error accumulation is observed in PPNN for both scenarios. 

\subsection*{When encoding completely mis-specified PDE terms}

In the scenarios we have presented so far, the preserved PDE operators are incomplete but not entirely incorrect, which allows the PPNN model to outperform the black-box baseline. However, in certain situations, our prior knowledge about the target system may sometimes be entirely incorrect. In this section, we consider an extreme case where the preserved physics are completely mis-specified.

To investigate this, we consider a system governed by the viscous Burgers' equation (Eq.\ref{eq:burgers}), but we preserving a reaction term (Eq.\ref{eq:reaction}) in the PPNN that does not reflect the actual physical processes at all. This experiment aims to assess our model's performance when the physics are completely mis-specified and determine how this mismatch affects the overall model performance.
\begin{figure}[!htp]
    \centering
    \includegraphics[width=0.4\textwidth ]{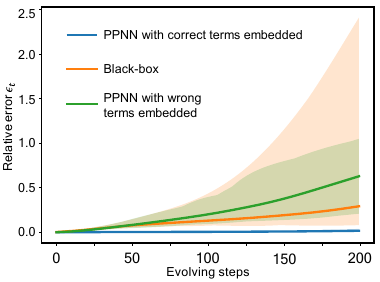}
    \caption{ Relative error $\epsilon_t$ comparison when wrong terms are embedded in  partial differential equation preserved neural network (PPNN), tested on 2D Burgers' equation. The relative error of PPNN (blue line \blueline), its black-box counterpart (orange line\orangeline), and PPNN with completely wrong partial differential equation terms (green line \greenline) tested on unseen parameters is shown in the figure. Solid lines show the mean relative error, while the shaded areas show the distribution of all the 100 sample trajectories.}
    \label{fig:wrongEmbed}
\end{figure}
These results show the model's behavior under the extreme conditions, when the underlying physics might be either completely unknown or inaccurately specified. As depicted in Fig.~\ref{fig:wrongEmbed}, the performance of the PPNN model suffers when the embedded PDE terms diverges significantly from the actual physics. In such cases, the performance of the PPNN model is adversely affected, with its predictions being worse than those of the black-box method. As expected, this result suggests that an certain level of alignment between the embedded PDEs and the underlying physics is essential for optimal performance. Particularly, the error distribution range of the PPNN model is significantly narrower than that of the black-box baseline, indicating that mis-specified embedded PDEs also impose an inductive bias to the deep learning model.

\subsection*{Training and inference cost}
We have demonstrated that the proposed PPNN significantly improves the accuracy, generalizability, and robustness of next-step neural predictors by preserving the mathematical structure of the governing PDEs. Since the PPNN has a more complex network structure than the black-box baseline, it is worthwhile to discuss the training and inference costs of the PPNN and its comparison with the corresponding black-box baseline and the reference numerical solvers.

\begin{figure}[t!]
    \centering
    \includegraphics[width=\textwidth]{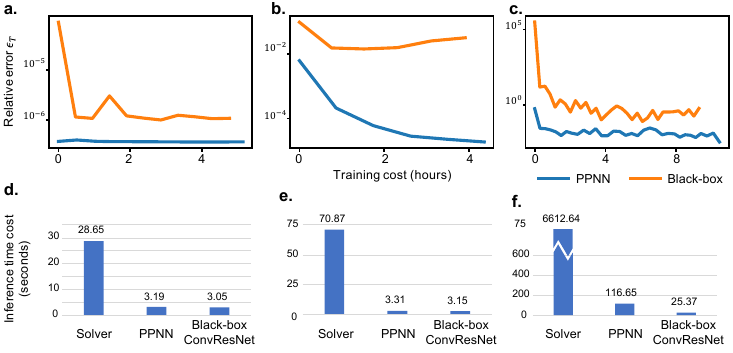}
    \caption{Testing error $\epsilon_T$ during training and the inference cost of  partial differential equation (PDE)-preserved neural network (PPNN), black-box baseline and numerical solvers. \textbf{a}-\textbf{c}, Averaged relative test error at the last time step $\epsilon_T$ of PPNN (blue lines) and black-box ConvResNet (orange lines) in training process of different cases in the section \nameref{sec:fully_known} (governed by reaction-diffusion (\textbf{a}), Burgers' equation (\textbf{b}), and Naiver-stokes equations (\textbf{c})). \textbf{d}-\textbf{f}, inference time cost of numerical solver, PPNN and black-box ConvResNet in the case (governed by reaction-diffusion (\textbf{d}), Burgers' equation (\textbf{e}), and Naiver-stokes equations (\textbf{f})). The reaction-diffusion and Burgers cases are inferred (simulated) on a NVIDIA RTX 3070 GPU and the time is measured for infer/simulate 10 trajectories for 200 time steps. The Navier-Stocks case is inferred/simulated on a single of Intel Xeon Gold 6138 CPU and the time is measured for infer/simulate 1 trajectory for 219 time steps.}
    \label{fig:cost}
\end{figure}

\paragraph{Training cost} As shown in Fig.~\ref{fig:cost}\textbf{a}-\textbf{c}, the averaged relative (rollout) prediction error $\epsilon_T$ on $n$ testing parameters $\lambda$ at the last time step $T$ in the training process ($n=8$ in RD, $n=6$ in Burgers and $n=5$ in NS). For all the cases, PPNN features a significantly (orders of magnitude) lower error than the black-box model from a very early training stage. This means that, to achieve the same (if not higher) level of accuracy, our PPNN requires significantly less training cost compared to the black-box baseline. In addition, under the same training budget, the PPNN is much more accurate than the black-box baseline, demonstrating the merit of PPNN by leveraging the prior knowledge for network architecture design.

\paragraph{Inference cost} The inference costs of different neural networks and numerical solvers on the three testing cases (see section~\nameref{sec:fully_known}) with the model rollout length of $T$ are summarized in Fig.~\ref{fig:cost}\textbf{b}-\textbf{f}. Due to the fast inference speed of neural networks, both next-step neural models show significant speedup compared to the high-fidelity numerical solvers. In particular, the speedup by the PPNN varies from $10\times$ to $60\times$ without significantly sacrificing the prediction accuracy. Such speedup will become more tremendous considering a longer model rollout and enormous repeated model queries on a large number of different parameter settings, which are commonly required in many-query applications such as optimization design, inverse problems, and uncertainty quantification. Note that all models are compared on the same hardware (GPU or CPU) to eliminate the difference introduced by hardware. However, as most legacy numerical solvers can only run on CPUs, the speedup by neural models can be much more significant if they leverage massive GPU parallelism. Admittedly, adding the PDE-preserving part inevitably increases the inference cost compared to the black-box baseline, but the huge performance improvement by doing so outweighs the slight computational overhead, as demonstrated in section~\nameref{sec:fully_known}. We have to point out that the computation of the PDE-preserving portion is not fully optimized, particularly in the NS case, where low-speed I/O interactions reduce the overall speedup ratio compared to the numerical solver based on the mature CFD platform OpenFOAM. Further performance improvements are expected by customized code optimizations in future work.     

\subsection*{Relationship between the PDE-preserving portion and numerical solvers}
\begin{figure}[!htp]
    \centering
    \includegraphics[height=0.81\textheight]{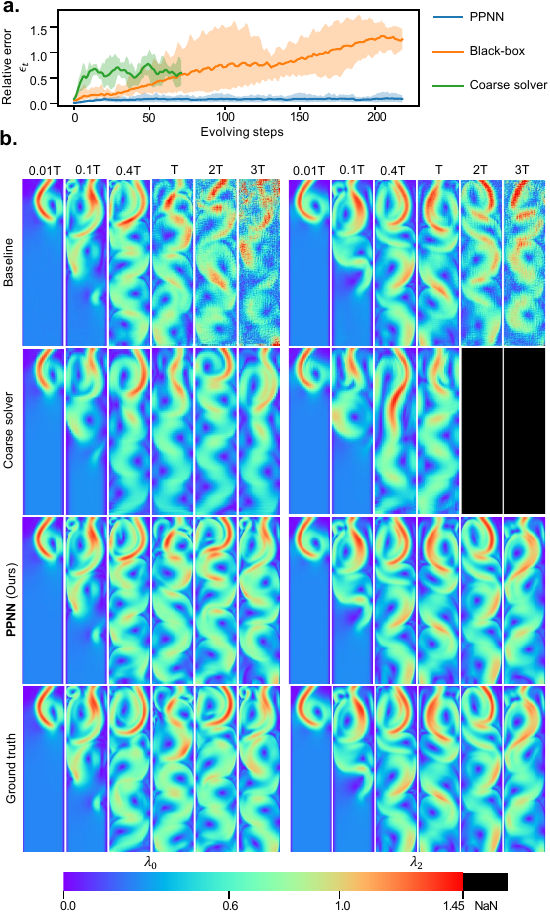}
    \caption{Prediction comparison between  partial differential equation (PDE)-preserved neural network (PPNN), the PDE-preserving part of PPNN (numerical solver results on a coarse mesh), the black-box baseline, and the label data. \textbf{a}, Relative error at different time steps of PPNN (blue line \blueline), Black-Box neural network (orange line \orangeline) and the coarse solver (green line \greenline) compared to the ground truth results obtained by icoFoam on a fine mesh . The relative error is an averaged value of 5 test trajectories with randomly sampled parameters, these parameters are not in the training set. The shaded area shows the maximum and minimum relative error of these testing trajectories. In coarse solver, 2 of the testing trajectories diverged (NaN) at the $72^{nd}$ step thus the green curve (\greenline) stops at the $71^{st}$ step.
    \textbf{b}, The contours show predicted solution snapshots of velocity magnitude $\lVert u\rVert_2$ for the NS equations, obtained by black-box ConvResNet (baseline), PDE-preserving part only(coarse solver), and PPNN (ours); compared against ground truth (high-resolution numerical simulation), where $\bm{\lambda}_0 = [2500, 0.325]^T$ and $\bm{\lambda}_2 = [9000, 0.475]^T$, which are unseen in the training set. The black color indicates NaN, i.e., solution blow up.}
    \label{fig:dis:c}
\end{figure}
The advantages of the proposed PPNN over the pure black-box baseline mainly come from ``baking" the prior knowledge into the network architecture. As discussed above, the mathematical structures of the governing physics are encoded into the PPNN based on the relationship between neural network structures and differential equations. From the numerical modeling perspective, if our understanding of the underlying physics is complete and accurate (i.e., complete governing PDEs are available), the PDE-preserving portion in PPNN can be interpreted as a numerical solver with the explicit forward Euler scheme defined on a coarse mesh. For simplicity, we here refer to this numerical solver derived from the fully-known PDE preserving part as the ``coarse solver". It is interesting to see how well it performs by the coarse solver only when governing equations and IC/BCs/physics properties are fully known.

We use the NS case as an example. Fig.~\ref{fig:dis:c}b shows the magnitude of velocity $\lVert\bm{u}\rVert_2$ predicted by the PPNN, black-box ConvResNet and coarse solver, respectively, compared against the reference solution. Two representative testing parameters are studied here, one is at a lower Reynolds number $\mathrm{Re} = 2500, y_0 = 0.325$ (Fig.~\ref{fig:dis:c}b, $\lambda_0$), and the other is at a higher Reynolds number $\mathrm{Re} = 9000, y_0 = 0.475$ (Fig.~\ref{fig:dis:c}b, $\lambda_2$). It is clear that the predictions by the coarse solver noticeably deviate from the reference solution from $0.4T$, and most vortices are damped out due to the coarse spatial discretization. This becomes worse in the higher Reynolds number scenario, where the coarse solver predicted flow field is unphysical at $0.1T$ and the simulation completely diverged at $t = 1.16T$, because of the large learning timestep making traditional numerical solvers fail to satisfy the stability constraint.

As shown in the error propagation curves in Fig.~\ref{fig:dis:c}a, the coarse solver has large prediction errors over the testing parameter set from the very beginning, which is much higher than that of the black-box data-driven baseline. Since several of the testing trajectories by the coarse solver diverges quickly after 70 evolving steps, the error propagation curve stops. 

This figure again empirically demonstrates that the PPNN structure not only overcomes the error accumulation problem in black-box methods, but also significantly outperforms numerical solvers by simply coarsen the spatiotemporal grids. On the other hand, for those trajectories that do not diverge, the coarse solver's relative errors are limited to a certain level, which is in contrast to black-box, data-driven methods where the error constantly grows due to the error accumulation. This phenomenon implies that preserving PDEs plays a critical role in addressing the issue of error accumulation, which does not simply provide a rough estimation of the next step, but carries underlying physics information that guides the longer-term prediction.

\subsection*{PPNN as a general framework for embedding known physics}
\label{sec:unetViT}
In the previous sections, we have demonstrated the performance enhancement achieved by PPNN based on ConvResNet architecture. However, the proposed approach is not restricted to a particular neural network structures. In this section, we showcase the flexibility of PPNN as a general framework by integrating the PDE-preserving part into various DNN architectures, specifically U-Net and Vision Transformer (ViT). More details on the particular U-Net and ViT architectures employed in our study are provided in the supplementary information. To illustrate the versatility of PPNN, we tested it alongside its corresponding baseline in the context of the viscous Burgers' equation, as discussed in Section \nameref{sec:burgers}.

\begin{figure}[!htp]
    \centering
    \includegraphics[width=0.9\textwidth]{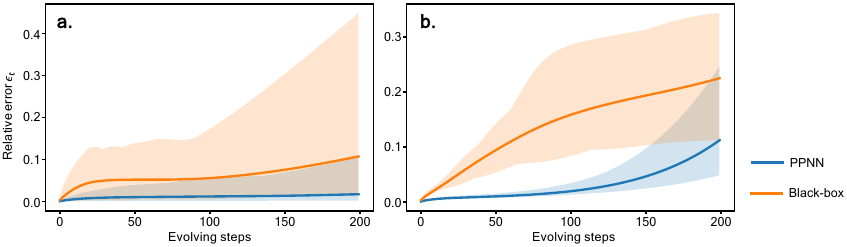}
    \caption{Prediction comparison when  partial differential equation preserved neural network (PPNN) using different deep neural networks as the trainable part. The relative error $\epsilon_t$ of 100 randomly sampled testing parameters $\bm{\lambda}$ is showed in this figure. The solid lines shows the averaged error over these 100 samples with shaded area shows the maximum and minimum relative errors of all testing trajectories. Blue (\blueline) indicates the error of PPNN while orange (\orangeline) represents the corresponding black-box method. \textbf{a}. shows the relative error of ViT and its PPNN variant. \textbf{b}. shows the relative error of U-Net and its PPNN counterpart.}
    \label{fig:unetViT}
\end{figure}
Figure~\ref{fig:unetViT} presents the relative error $\epsilon_t$ gathered from 100 randomly selected unseen input parameters $\bm{\lambda}$ over 200 testing time steps for both the U-Net and ViT scenarios. In both cases, the PPNN variant considerably outperforms its black-box counterpart, achieving much lower relative errors. Furthermore, the error distributions of the PPNN exhibit narrower ranges in comparison to those of the baseline models. When compared to the ConvResNet (see Figure~\ref{fig:rd_bg}\textbf{f}), both the U-Net and ViT baselines exhibit significantly enhanced performance in terms of the average relative error. It is noteworthy that while the baselines exhibit improved performance, the PPNN variant demonstrates does not have the same performance gain, albeit still superior to their corresponding baseline models. This observation suggests a potential overfitting issue in the PPNN variant, warranting further investigation.

By successfully incorporating PPNN with a variety of DNN architectures and exhibiting its superior performance in the setting of the viscous Burgers' equation, we furnish compelling evidence that PPNN operates as a flexible framework for integrating known physics into deep neural networks. This underlines its potential for enhancing the predictive accuracy and robustness across various neural architectures.Moreover, our approach not only demonstrates compatibility with different neural networks but also shows impressive generalizability across varying boundary conditions. For additional insights into the application of PPNN on diverse boundary value problems, we invite readers to refer to the Section \ref{sec:bc} in the supplementary information.

\subsection*{Comparison with existing SOTA methods for neural operator learning}
The backbone of the proposed PPNN method is a next-step auto-regressive model, which learns the transition dynamics of a spatiotemporal process, by mapping the solution fields from previous time steps to the next ones, and the whole trajectory prediction is obtained by rolling out the learned transition model autoregressively. Since the PPNN prediction is also conditioned on parameters $\boldsymbol{\lambda}$, the proposed model can be interpreted as learning an operator $\mathcal{G}$ in a discrete manner,
\begin{equation}
    \mathcal{G}: \bm{u}(\bm{\xi};\bm{\lambda}) \mapsto \mathcal{G}\large[\bm{u}(\bm{\xi};\bm{\lambda})\large],
    \label{eq:operator}
\end{equation}
where $\bm{\xi} = [\bm{x}, t]$ represents spatial and temporal coordinates and $\bm{u}\in\mathbb{R}^n$ is the $n$-dimensional state variable. In addition to the auto-regressive formulation, one can directly learn the operator $\mathcal{G}$ using deep neural networks in a continuous manner, generally referred to as neural operators. In the past few years, several continuous neural operator learning methods have been proposed, e.g., DeepONet~\cite{lu2021learning, lu2022comprehensive} and Fourier Neural Operator (FNO)~\cite{li2020fourier}. Although many of them have shown great success for a handful of PDE-governed systems, it remains unclear how these methods perform compared to our proposed PPNN on the challenging scenarios studied in this work,
\begin{itemize}
    \item Problems with high-dimensional parameter space, i.e., $\bm{\lambda} \in \mathbb{R}^d, d\gg1$.
    \item Limited training data for good generalizability in parameter space and temporal domain. 
\end{itemize}
Therefore, we conduct a comprehensive comparison of PPNN with existing state-of-the-art (SOTA) neural operators, including physics-informed neural network (PINN)~\cite{raissi2019physics}, DeepONet~\cite{lu2021learning, lu2022comprehensive}, 
and Fourier Neural Operator (FNO)~\cite{li2020fourier}, on one of the previous test cases, \nameref{sec:burgers}, where the PDE is fully known. (Strictly speaking, original PINN by Rassi et al.~\cite{raissi2019physics} is not an operator learner, but can be easily extended to achieving so by augmenting the network input layer with the parameter dimension, as shown in~\cite{sun2020surrogate}.) For a fair comparison, the problem setting and training data remain the same for all the methods and the number of trainable parameters of each models are comparable (PINN: 1.94M parameters; DeepONet: 1.51M parameters; PPNN: 1.56M parameters. Please note in DeepONet, we used two separate but identical neural networks to learn the two components $u_x$, $u_y$ of velocity $\bm{u}$ respectively to achieve optimal performance; each network contains 0.755M trainable parameters). Except for FNO, which has 0.58M trainable parameters due to the spatial Fourier transformation in FNO is too memory-hungry for a larger model to fit into the GPU used for training (RTX A6000 48GB RAM). It is worth noting that FNO could be formulated either as a continuous operator or as an autoregressive model. Here we show the performance of the continuous FNO. The performance of autoregressive FNO (named as aFNO) is shown in the Section~\ref{sec:fno_more} supplementary information, which is slightly better compared to continuous FNO in terms of testing error with unseen parameters. Besides, we also include a DeepONet with significantly more trainable parameters (79.19M) to show the highest possible performance DeepONet would achieve, which is named as DeepONet-L. Note that since some of these models' original forms cannot be directly applied to learn parametric spatiotemporal dynamics in multi-variable settings, necessary modifications and improvement has been made. The implementation details and hyper-parameters of these models are provided in the supplementary information (see \ref{sec:nn}). 

\begin{figure}[t!]
    \centering
    \includegraphics[width=\textwidth]{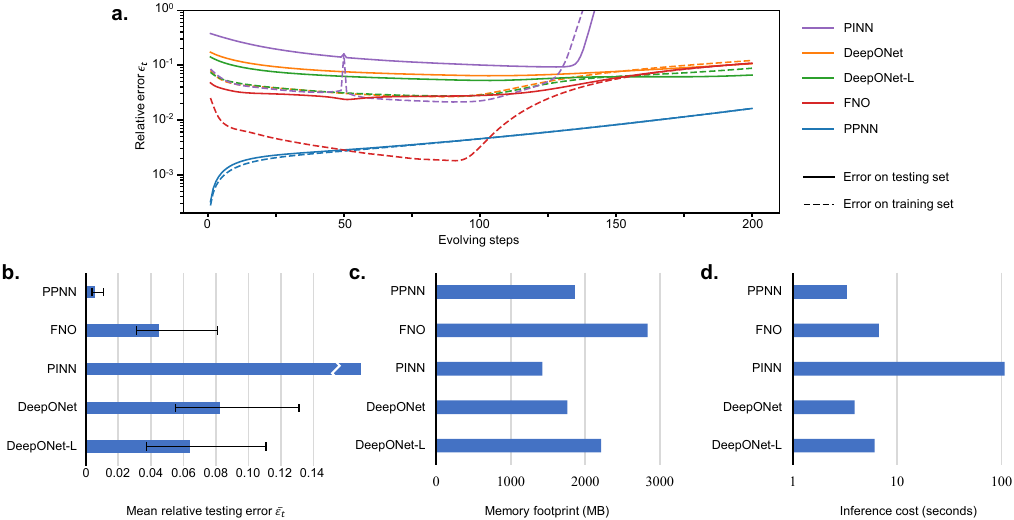}
    \caption{Comparison between  partial differential equation preserved neural network (PPNN) and various neural operators. \textbf{a}. Comparison of the relative error $\epsilon_t$ of the physics-informed neural network (PINN, purple lines \purpleline),  deep operator networks (DeepONet, orange lines \orangeline), DeepONet-L (green lines \greenline), Fourier neural operator (FNO, red lines \redline), and PPNN (blue lines \blueline) of the velocity $\bm{u}$ in the viscous Burgers' equation evaluated on 100 randomly generated testing (unseen) parameters (solid lines), and 100 randomly selected training parameters (dashed lines). Only the first $100$ time steps are used for training. Note that the y axis is in log scale. \textbf{b}. Relative testing error averaged over all testing parameters and time steps $\bar{\epsilon_t}$ with error bar. Note that PINN has a much higher error than the other models which cannot be completely shown in this figure. The error bar here indicates the lowest and highest relative testing error $\epsilon_t$ among all the testing parameters, while the blue bar shows the mean relative testing error $\bar{\epsilon_t}$. \textbf{c}. The memory footprint of different methods when testing 10 trajectories. Please note that, the time cost and memory footprint measured of PINN is the amount required for inferring a single row of the target mesh at a time. Inferring the whole field requires more memory, which exceeds the inference device's capacity. \textbf{d}. The inference cost of testing 10 trajectories on a NVIDIA RTX 3070 GPU. Please note the inference time of PINN is measured for the inference-optimized variants of the original models, which are significantly faster than the original form used for training.}
    \label{fig:baseline-compare}
\end{figure}

\paragraph{Predictive performance comparison}
All the models are used to predict the spatiotemporal dynamics of 100 randomly generated initial fields which are not seen in training. The relative prediction errors $\epsilon_t$ of the existing SOTA neural operators and PPNN are compared in Fig.~\ref{fig:baseline-compare}. As shown in Fig.~\ref{fig:baseline-compare}a and~\ref{fig:baseline-compare}b, PPNN significantly outperforms all the other SOTA baselines for all the time steps in both training and testing parameter regimes. All the existing SOTA neural operators have much higher prediction errors (several orders of magnitude higher) compared to PPNN, especially when entering the extrapolation range (after 100 time steps), where the error grows rapidly. In contrast, the relative error of PPNN predictions remains very low ($10^{-3}$) and barely accumulated evolving with time (shown in Fig~\ref{fig:baseline-compare}a). The prediction errors of most continuous neural operators do not grow monotonically since their predictions do not rely on auto-regressive model rollout and thus does not have error accumulation issue. However, the overall accuracy of all continuous neural operators (particularly in extrapolation range) is much lower than that of PPNN. Besides, PPNN exhibits a much smaller error scattering over different testing samples (shown in Fig.~\ref{fig:baseline-compare}b), indicating significantly higher robustness compared to existing SOTA methods. All of these observations suggest the obvious superiority of the PPNN in terms of extrapolability in time.  

The comparison of the generalizability in parameter space of all methods is shown in Fig.~\ref{fig:baseline-compare}a, where the dashed lines represent the averaged testing errors on the training parameter set, while the solid lines indicate the errors on the testing parameter set. It is clear that all the continuous neural operators have a significantly higher prediction error on the testing set than that on the training set, while the PPNN's prediction errors are almost the same on both the testing and training sets, which are much lower than all the other methods, indicating a much better generalizability. 

It is worth mentioning that the notable overfitting issue is observed in DeepONet with increased trainable parameters, i.e., DeepONet-L. It can be seen that although the prediction errors of DeepONet-L and FNO are relatively lower on training parameter sets and interpolation regimes, they rapidly increase when stepping into extrapolation ranges and unseen parameter regimes. We would like to point out that there are ``physics-informed'' variants of DeepONet and FNO~\cite{wang2021learning, li2021physics}, which regulate the DNN training by minimizing the residual of governing PDEs, in conjunction with data loss. However, these approaches typically necessitate the knowledge of the complete equation forms to formulate the physics-informed loss, while our method excel at integrating partially known physics, such as individual PDE operators, into the neural network structures. Moreover, the challenge of balancing DNN training with equation loss and label data is a well-documented issue, often requiring sophisticated hyperparameter tunning to adjust the weights between equation loss and data loss~\cite{wang2022and, chen2018gradnorm, mcclenny2020self}. In addition, the use of equation loss for problems with a high-dimensional parameter space poses a significant challenge in minimizing the composed loss function, leading to marginal and often unstable improvement over purely data-driven methods. We provide a more detailed comparison and discussion regarding the performance of these physics-informed variants of FNO/DeepONet in the supplementary information (see \ref{sec:pideeponet}).

\paragraph{Cost comparison}
Figures~\ref{fig:baseline-compare}c and~\ref{fig:baseline-compare}d show the time cost and memory footprint in the inference phase. Even compared to the fastest baseline DeepONet, PPNN is still about $20\%$ faster and the memory footprint of PPNN is very close to the model with the smallest memory footprint: DeepONet. It should be noted that all the models, including our PPNN are not exhaustively fine-tuned. Although carefully tuning hyperparameters may further improve the performance of each model, issues such as generalizability or robustness presented above cannot be addressed by hyperparameter tuning.    

\subsection*{Limitations and future directions}
Despite the significant advancements of the PPNN model, it is important to acknowledge some inherent limitations and potential areas of improvement. One such limitation is the minor accumulation of error when extrapolating over a significantly long duration. This can be attributed to the one-step prediction formulation currently utilized by the PPNN model. It is worth noting that the trainable aspect of PPNN is mathematically equivalent to learning closure models under the current design. However, a crucial distinction between our proposed method and existing closure model approaches lies in the ability to propagate gradients through PPNN, encompassing both the PDE-preserving segment and the trainable portion. This capability arises from our representation of physics priors through convolutional neural networks, making it differentiable. This differentiable feature empowers us to conduct end-to-end training with long-term model rollouts, an achievement unattainable in traditional closure model learning, where the physics priors (i.e., numerical solvers) lack differentiability, often necessitating direct labels for the discrepancy of such prior models. More importantly, this study underscores the connection between numerical PDE operators and neural architecture components, which can pave the way for innovative neural solver designs that go beyond classic PDE solvers augmented with DNN closures. From this viewpoint, conventional numerical PDE solvers can be conceptualized as a specific instance of neural networks. Their architecture details, including elements like convolution kernels, residual connections, or recurrent structure, are completely determined by the governing PDEs and their associated numerical schemes. In contrast, fully trainable neural networks are completely flexible and derive their structural parameters purely from data. Nonetheless, it is important to note that in this work, we did not fully explore the potential of PPNN through extended model rollouts during training. Conventional time-marching solvers operate based on predefined time integration schemes. In contrast, the PPNN framework offers the potential to weave various numerical PDE operators and trainable components to construct a modern DL architecture such as LSTM or transformer, clearly a departure from standard PDE solvers. This paper represents our initial step into the field of differentiable hybrid neural modeling, primarily aiming to explore and demonstrate the merit of PDE-integrated neural models. As such, the design and comparison of various hybrid PDE-neural architectures fall outside the scope of this work.

Our present work primarily focuses on structured data or meshes, a choice driven by their simplicity and computational efficiency. These are commonly used in many computational physics problems with relatively simple geometries. The novelty of our approach lies in the innovative use of structured meshes to design a PDE-preserving neural network. This is accomplished by mapping known PDE operators onto convolutional filters, thereby transposing the laws of physics into the language of deep learning. However, this focus on structured data does not mean that our model is inherently limited to such data. We see our demonstration on structured meshes as an essential first step towards extending the approach to more complex geometries and unstructured data. Addressing concerns about the applicability of our method to unstructured data and irregular geometries, we note that our current method could be extended using graph neural networks. The graph convolution operation, interpreted as a localized spectral filtering on unstructured data, can be viewed as a generalization of the CNN's convolution operation. By carefully designing spectral filters, the concept of ``PDE-preserving'' can be incorporated into desired spatial PDE operators through finite-volume-based or finite-element-based kernel functions. Although such an extension would require rigorous mathematical derivations and extensive empirical studies, we believe it serves as an intriguing direction for future research.

Furthermore, the ConvResNet formulation in the current version of PPNN is not mesh-invariant due to the discrete convolution operation, suggesting it cannot directly process data represented on different meshes without interpolations. However, the proposed PPNN framework can be extended to accommodate mesh invariance. One potential way to achieve this is to use mesh-invariant convolutional layers, which apply the same operations to the input data regardless of the underlying mesh structure. This could be realized, for instance, by employing geodesic convolutions or graph convolution kernel in spectral domain, allowing the model to adapt to variations in the mesh resolutions. Additionally, integrating adaptive mesh refinement techniques into the training process might provide another route towards mesh invariance. This strategy would involve dynamically adjusting the mesh resolution by incorporating mesh info $\Delta x$ into the model, allowing the model to capture the mesh variations.

In real-world applications, training data can be gathered from experiments or in-situ sensing, where data uncertainty may arise due to measurement noises in both inputs and training labels. Our current PPNN model does not include an uncertainty quantification (UQ) capability, but uncertainty propagation and quantification represent fascinating directions for future research. Extending the PPNN model to incorporate Bayesian learning could be a potential solution. Techniques like Bayesian neural networks using variational inference~\cite{kucukelbir2017automatic,graves2011practical,hoffman2013stochastic,gal2016dropout} or deep ensemble methods~\cite{lakshminarayanan2017simple,ovadia2019can,rahaman2021uncertainty} may offer promising avenues for expanding the PPNN model to include UQ capabilities.

Spatiotemporal dynamics constitute a fundamental aspect of numerous physics systems, ranging from classical fields like fluid dynamics, acoustics, and electromagnetics to the intricate realm of Quantum mechanics. The governing equations for such dynamics often fall within the domain of partial differential equations. Consequently, the ability to effectively solve these PDEs is imperative for comprehending, modeling, and controlling the underlying physical processes. By integrating the PDE structure into deep neural networks, PPNN represents a powerful tool for modeling such PDEs. In the context of various physics applications, PPNN exhibits considerable potential. Contrasted with traditional numerical solvers or earlier physics-informed neural networks, PPNN demonstrates lower training and inferring cost and the capacity to assimilate unknown physics from data. Additionally, when compared to purely data-driven methods, PPNN provides enhanced accuracy in out-of-sample scenarios while maintaining stability over prolonged model rollouts. The versatile nature of PPNN makes it a promising candidate for applications in modeling and predicting dynamic physics systems, including heat transfer, turbulent flow, and electromagnetic fields. While not delving into the specifics of each application, it is evident that PPNN holds significant promise for speeding up the study and understanding of complex spatiotemporal dynamics across various physics domains.

\section*{Conclusion}
In this work, we proposed a physics-inspired deep learning framework, PDE-preserved neural network (PPNN), aiming to learn parametric spatiotemporal physics, where the (partially) known governing PDE structures are preserved via fixed convolutional residual connection blocks in a multi-resolution setting. The PDE-preserving ConvResNet blocks together with trainable blocks in an encoding-decoding manner bring the PPNN significant advantages in long-term model rollout accuracy, spatiotemporal/parameter generalizability, and training efficiency. The effectiveness and merit have been demonstrated over a handful of challenging spatiotemporal prediction tasks, including the FitzHugh–Nagumo reaction diffusion equations, viscous Burgers equations and Naiver-Stokes equations, compared to the existing baselines, including ConvResNet, U-Net, Vision transformer, PINN, DeepONet, and FNO. The proposed PPNN shows satisfactory predictive accuracy in testing regimes and significantly lower error-accumulation effect for long-term model rollout in time, even if the preserved physics is incomplete or inaccurate. 
Finally, the discussion on the inference and training costs shows the great potential of the proposed model to serve as a reliable and efficient surrogate model for spatiotemporal dynamics in many applications that require repeated model queries, e.g., design optimization, data assimilation, uncertainty quantification, and inverse problems. While Direct Numerical Simulations (DNS) are used as the source of labeled training data in our study, the data could just as well originate from experimental results or field observations. A unique feature of PPNN, and one of its significant advances, lies in its ability to generalize to different physical parameters and initial/boundary conditions. Unlike most label-free PINN techniques, which act as PDE solvers for a given set of parameters and conditions, PPNN's ability to adapt to varying parameters and conditions underscores its capability to learn the PDE system. In general, this work explored a creative design of leveraging physics-inductive bias in scientific machine/deep learning and showcased how to use physical prior knowledge to inform the learning architecture design, shedding new light on physics-informed deep learning from a different aspect. Therefore, this work represents a inventive PiDL development and a significant advance in the realm of SciML.

\section*{METHODS}
\label{sec:methodology}
\subsection*{Problem formulation}
We are interested in predictive modeling of physical systems with spatiotemporal dynamics, which can be described by a set of parametric coupled PDEs in the general form,
\begin{linenomath*}
\begin{subequations}\label{eq:pde}
	\begin{alignat}{3}
	\frac{\partial \bm{u}}{\partial t} + \mathscr{F} \bigg[\bm{u}, \bm{u}^2, ..., \nabla_{\bm{x}}\bm{u}, \nabla_{\bm{x}}^2\bm{u}, \nabla_{\bm{x}}\bm{u}\cdot\bm{u}, ...; \boldsymbol{\lambda} \bigg] &= \bm{0}, \qquad & &\bm{x}, t \in \Omega \times [0, T], \boldsymbol{\lambda} \in \mathbb{R}^{d},\\ 
	\mathcal{I}\bigg[\bm{x}, \bm{u}, \nabla_{\bm{x}}^2\bm{u}, \nabla_{\bm{x}}\bm{u}\cdot\bm{u}; \boldsymbol{\lambda}\bigg] &= \bm{0}, \qquad & &\bm{x} \in \Omega, t =0,\boldsymbol{\lambda} \in \mathbb{R}^{d},\\
	\mathcal{B}\bigg[t, \bm{x}, \bm{u}, \nabla_{\bm{x}}^2\bm{u}, \nabla_{\bm{x}}\bm{u}\cdot\bm{u}; \boldsymbol{\lambda}\bigg] &= \bm{0}, \qquad & &\bm{x}, t \in \partial\Omega \times [0, T], \boldsymbol{\lambda} \in \mathbb{R}^{d},
	\end{alignat}
\end{subequations}
\end{linenomath*}
where $\bm{u} = \bm{u}(\bm{x}, t; \boldsymbol{\lambda}) \in\mathbb{R}^{n}$ is the $n$-dimensional state variable; $t$ denotes time and $\bm{x}\in\Omega$ specifies the space; $\mathscr{F}[\cdot]$ is a complex nonlinear functional governing the physics, while differential operators $\mathcal{I}[\cdot]$ and $\mathcal{B}[\cdot]$ describe the initial and boundary conditions (I/BCs) of the system, respectively; $\boldsymbol{\lambda}\in\mathbb{R}^d$ is a $d$-dimensional vector, representing physical/modeling parameters in the governing PDEs and/or I/BCs. Solving this parametric spatiotemporal PDE system typically relies on traditional FD/FV/FE methods, which are computationally expensive in most cases. This is due to the spatiotemporal discretization of the PDEs into a high-dimensional algebraic system, making the numerical simulation time-consuming, particularly considering that a tiny step is often required for the time integration to satisfy the numerical stability constraint. Moreover, as the system solution $\bm{u}(\bm{x}, t; \boldsymbol{\lambda})$ is parameter-dependent, we have to start over and conduct the entire simulation given a new parameter $\boldsymbol{\lambda}$, making it infeasible for application scenarios that require many model queries, e.g., parameter inference, optimization, and uncertainty quantification. Therefore, our objective is to develop a data-driven neural solver for rapid spatiotemporal prediction, enabled by efficient time-stepping with coarse-gaining and fast inference speed of neural networks. In particular, this study focuses on the learning architecture design by preserving known PDE structures for improving the robustness, stability, and generalizability of data-driven auto-regressive predicting models.  

\subsection*{Next-step prediction models based on convolutional ResNets}
\label{sec:methd2}
The next-step DNN predictors are commonly used for emulating spatiotemporal dynamics in an autoregressive manner,
\begin{linenomath*}
\begin{equation}
    \bm{u}_t = f_{\theta}(\bm{u}_{t-1}, \bm{\lambda}\mid\bm{\theta}),
\end{equation}
\end{linenomath*}
where the state solution $\bm{u}_t$ at time step $t$ is approximated by a neural network function $f_{\theta}: \mathbb{R}^n \times \mathbb{R}^d \to \mathbb{R}^n$, taking the previous state $\bm{u}_{t-1}$ and physical parameters $\bm{\lambda}$ as the input features. The function $f_{\theta}(\cdot\mid\bm{\theta})$ is parameterized by trainable weight vector $\bm{\theta}$ that can be optimized based on training labels. Once the model is fully trained, it can be used to predict spatiotemporal dynamics by autoregressive model rollouts given only the initial condition $\bm{u}_0$ and a specific set of physical parameters $\bm{\lambda}$. In general, the next-step model is built based on residual network (ResNet) blocks, which have recently improved the state-of-the-art (SOTA) in many benchmarks learning tasks~\cite{pfaff2020learning}. Given the input features $\bm{z}_0$, a ResNet block with $N$ layers outputs $\bm{z}_N$ as,
\begin{linenomath*}
\begin{equation}\label{eq:resNet}
    \bm{z}_{j+1} = \bm{z}_j + f^{(j)}(\bm{z}_j\mid\bm{\theta}^{(j)}), \hspace{1.5em} j = 0, \cdots N-1,
\end{equation}
\end{linenomath*}
where $f^{(j)}$ represents the generic neural network function of $j^\mathrm{th}$ layer and $\bm{\theta}^{(j)}$ are corresponding weights. For end-to-end spatiotemporal learning, $f^{(j)}$ are often formulated by (graph) convolutional neural networks with trainable convolution stencils and biases. In a ResNet block, the dimension of the feature vectors (i.e., the image resolution and the number of channels) should remain the same across all layers. The ResNet-based next-step models have been demonstrated powerful and effective for predicting complex spatiotemporal physics. One of the examples is the MeshGraphNet~\cite{pfaff2020learning}, which is a GNN-based ResNets and shows the SOTA performance in spatiotemporal learning with unstructured mesh data. 

\begin{figure}[!t]
    \centering
    \includegraphics[width=1.0\textwidth]{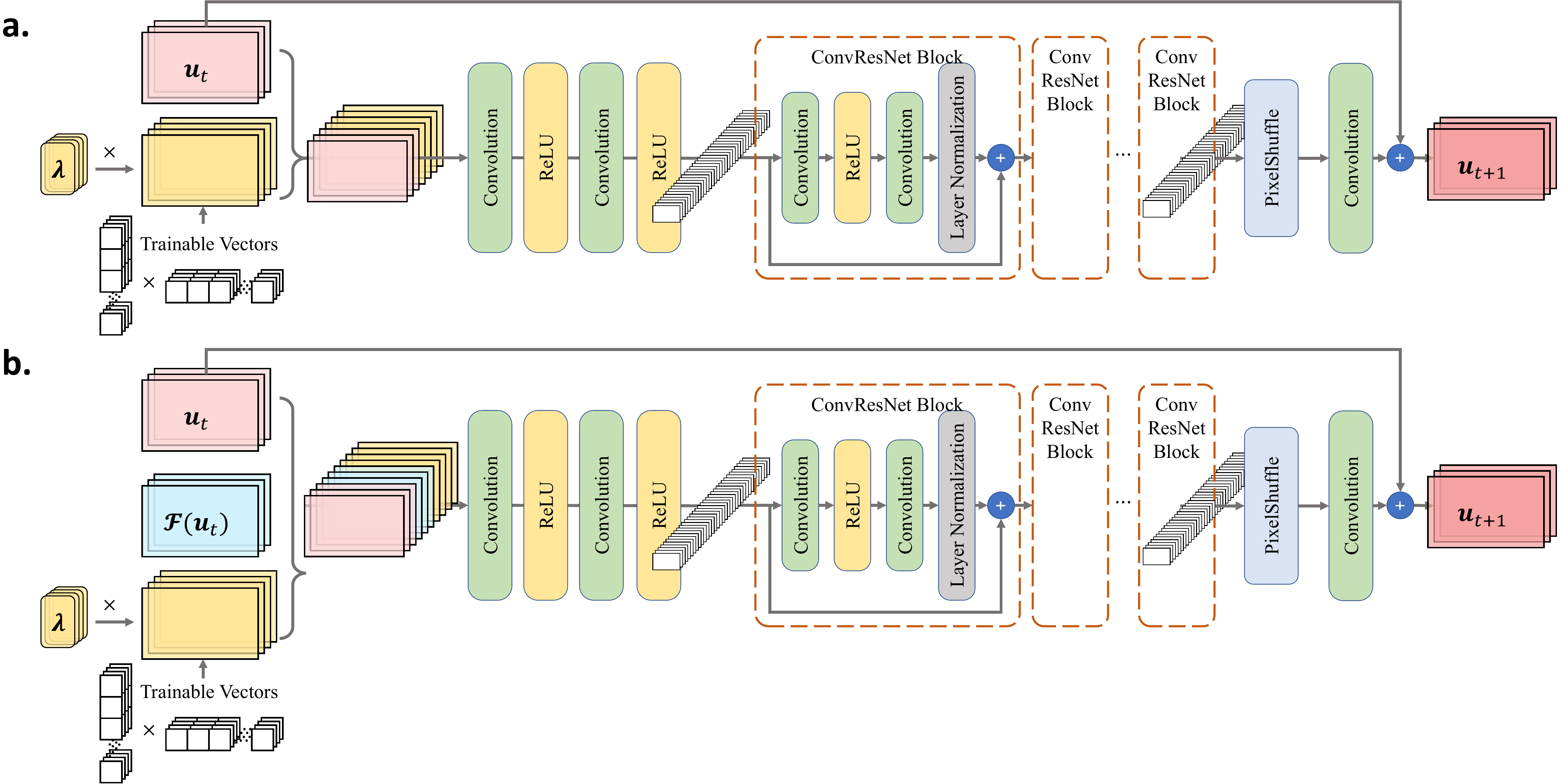}
    \caption{Schematics of the deep neural network structures used in this work. \textbf{a}, Network architecture of the baseline black-box ConvResNet-based next-step model. \textbf{b}, Network architecture of the trainable portion of  partial differential equation (PDE)-preserved neural network (PPNN). The only difference between them is trainable portion of PPNN has an extra input variable $\mathcal{F}(\bm{u}_t)$, provided by the PDE-preserving portion of PPNN.}
    \label{fig:NN}
\end{figure}

In this work, as we limit ourselves to structured data within regular domains, a CNN variant of the MeshGraphNet, Convolutional ResNet (ConvResNet)-based next-step model, is used as one of the baseline black-box models in this work, whose network structure is shown in Fig.~\ref{fig:NN}a. The ConvResNet takes the previous state and physical parameters as the input and predicts the next-step state using a residual connection across the entire hidden ConvResNet layers after a pixel shuffle layer. The hidden layers consist of several ConvResNet blocks, constructed based on standard convolution layers with residual connections and ReLU activation functions, followed by layer normalization. To learn the dependence of physical parameters $\bm{\lambda}$, each scalar component of the physical parameter vector is multiplied by a trainable matrix, which is obtained by vector multiplication of trainable weight vectors.

\subsection*{Neural network architecture and differential equations}
Recent studies have shown the relationship between DNN architectures and differential equations: ResNets can be interpreted as discretized forms of ODEs/PDEs, while differential equations can be treated as a continuous interpretation of ResNet blocks with infinite depth.

\paragraph{Residual connections and ODEs} 
As discussed in~\cite{haber2017stable,chen2018neural}, the residual connection as defined in Eq.~\ref{eq:resNet} can be seen as a forward Euler discretization of a ODE,
\begin{linenomath*}
\begin{equation}\label{eq:resNetC}
    \frac{\partial \bm{z}(t)}{\partial t} = F(\bm{z}(t) \mid\bm{\theta}(t)), \hspace{1.5em} \mathrm{for} \hspace{1em} t\in(0, T],
\end{equation}
\end{linenomath*}
where $\bm{z}(t=0) = \bm{z}_0$ and $T$ is total time. In ResNets, a fixed time step size of $\Delta t = 1$ is set for the entire time span and $N\cdot\Delta t = T$. Namely, the depth of the residual connection (i.e., the number of layers in a ResNet block) can be controlled by changing the total time $T$. On the other hand, an ODE as given by Eq.~\ref{eq:resNetC} can be interpreted as a continuous ResNet block with infinite number of layers (i.e., infinite depth). Based on this observation, the classic ResNet structure can be extended by discretizing an ODE using different time-stepping schemes (e.g., Euler, Runge-Kutta, leapfrog, etc.). Moreover, we can also define a residual connection block by directly coupling a differentiable ODE solver with a multi-layer perception (MLP) representing $F(\cdot)$, where the hybrid ODE-MLP is trained as a whole differentiable program using back-propagation, which is known as a neural-ODE block~\cite{chen2018neural}.   

\paragraph{Convolution operations and PDEs} In the neuralODE, MLP is used to define $F(\cdot)$, which, however, can be any neural network structure in a general setting. When dealing with structured data (e.g., images, videos, physical fields), the features $\bm{z}(t,\bm{x})$ can be seen as spatial fields, and convolution operations are often used to construct a CNN-based $F(\cdot)$. A profound relationship between convolutions and differentiations has been presented in~\cite{dong2017image,lu2018beyond,long2018pde}. Following the deep learning convention, a 2D convolution is defined as,
\begin{linenomath*}
\begin{equation}\label{eq:conv}
    \textrm{conv}\left(\bm{z}, h^{(\theta)}\right) = \int{\bm{z(\bm{x}'-\bm{x})} h^{(\theta)}(\bm{x})}d\bm{x},
\end{equation}
\end{linenomath*}
where $h$ represent convolution kernel parameterized by $\bm{\theta}$. Based on the order of sum rules, the kernel $h$ can be designed to approximate any differential operator with prescribed order of accuracy~\cite{long2019pde}, and thus the convolution in Eq.~\ref{eq:conv} can be expressed as~\cite{eliasof2021pde},
\begin{linenomath*}
\begin{equation}\label{eq:convdiff}
    \textrm{conv}\left(\bm{z}, h^{(\theta)}\right) = \mathscr{D} \bigg[\bm{u}, ..., \nabla_{\bm{x}}\bm{u}, \nabla_{\bm{x}}^2\bm{u}, \nabla_{\bm{x}}\bm{u}\cdot\bm{u}, ...; \boldsymbol{\theta} \bigg],
\end{equation}
\end{linenomath*}
where $\mathscr{D}$ is a discrete differential operator based FD/FV/FE methods. For example, from the point of view of FDM, convolution filters can be seen as the finite difference stencils of certain can be interpreted as the discrete forms of certain PDEs, and thus the PDEs can be used to inform ConvResNet architecture design.

\subsection*{Multi-resolution PDE-preserved Neural Network (PPNN) architecture}
It is well known that auto-regressive models suffer from error accumulation, which is particularly severe for the next-step formulation. Although remedies such as using training noises~\cite{sanchez2020learning} or sequence models~\cite{han2022predicting} have been explored, the error accumulation issue cannot be easily mitigated, and the model usually fails to operate in a long-span rollout. Inspired by the relationship between network architectures and differential equations, we hypothesize that the performance of an auto-regressive ConvResNet model for spatiotemporal learning can be significantly improved if the network is constructed by preserving (partially) known governing physics (i.e., PDEs) of the spatiotemporal dynamics. Therefore, we propose a multi-resolution PDE-preserved neural network (PPNN) framework, where the discrete governing PDEs are preserved in residual connection blocks using grids with multiple resolutions.

As shown in Fig.~\ref{fig:PPNN}, the PPNN has the same backbone ResNet structure as the black-box next-step baseline model, where a residual connection is applied across the entire hidden ConvResNet layers. The hidden ConvResNet consists of two portions: PDE-preserving ConvRes layers and trainable ConvRes layers, coupled in an encoding-decoding manner. In the PDE-preserving portion, the ConvRes connection is constructed based on the convolution operators defined by the discrete differential operators of the governing PDEs using finite-difference stencils. The preserved-PDE ConvRes layers are operated on low-resolution grids by taking in the downsampled input solution fields using bi-linear algorithm and the output is upsampled back to the original resolution using bi-cubic algorithm, which improves the model rollout stability with large evolving steps, meanwhile reducing the cost overhead during the model inference. This structure resembles the multgrid method which significantly improves the speed and reduce the cost by solving PDEs on different mesh resolutions. The trainable portion takes the high-resolution solution fields, together with the output of the PDE preserving part, as the input and contains a few classic ConvResNet blocks. For a fair comparison, the network architecture of the trainable portion is exactly the same as that of the black-box ConvResNet baseline, except that the trainable portion of PPNN takes the output of the PDE-preserving portion (see Fig.~\ref{fig:NN}). The PDE-preserving part and trainable part are connected via bi-cubic up-sampling operation. Overall, the PDE-preserving part enhance the trainable part by (a) preserving a time integration scheme (b) providing input feature enrichment. An ablation study of these two components can be found in the section \ref{sec:ablation} in supplementary information. Note that a smaller time step $\Delta t'$ can be used within the PDE-preserving portion via inner-iteration to stabilize model rollout. In general, the combination of the two portions can be seen as a ConvResNet architecture that preserves the mathematical structure of the underlying physics behind the spatiotemporal dynamics to be modeled.

\section*{Data availability} 
All the used datasets in this study can be generated by the openly available Python scripts on GitHub at \url{https://github.com/jx-wang-s-group/ppnn} upon publication.

\section*{Code availability} 
All the source codes to reproduce the results in this study will be openly available on GitHub at \url{https://github.com/jx-wang-s-group/ppnn} upon publication.

\vspace{24pt}
\noindent\textbf{Acknowledgement:}
X.Y.L. and J.X.W. would like to acknowledge the funds from 
Office of Naval Research under award numbers N00014-23-1-2071 and National Science Foundation, under award numbers OAC-2047127. H.S. acknowledges the funds from the National Natural Science Foundation of China (No. 62276269) and the Beijing Natural Science Foundation (No. 1232009). L.L. was supported by the funds from U.S. Department of Energy under award number No. DE-SC0022953. We would like to express our sincere gratitude to the three anonymous reviewers and the editor for their valuable comments and suggestions, which contributed to enhancing the quality of this paper.\\

\noindent\textbf{Author contributions:} X.Y.L., H.S. and J.X.W. contributed to the ideation and design of the research; X.Y.L. and J.X.W. performed the research (implemented the model, conducted numerical experiments, analyzed the data, contributed materials/analysis tools); X.Y.L, M.Z. and L.L. contributed to the comparison study with other models; X.Y.L. and J.X.W. wrote the manuscript; H.S. and L.L. contributed to manuscript editing. \\

\noindent\textbf{Corresponding authors:} Jian-Xun Wang  (\url{jwang33@nd.edu}). \\ 

\noindent\textbf{Competing interests:}
The authors declare no competing interests. \\

\noindent\textbf{Supplementary information:}
The supplementary information is attached.
\clearpage

\beginsupplement

\section*{\textbf{Supplementary Information}: \\ Multi-resolution partial differential equations preserved learning framework for spatiotemporal dynamics}







\section{Generalizability over boundary value problems}
\label{sec:bc}
We applied the ConvResNet-based PPNN and its black-box counterpart to a one-dimensional environment governed by the diffusion equation:
\begin{equation}
    \frac{\partial u}{\partial t} = \frac{\partial^2 u}{\partial x^2}, \quad t\in[0,100], \quad x\in\Omega = [0,L]
\end{equation}
where $u$ denotes the diffusing material density, $t$ and $x$ representing the time and space coordinates, respectively. The simulation space $\Omega$ is defined as $\Omega=[0,L],\,L=16\pi$. The boundary conditions were randomly sampled from two types: Dirichlet and Neumann boundaries, as below
\begin{equation}
    \begin{cases}
    u(x, t) = \beta b, \quad c=0\\
    \displaystyle\frac{\partial u(x,t)}{\partial x}  = b, \quad c=1
\end{cases}x\in\partial\Omega
\end{equation} 
where $\beta=5$ is a scaling factor. We use $b$ as the boundary value parameter and $c$ to represent the boundary type. In this case, $c=0$ corresponds to Dirichlet boundary and $c=1$ indicates Neumann boundary. The sampling is based on the following distributions,
\begin{equation}
\begin{split}
P(c=0)&=P(c=1)=0.5\\ &b \sim U(-1,1)
\end{split}
\end{equation}
where $P(\cdot)$ represent probability density and $U(-1, 1)$ is uniform distribution over range $(-1,1)$. 
Meanwhile, the initial condition is sampled from a $256$ dimensional space:
\begin{equation}
    u(x, 0) = \sum_{i=1}^{16}\sum_{j=1}^{16} \kappa_{i,j} \sin({i\cdot x/L+ \varphi_{i,j}})
\end{equation} where $\kappa_{i,j}$ and $\varphi_{i,j}$ are random variables sampled from uniform distribution over $[-0.5,0.5)$ and $[0,1)$, respectively. The training and testing data are generated by a finite difference solver using forward Euler and second order of accuracy central difference scheme. The spatial domain is discretized into $128$ grid points while the numerical time step is $\delta t = 0.01$ and learning step size is $\Delta t = 50\delta t$. 
The boundary value problem studied here is of a high-dimensional parameter space consisting of $b, c$ and the initial condition space, denoted as $\bm{\lambda}$.
\begin{figure}[!htp]
    \centering
    \includegraphics[width=0.5\textwidth]{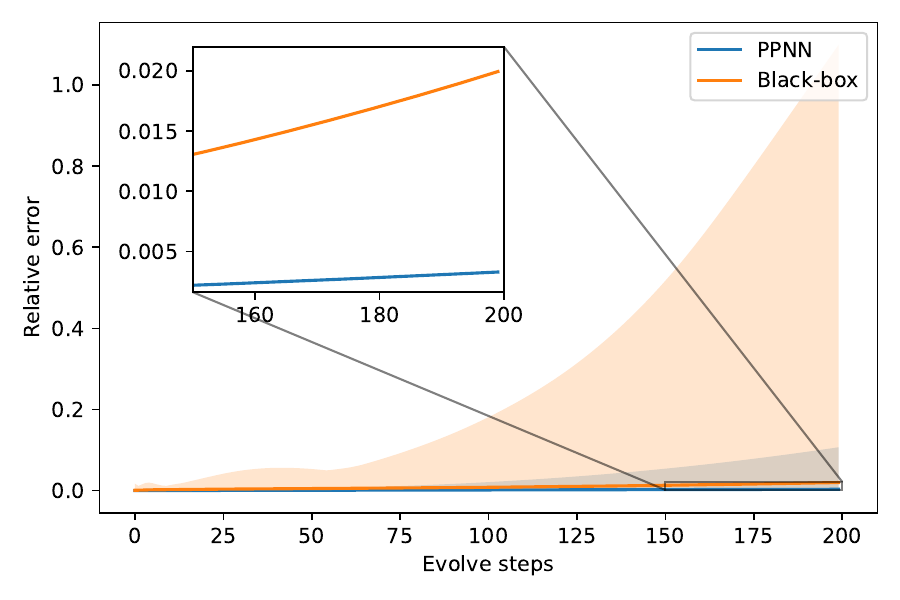}
    \caption{The error of 256 randomly selected parameters $\bm{\lambda}$, predicted by PPNN (blue line) compared to black-box baseline (orange line). Shaded areas indicate the error distribution range.}
    \label{fig:bverror}
\end{figure}
\begin{figure}[!h]
    \centering
    \includegraphics[width=\textwidth]{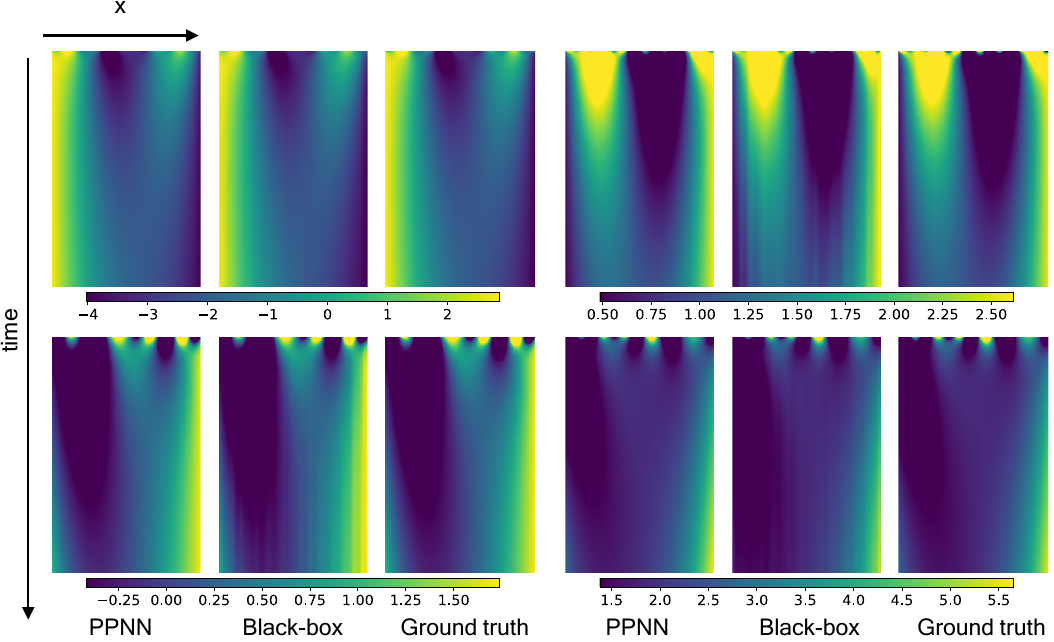}
    \caption{Prediction of $u$ of $4$ randomly selected $\bm{\lambda}$ by PPNN and black-box compared with the ground truth}
    \label{fig:bvsnap}
\end{figure}

Figure~\ref{fig:bverror} shows the relative error $\epsilon_t$ of PPNN compared with its corresponding black-box method tested with 256 randomly sampled unseen parameters $\bm{\lambda}$ for $200$ time steps. Similar to other cases, PPNN shows significantly lower relative error with an average error of merely $0.328\%$ at the last time step, which is almost one order of magnitude smaller than that of its black-box counterpart ($1.996\%$). Meanwhile, the error distribution range is significantly narrower.

Fig.~\ref{fig:bvsnap} compares the PPNN prediction against the black-box results and ground truth. We can see the PPNN results are visually identical to the ground truth while for the black-box method, while unphysical discontinuity can be observed especially close to the end of each trajectory.

\section{Ablation study of PPNN structure}
\label{sec:ablation}
We conducted ablation studies to understand the influence of the two components of the PDE-preserving portion: (a) the residual connection of the preserved PDE on coarse grids and (b) the feature expansion (input enrichment) derived from the PDE-preserving module for the trainable network. We çreated two variants of the ConvResNet-based PPNN: one only included component (a) and was referred to as PPNN-Residual only (PPNNRo), and the other, referred to as PPNN-Enrichment only (PPNNEo), exclusively utilized the output of the PDE-preserving part for feature enrichment. Burgers' equation is used for this study.
\begin{figure}[!htp]
    \centering
    \subfloat[$500$ epochs]{\includegraphics[width=0.48\textwidth]{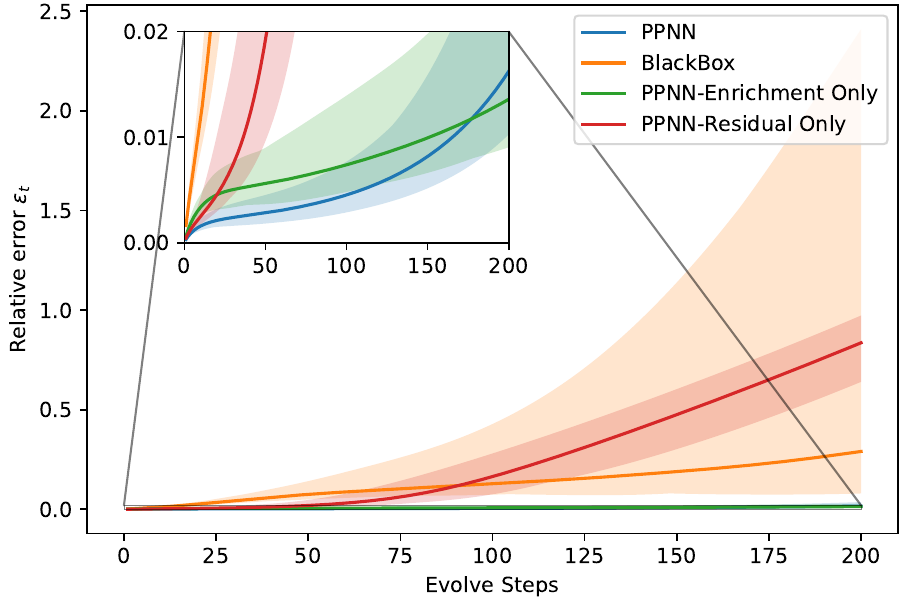}} 
    \subfloat[$50$ epochs]{\includegraphics[width=0.48\textwidth]{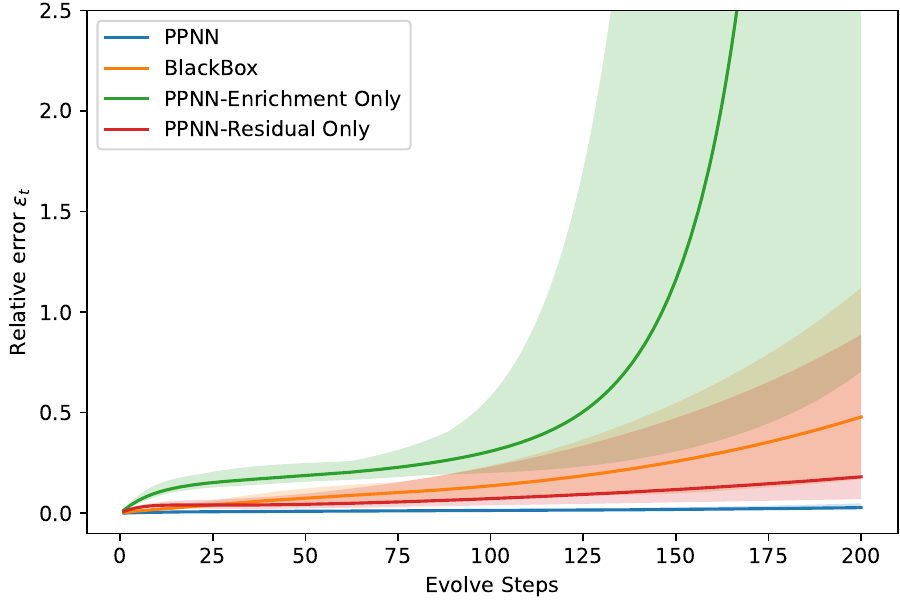}}
    \caption{Ablation study of the PPNN framework. Relative error $\epsilon_t$ of 100 randomly selected testing parameters $\bm{\lambda}$ in the viscous Burgers' case predicted by PPNN, black-box ConvResNet, PPNN with input enrichment only (PPNNEo), PPNN with PDE residual only (PPNNRo). (a) shows the relative error after sufficient training epochs ($500$ training epochs), while (b) shows the error with insufficient training epochs (i.e. $50$ epochs). solid lines indicate the average error of the $100$ testing parameters while the shaded areas represent the error distribution range.}
    \label{fig:noRP}
\end{figure}
As depicted in Fig.~\ref{fig:noRP}, the relative error $\epsilon_t$ of these two PPNN variants is compared with the complete PPNN and the black-box method, using unseen testing parameters. Fig.~\ref{fig:noRP}(a), illustrating the relative error after sufficient training (500 epochs), shows that while PPNNEo's performance closely mirrors that of the complete PPNN, the full PPNN nonetheless delivers the lowest relative error and the tightest error distribution range over the majority of testing steps. Meanwhile, PPNNRo struggles from significant error accumulation, suggesting that training becomes notoriously challenging when PDE residuals are connected but not included as input features in the trainable networks. Even so, PPNNRo exhibits a significantly narrower error distribution range than the black-box baseline, indicating the value of embedding PDEs for biasing the final results. The performance gap between PPNNRo and the full PPNN or PPNNEo could be narrowed by using a higher-resolution mesh or a higher-accuracy scheme in the PDE-preserving part, though this would come with increased inference time.

However, with insufficient training (50 epochs), as depicted in Fig.~\ref{fig:noRP}(b), PPNNEo experiences rapid error accumulation after rolling out for $100$ steps and exhibits a wide error distribution range—worse than that of the black-box baseline—indicating it is more challenging to train due to feature expansion. This can be substantially mitigated by adding the PDE-based residual connection, benefiting from the robust bias introduced by the PDE operators preserved on coarse grids. Throughout all stages, the complete PPNN retains the best performance.

The results indicate that both the feature enrichment and the residual connection elements are crucial to our model's overall performance. The residual connection in particular stabilizes training—especially in the early stages, which is particularly necessary when training in a sequence-to-sequence (Seq2Seq) style—and significantly reduces training cost, while input enrichment can eventually enhance accuracy at the final stage with sufficient training.

\section{Details of the PPNN and baseline methods}
\label{sec:nn}
In this section, we provide the details of the implementations of PPNN and the baseline methods that are omitted in the main text. 
Table.~\ref{tab:resolution} shows the resolution used by the PPNN in different cases.
\begin{table}[!h]
    \centering
    \caption{Resolution used in the PPNN}
    \begin{tabular}{c|c|c|c}
    \hline
        & Reaction-diffusion & viscous Burgers' & Navier-Stocks\\
        \hline
       Output resolution & $256\times256$ & $256\times256$ & $400\times100$\\
       \hline
       PDE-preserving part resolution & $48\times48$ & $32\times32$ & $100\times25$\\
       \hline
    \end{tabular}
    \label{tab:resolution}
\end{table}

To make a fair comparison, the neural networks we used share a very similar encoder structure, which leverages multiple convolution layers (except FNO which uses channel-wise MLP) to compress the multi-channel input field to a smaller field (for PPNN, black-box ConvResNet and FNO) or to a hidden vector (for PINN and DeepONet). The physical parameters are first mapped to a field with the same shape as the input field and then passed to the CNN encoder as an additional channel. Here we use two trainable vectors (except DeepONet) of shape $[n_x, 1]$ and $[1, n_y]$ (where $n_x$, $n_y$ are the number of evaluating points in the input field in $x$ and $y$ directions respectively) to generate a matrix and multiply with the physical parameters to mapping to a field.


\subsection{PPNN and Black-box ConvResNet}
The black-box ConvResNet baseline and the trainable portion of PPNN share an almost identical structure for every case, as shown in the Figure 10 in the main manuscript, respectively. In the RD equations cases and Burgers' equations cases, the encoder part consists of two layers of convolutional layers, using a $6\times6$ kernel, with zero padding of $2$ and a stride of $2$. Afterwards, there are three ConvResNet blocks, each with a $5\times5$ kernel, $48$ channels and a zero padding of $2$. In NS cases, four ConvResNet blocks are used, each of which has a $7\times7$ kernel, 96 channels and a zero padding of $3$. The following decoder includes a pixelshuffle with upscale factor equals $4$ and a convolution layer with a $5\times5$ kernel and zero padding of 2. The number of trainable parameters are shown in Table.~\ref{tab:infer_cost1}. All the neural networks are implemented on the deep learning platform PyTorch~\cite{NEURIPS2019_9015} and trained on a NVIDIA RTX A6000 GPU. 

\begin{table}[ht!]
\caption{Number of trainable parameters}
\centering
    \begin{threeparttable}
        \begin{tabular*}{\textwidth}{@{\extracolsep{\fill}}c|clcl}
            \toprule
            \multirow{2}{10em}{\,\,\,\quad\qquad Cases} & \multicolumn{3}{c}{Number of trainable parameters}\\
                & PPNN      && Black-box ConvResNet \\
            \midrule
            RD/Burgers & 1,549,156 && 1,548,292  \\
            NS         & 5,624,869 && 5,622,277  \\
            \bottomrule
        \end{tabular*}
    \end{threeparttable}
\label{tab:infer_cost1}
\end{table}

\subsection{U-Net and its PPNN counterpart}
\begin{figure}[!htp]
    \centering
    \includegraphics[width=\textwidth]{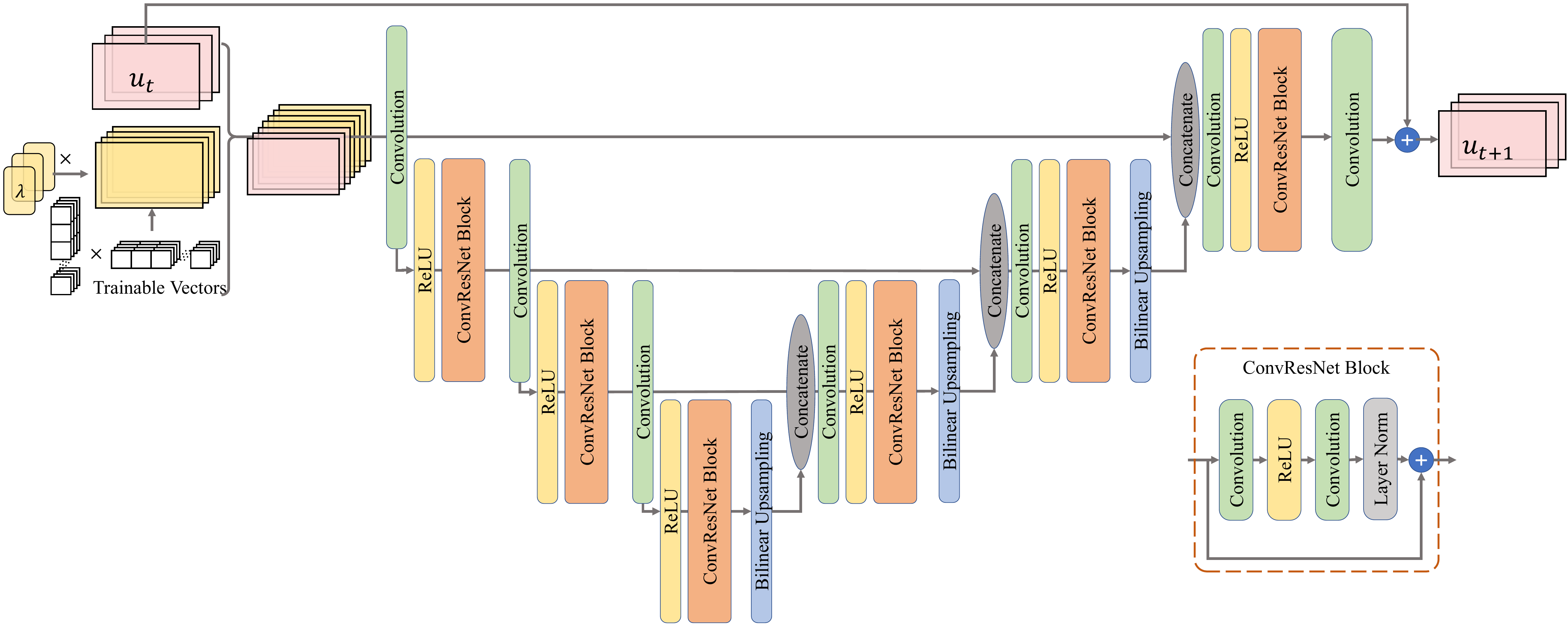}
    \caption{Schematic of the Unet used in this paper}
    \label{fig:unet}
\end{figure}

The structure of the U-Net used in section "PPNN as a General Framework for Embedding Known Physics" is shown in Fig.~\ref{fig:unet}. We used a three-layer encoder-decoder and in each  
The Black box baseline has $2,283,532$ trainable parameters while PPNN has $2,284,396$ trainable parameters. 

\subsection{Vision transformer and its PPNN counterpart}
For the vision transformer used in the section ``PPNN as a General Framework for Embedding Known Physics'', we divided the spatial field (of size $256\times 256$) into $8\times8$ patches and each patch is of size $32\times32$. A four dimensional position embedding $Pe$ is employed to indicate the position of each patch. 
\begin{equation}
    Pe = [\sin(x), \cos(x), sin(y), cos(y)]\quad x,y\in\{0,1,2,\cdots, 7\}
\end{equation}
The ViT consists of $6$ layers of attention encoder and a single layer attention decoder. In each multi-head attention layer, we use $16$ attention heads, with a hidden dimension of $1024$, while the output of the encoder is a $2048$ dimensional vector for each patch. The overall trainable parameters for black-box baseline is $64,025,088$ while the corresponding PPNN has 66,126,336 trainable parameters.

\subsection{PINN}
\begin{figure}[ht!]
    \centering
    \subfloat[]{\includegraphics[width=0.5\textwidth]{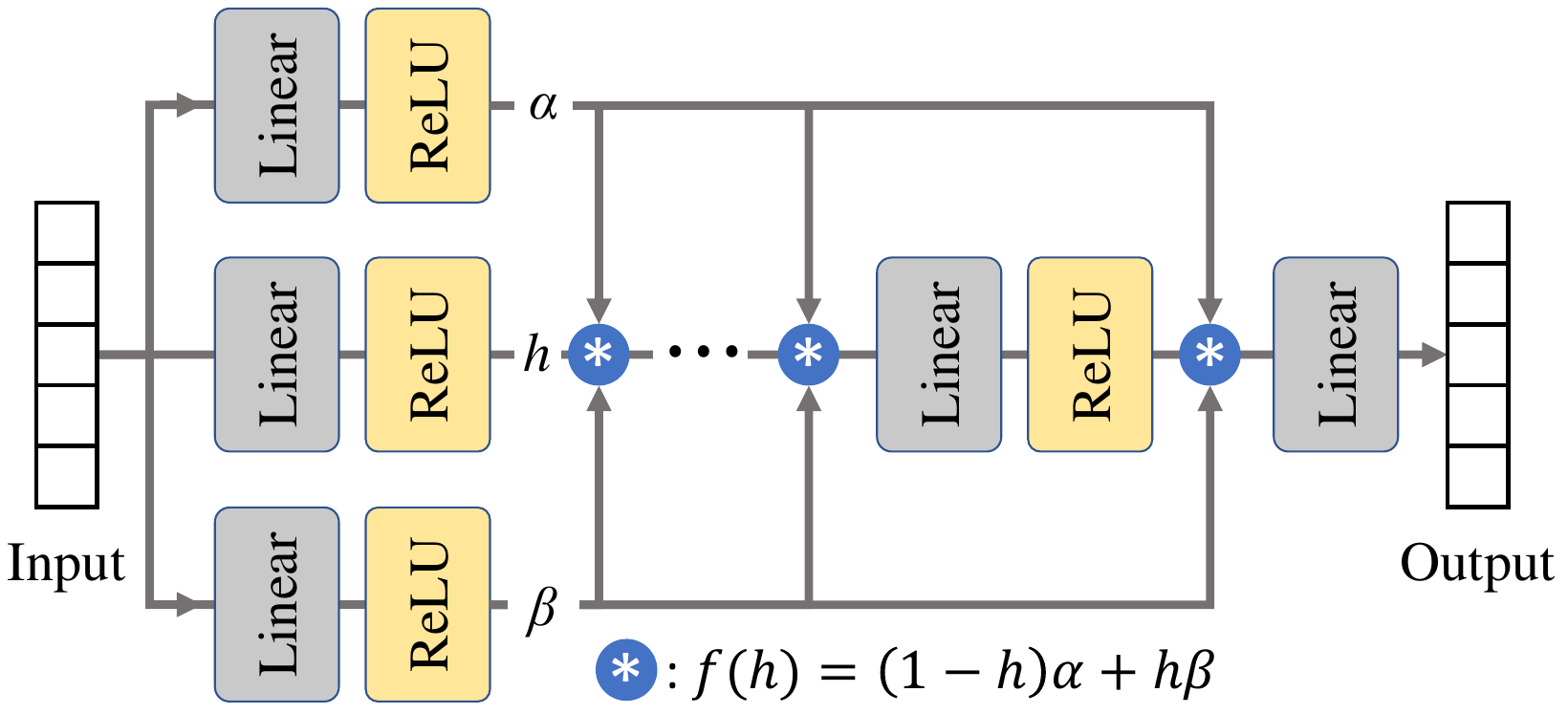}}\\
    \subfloat[]{\includegraphics[width=\textwidth]{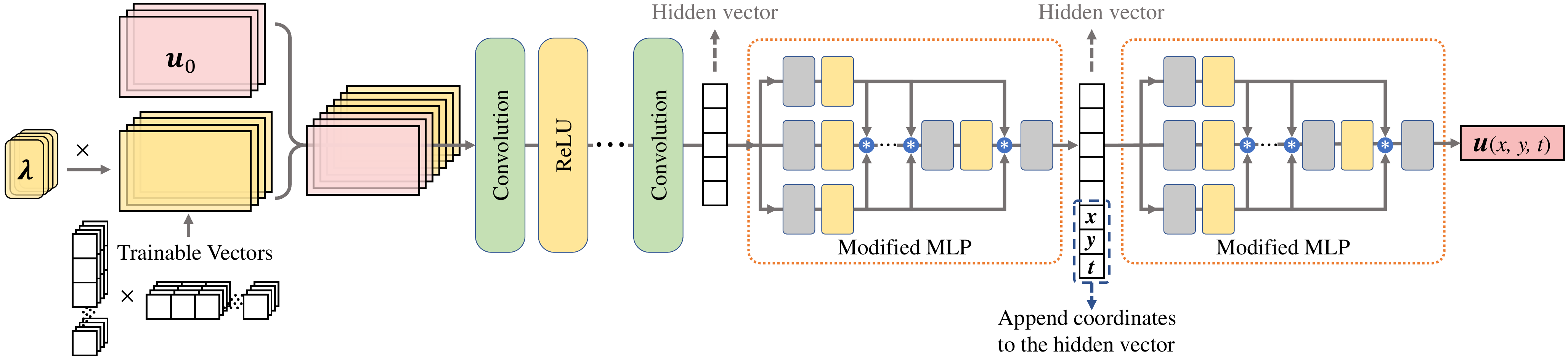}}
    \caption{(a)Structure of modified MLP. (b) The PINN structure we used}
    \label{fig:pinn}
\end{figure}

\paragraph{Modified MLP}
For PINN, we apply the modified, multi-layer perceptron (MLP) which has been ``proved to be empirically \& uniformly better than the conventional DeepONet architecture'' \cite{wang2022improved}. The structure of this modified MLP is shown in Fig.~\ref{fig:pinn}(a)

PINN consists of a CNN encoder followed by a modified MLP to compress the input parameters: initial condition and viscosity to a hidden vector. Then the space and time coordinates are appended to the hidden vector and passed to another modified MLP. The structure is shown in Fig.~\ref{fig:pinn}(b). The encoder consists of $4$ layers CNN, with kernel size equals to $8$, $6$, $6$, and $8$ respectively. The first three convolution layers has a stride equals to $3$, while in the last layer stride equals $1$. After the CNN, the field is compressed to a vector of size $48$. For both modified MLPs, each has $25$ hidden layers with $40000$ neurons in each layer.  

Because we have to evaluate the boundary values in PINN, the PINN has to deal with all the points, including those at the bottom and right boundaries which are omitted in PPNN and FNO due to these points have exact same value as those points at the top and left boundaries (periodic BC). The output is the predicted solution with two velocity components $u_x$ and $u_y$.
In training, the total loss $L$ is a weighted summation of $4$ components:  
\begin{equation}
    L = w_\mathrm{ic}L_\mathrm{ic} + w_\mathrm{eq}L_\mathrm{eq} + w_\mathrm{bc}L_\mathrm{bc}+ w_\mathrm{d}L_\mathrm{d}
    \label{eq:loss}
\end{equation}
where $w_\mathrm{ic}=20$, $w_\mathrm{bc}=1$, $w_\mathrm{eq}=1$ and $w_\mathrm{d}=20$ are the balancing weights for the loss terms initial condition loss $L_\mathrm{ic}$, equation loss $L_\mathrm{eq}$, boundary loss $L_\mathrm{bc}$ and data loss $L_\mathrm{d}$, respectively.

\subsection{DeepONet}
\begin{figure}[ht!]
    \centering
    \includegraphics[width=0.8\textwidth]{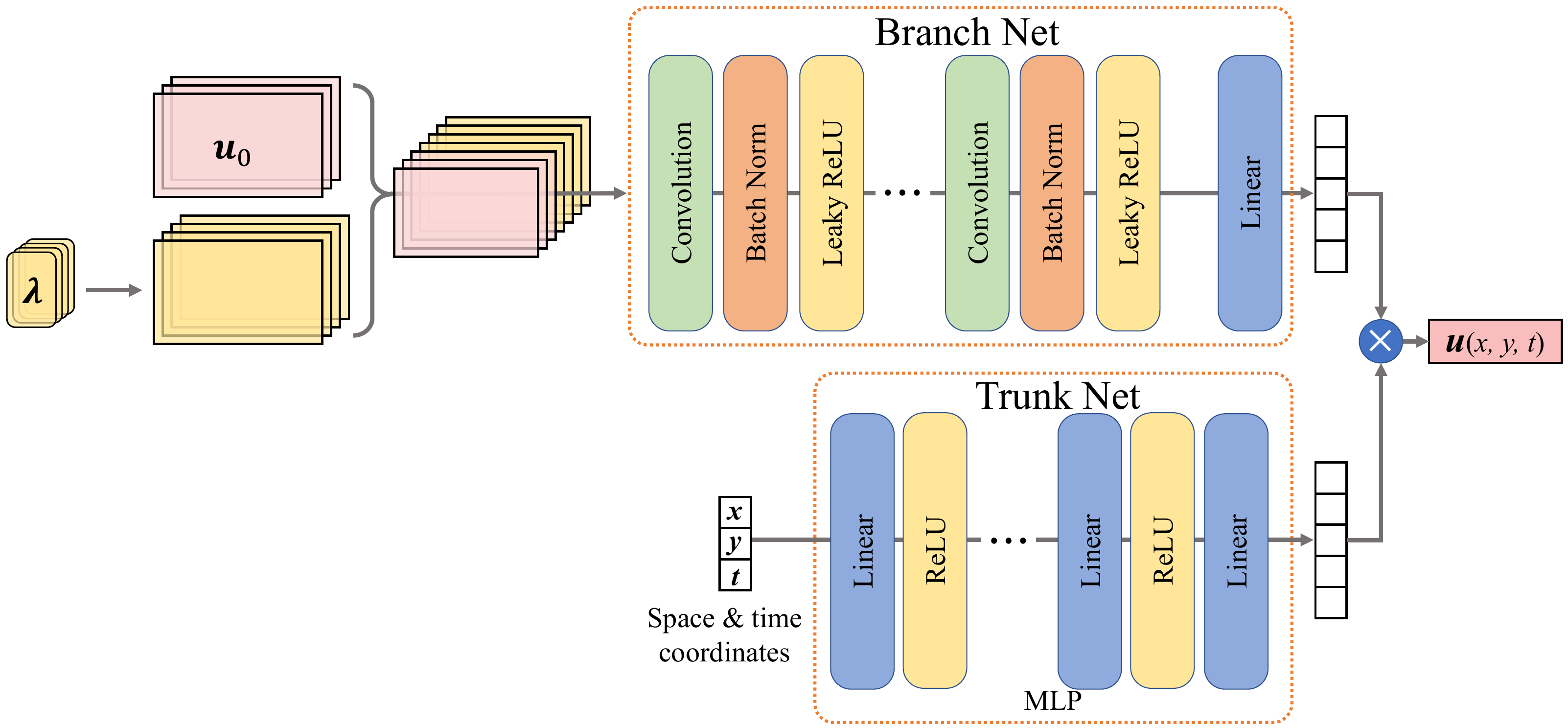}
    \caption{Structure of DeepONet (only shows the network used to predict $u_x$, the network used to predict $u_y$ shares completely identical structure.)}
    \label{fig:deeponet}
\end{figure}

We followed the description of DeepONet as proposed in \cite{lu2022comprehensive, lu2021learning}, the structure of DeepONet is shown in Fig.~\ref{fig:deeponet}. 
Based on the numerical experiments, we find that use two separate neural networks to predict the two velocity components achieves slightly better performance. DeepONet consists of two parts: branch net and trunk net. The branch net is responsible for handling the hidden vector encoded from the input parameters, while the trunk net deals with the space \& time coordinates. The output of these two sub-nets are element-wisely multiplied with each other and generates the output: the velocity components $u_x$ or $u_y$.  The DeepONet has $8$ convolution layers with batch norm in the branch net, while the trunk net consists of 4 linear layers. For DeepONet-L, the branch net contains $11$ convolution layers while trunk net has 4 linear layers. For more details please refer to the source code.

\subsection{Fourier Neural Operators (FNO)}
\label{sec:fno_more}
\begin{figure}[!ht]
    \centering
    \subfloat[]{\includegraphics[width=0.95\textwidth]{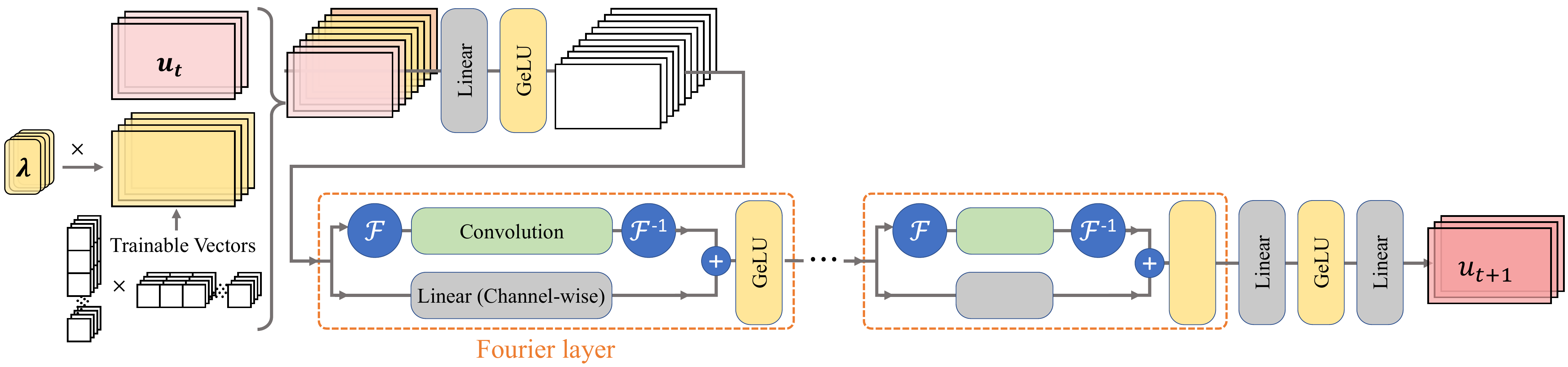}}\\
    \subfloat[]{\includegraphics[width=0.95\textwidth]{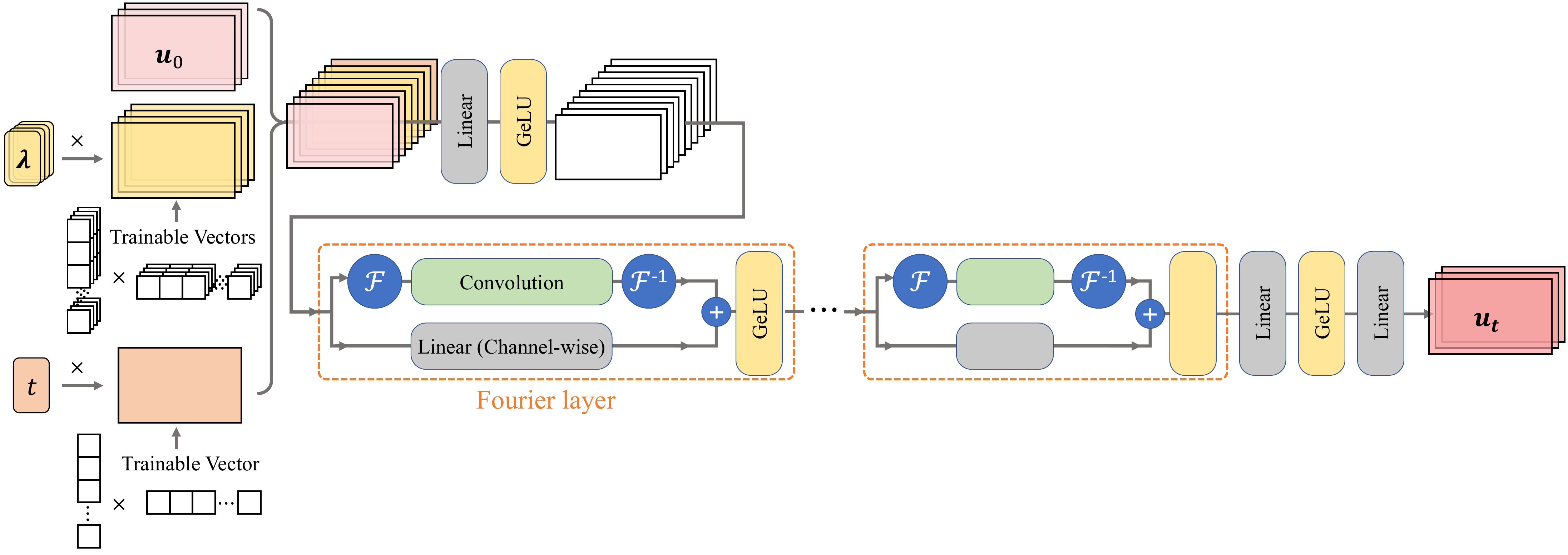}}
    \caption{Structure of Fourier neural operators (FNOs). \textbf{a}. Structure of the autoregressive FNO (aFNO). \textbf{b}. structure of continuous operator FNO (FNO)}
    \label{fig:fno}
\end{figure}

As discussed in the main text, FNO can be formulated either as a autoregressive model like PPNN or a continuous operator like DeepONet/PINN. To distinguish the two formulations, we name the autoregressive FNO as aFNO. aFNO learns the following mapping relationship $\mathcal{G}$:
\begin{equation}
    \mathcal{G}: \,\,\bm{u}_{t;\,\bm{\lambda}} \mapsto \bm{u}_{t+1;\,\bm{\lambda}}
\end{equation}
where $\bm{u}$ is the state variable, $\bm{\lambda}$ represents the physical parameters. 
The core component of FNO is the Fourier layer, which is shown in Fig.~\ref{fig:fno}. One Fourier layer consists of two parts (as shown in the orange dash box of Fig.~\ref{fig:fno}). One part contains spatial Fourier transformation, a convolution layer and inverse Fourier transformation. The other part is a channel-wise linear layer. The output of two parts are summed together. Since we want to learn a dynamic process and the evaluation position remain unchanged, we replace the original space coordinates input~\cite{li2020fourier} with the time coordinate. In total there are 5 Fourier layers and each deals with 20 channels and projects to/from 12 Fourier modes. Here we follow the original FNO paper to use GeLU~\cite{hendrycks2016gaussian} as the activation function for every layer

\begin{figure}
    \centering
    \includegraphics[width=0.5\textwidth]{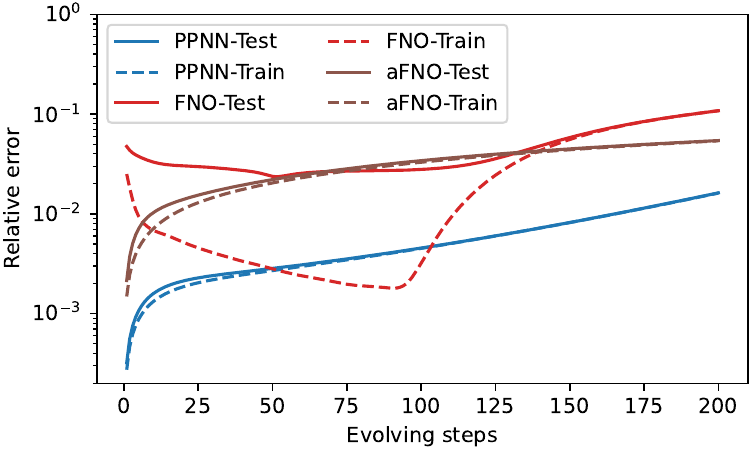}
    \caption{Comparison of the log-scale relative error $\epsilon_t$ on testing parameter set (solid lines) and training parameter set (dashed lines) of the PPNN, FNO, and aFNO respectively of the velocity $\bm{u}$ in the viscous Burgers' equation evaluated on 100 randomly generated testing (unseen) parameters (solid line), and 100 randomly selected training parameters. Only the first $100$ time steps are used for training.}
    \label{fig:fno_error}
\end{figure}

The difference of the two formulations of FNOs in terms of relative error is shown in Fig.~\ref{fig:fno_error}. Although FNO is free of the error accumulation issue, aFNO shows lower relative error in the extrapolation range. However, compared to PPNN, aFNO has a significant performance gap in both training set and testing set. 

\section{Details of the case settings and training data generation}
\label{sec:detail}

\subsection{FitzHugh–Nagumo reaction diffusion equations}
The RD equations are solved by a finite difference solver on a $256\times 256$ mesh. The spatial derivatives are computed by the $6^{th}$ order accuracy central difference scheme and time stepping is based on the forward Euler method with the time step of $\delta t = 1\times10^{-5}$. The reference solution snapshots at every other 200 timesteps are collected for the neural network training, and thus the timestep size of the neural model is $\Delta t = 200\delta t$. The whole training dataset includes trajectories of sparse reference solutions at $8$ different $\gamma$ uniformly distributed in $[0.6,\,1.3]$ and each $\gamma$ comes with $32$ randomly generated ICs. In total, the training set contains $256$ trajectories, and each trajectory includes $100$ snapshots covering the temporal range of $T = 100\Delta t$, which starts from the $2001^{st}$ numerical step. In PPNN, the PDE-preserving portion is defined on a coarse mesh of size $48\times48$ with the same learning timestep size as the trainable part ($\Delta t' = \Delta t = 200 \delta t$).

\subsection{Viscous Burgers equation}
The Burgers' equations are solved via a finite difference solver on a $256\times 256$ mesh, using a $3^{rd}$ order accuracy up-wind scheme for the convection term and $6^{th}$ order accuracy central difference scheme for the diffusion term. Forward Euler method is used for the time stepping, with a timestep size of $\delta t = 1\times10^{-4}$. Similar to RD case, the reference solution snapshots are collected for DNN training at every other 200 numerical timesteps ($\Delta t = 200\,\delta t$). The training dataset consists of sparse reference solutions of six different $\nu$, and there are $36$ randomly generated ICs for each $\nu$. In total, the training dataset includes $216$ trajectories and each trajectory includes 100 snapshots covering the temporal range of $T = 100\Delta t$, started from the first numerical step. In PPNN, the PDE-preserving portion is defined on a coarse mesh of size $32\times32$ with the same learning step size as the trainable part ($\Delta t' = \Delta t = 200\delta t$).

\subsection{Naiver-Stokes equations}
The N-S equation is solved on a fine mesh of $400\times100$ grids using the Pressure-Implicit with Splitting of Operators (PISO) algorithm. All the numerical discretizations are conducted using OpenFOAM, which is an open-source C++ library for FVM. We simulate the NS equation for 6000 numerical steps, with numerical timestep $\delta t = 0.01$. The reference snapshots are collected every other 20 numerical steps, beginning from the $160^{th}$ step, to form the training trajectory. Namely, the learning step is set as $\Delta t = 80\delta t$. In the training set, there are 45 trajectories, each of which includes 73 snapshots covering the total time of $T = 73\Delta t$. The same FVM discretization scheme is used to construct the PDE-preserving portion, which is defined on a coarse mesh of size $100\times25$ and operated by several sub-iterations with a step size of $\Delta t' = 2\delta t$. 
\section{Physics-informed variants of DeepONet and FNO}
\label{sec:pideeponet}
\subsection{Physics-informed DeepONet (Pi-DeepONet)}

Pi-DeepONet was initially proposed by Wang et al.~\cite{wang2021learning}, where the PDE residual is minimized to regulate the DeepONet training. A point-wise PDE residual is evaluated using automatic differentiation. Here, we expand Pi-DeepONet to address three-dimensional spatiotemporal dynamics with parameterized physical properties. In this section, we test Pi-DeepONet with structures similar to the DeepONet used in previous sections on the Burgers' equation. However, we replace the linear layers with the modified MLP (as shown in Fig.~\ref{fig:pinn}), which is also employed in the original Pi-DeepONet. Another notable distinction between DeepONet and Pi-DeepONet is that Pi-DeepONet must predict the $x$ and $y$ components of the velocity field simultaneously to evaluate the equation residual, while the data-driven DeepONet predicts the velocity components separately. Here, we also examine Pi-DeepONet with various combinations of weighting hyperparameters ($w_\mathrm{ic}$, $w_\mathrm{bc}$, $w_\mathrm{eq}$, and $w_\mathrm{d}$, see Supplementary Eq.~\ref{eq:loss}) for different components in the loss function. These variants of Pi-DeepONet are named as Pi-DeepONet$1\sim3$ and Pi-DeepONet*, with the weighting parameters presented in Table~\ref{tab:weight}.

\begin{figure}[!htp]
    \centering
    \includegraphics[width=0.65\textwidth]{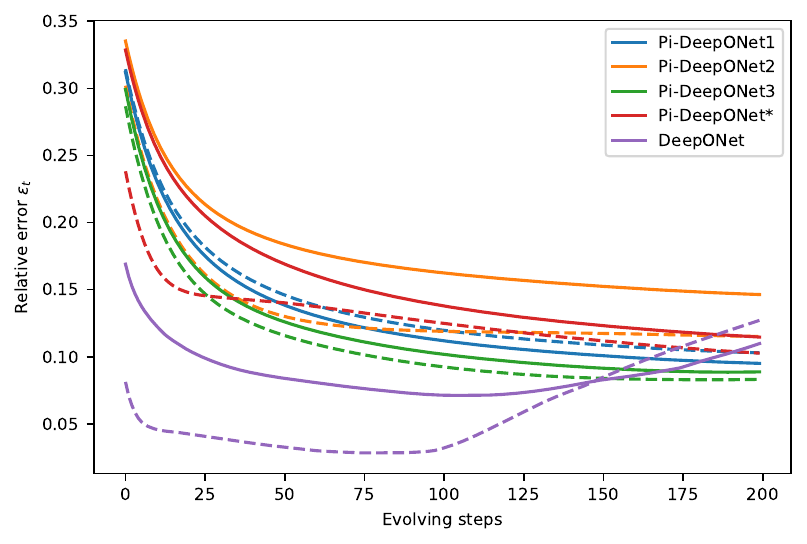}
    \caption{Relative error $\epsilon_t$ of different variants of (Pi-)DeepONet in the Burgers' equation case, averaged over $100$ randomly selected parameters $\bm{\lambda}$ from the training set (dashed lines) and $100$ from the testing set, including unseen initial conditions and physical parameters, i.e. viscosity (solid lines), respectively. Pi-DeepONet 1(blue), Pi-DeepONet 2 (orange), Pi-DeepONet 3 (green), Pi-DeepONet* (red), DeepONet (purple).}
    \label{fig:pideeponet}
\end{figure}

\begin{table}[!h]
    \centering
    \begin{tabular}{c|c|c|c|c||c|c|c|c|c}
    \hline
                     & $w_\mathrm{ic}$ & $w_\mathrm{bc}$ & $w_\mathrm{eq}$ & $w_\mathrm{d}$ & & $w_\mathrm{ic}$ & $w_\mathrm{bc}$ & $w_\mathrm{eq}$ & $w_\mathrm{d}$ \\
                     \hline
      Pi-DeepONet 1  & 20        & 1        & 1        & 20 & PINO-L       & 1        & 1        & 1        & 5\\
      \hline
      Pi-DeepONet 2  & 5        & 1        & 1        & 20 & PINO-L 2     & 1        & 1        & 1        & 1\\
      \hline
      Pi-DeepONet 3  & 20        & 1        & 1        & 1 & PINO-L 3     & 1        & 1        & 1        & 10\\
      \hline
      Pi-DeepONet*  & 0        & 0        & 0        & 1 & PINO-L 4     & 5        & 0        & 2        & 5\\
        \hline
        
    \end{tabular}
    \caption{Balancing weight used in Pi-DeepONet and PINO-L}
    \label{tab:weight}
\end{table}
Fig.~\ref{fig:pideeponet} presents the relative error of Pi-DeepONet compared to the DeepONet. The dashed lines represent performance on the training set, while the solid lines illustrate the relative error on the testing set. The performance of Pi-DeepONet varies significantly when trained with different loss weighting parameters; in most cases, it underperforms compared to the DeepONet. This is probably due to the fact that original Pi-DeepONet and DeepONet do not share the same DNN structure. To isolate the influence of the physics-informed loss, we contrast the results of Pi-DeepONet $1\sim3$ with Pi-DeepONet*, which has an identical DNN structure but is purely data-driven as the PDE weighting terms $w_{ic}$, $w_{bc}$, and $w_{eq}$ are set to zero. We find that even with an identical DNN structure, an improvement in prediction accuracy through physics-informed loss is not guaranteed. Instead, any potential improvement depends heavily on the proper tuning of weighting hyperparameters. Furthermore, it's worth highlighting that most of the Pi-DeepONets display a larger prediction error at the start than at the end of trajectories. This is likely because the initial condition is sampled from a very high-dimensional space, posing a learning challenge for Pi-DeepONet. However, as more time steps are rolled out, the velocity decays, and the overall dimensions of the velocity field decrease, making it easier for Pi-DeepONet to predict.

\subsection{Physics-informed neural operator (PINO)}
PINO~\cite{li2021physics}, derived from FNO, utilizes the PDE residual evaluated in the Fourier space to regulate DNN training. Utilizing the Fourier operator, PINO demonstrates discretization-invariance~\cite{li2021physics} and shows good generalizability in the initial condition space with moderate dimensions. However, despite incorporating the governing equation into the loss function, PINO still suffers from poor generalizability across different physical parameters. In particular, PINO heavily relies on fine-tuning to work with unseen physical parameters (e.g., Reynolds number, viscosity)~\cite{li2021physics}, undermining its data-efficiency within the training range. The issue could be partly attributed to its design that does not include the physical parameters as input information. To rectify this, we extended the PINO structure to accommodate physical parameter information by introducing these parameters as an extra channel. We purposely controlled the number of trainable parameters in PINO (1.57M) to be comparable to other baseline methods. To explore the potential performance when employing more trainable parameters, we included four additional models with a larger number of trainable parameters (about 6.30M). Similar to Section~\ref{sec:pideeponet}, these models are trained with different loss weighting hyperparameters (see Table~\ref{tab:weight}) and are named PINO-L, PINO-L2, PINO-L3, and PINO-L4, respectively. Since the original PINO paper used a three-dimensional FNO (3D-FNO) as the core structure for 2D dynamic problems, a structure fundamentally different from the FNO we have previously discussed, we also introduced a purely data-driven three-dimensional FNO as a baseline, which we refer to as FNO3d.

\begin{figure}[!htp]
    \centering    
    \subfloat{\includegraphics[width=0.5\textwidth]{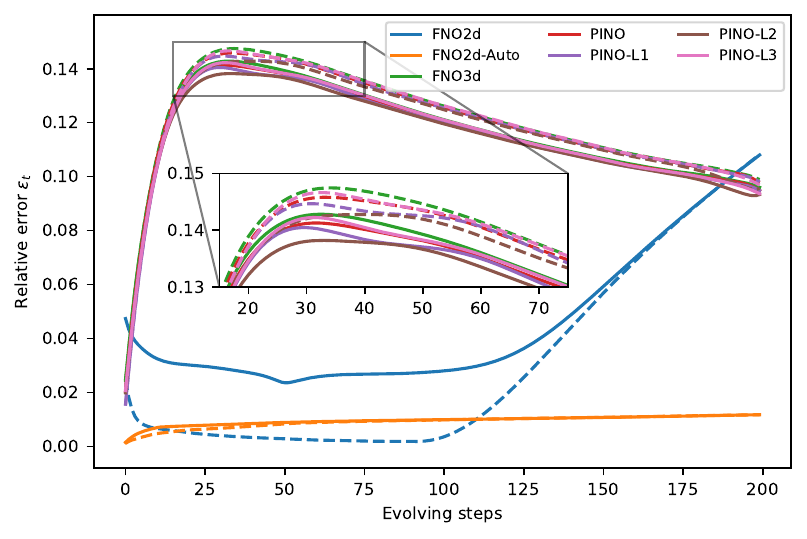}}
    \subfloat{\includegraphics[width=0.5\textwidth]{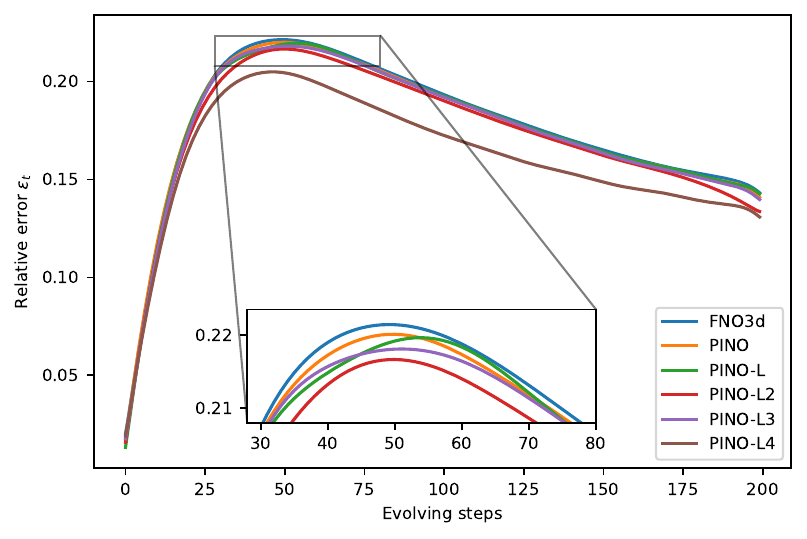}}
    \caption{Relative error $\epsilon_t$ of different variants of FNO/PINOs in the Burgers' equation case. Left panel shows the averaged relative error of FNO (blue), aFNO (orange), FNO3d (green), PINO (red), PINO-L (purple), PINO-L2 (brown), and PINO-L3 (pink) on $100$ randomly selected parameters $\bm{\lambda}$ from the training set (dashed lines) and $100$ from the testing set (including unseen initial conditions and physical parameters (i.e. viscosity), solid lines), respectively. The right panel shows the averaged relative error $\epsilon_t$ tested with one specific viscosity ($\nu=0.02$) from the training set.}
    \label{fig:fnos}
\end{figure}
Fig.~\ref{fig:fnos} compares the relative prediction errors of different variants of FNO/PINO. When comparing PINOs with FNO3d, which share the identical DNN structure, the improvement is quite marginal. This indicates that, like Pi-DeepONets, the physics-informed loss does not notably improve the model performance. This marginal improvement also aligns with the findings reported in the original PINO paper~\cite{li2021physics}. However, it's worth noting that in the original PINO paper, PINO was compared to FNO in a scenario of sparse label data, which is a different context than our comparison. Given our full-resolution training dataset, the improvement of PINO over FNO appears even more marginal here. However, unlike Pi-DeepONet, we didn't observe a large variation in prediction accuracy when altering the weighting terms of loss function components, suggesting that PINO is relatively less sensitive to the weighting hyperparameters.Moreover, we observe that the FNO2d structures show much better performance than the FNO3d structures (used in PINO). We further tested PINO's performance under a specific physical parameter (i.e., viscosity $\nu=0.02$), as shown in the right panel of Fig.~\ref{fig:fnos}. This is the scenario studied in the original PINO paper. The brown curve represents the relative error of PINO-L4, trained exclusively with viscosity $\nu=0.02$. In comparison with other PINOs trained with a variety of viscosities, PINO-L4 exhibits a lower prediction error, suggesting that PINO struggles to generalize across different physical parameters.

\section{Extra contours}
\label{sec:contours}
\subsection{Contour of the ``unknown'' magnetic field added to the Naiver-Stokes equation}
\begin{figure}[!h]
    \centering
    \includegraphics[width=0.8\textwidth]{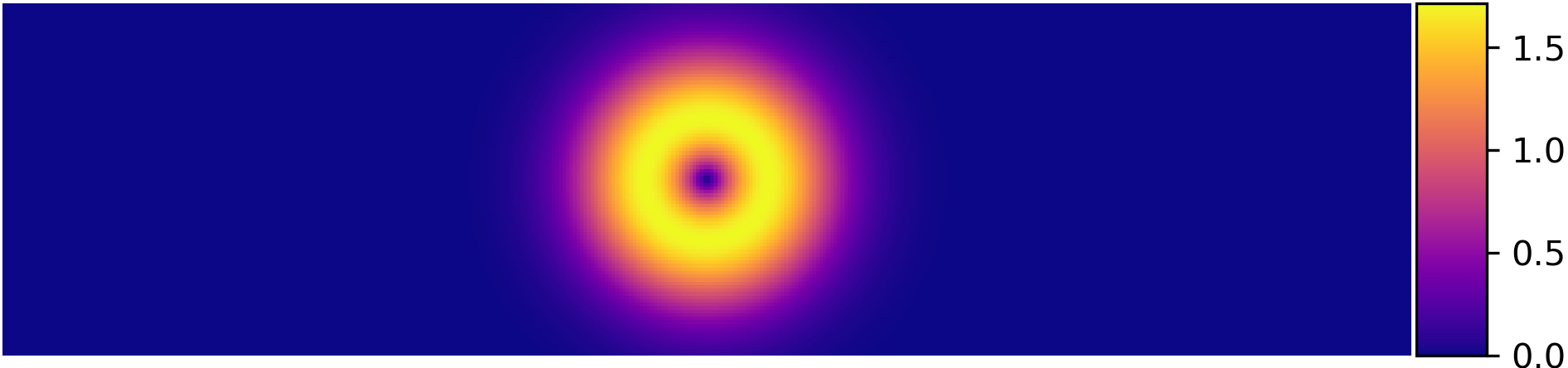}
    \caption{The magnitude distribution in the computational domain of the ``unknown'' source term added to the Naiver-Stokes equation}
    \label{fig:magnetic}
\end{figure}
\subsection{Extra comparison with black-box baselines}
\begin{figure}[!h]
    \centering
    \includegraphics[width=\textwidth]{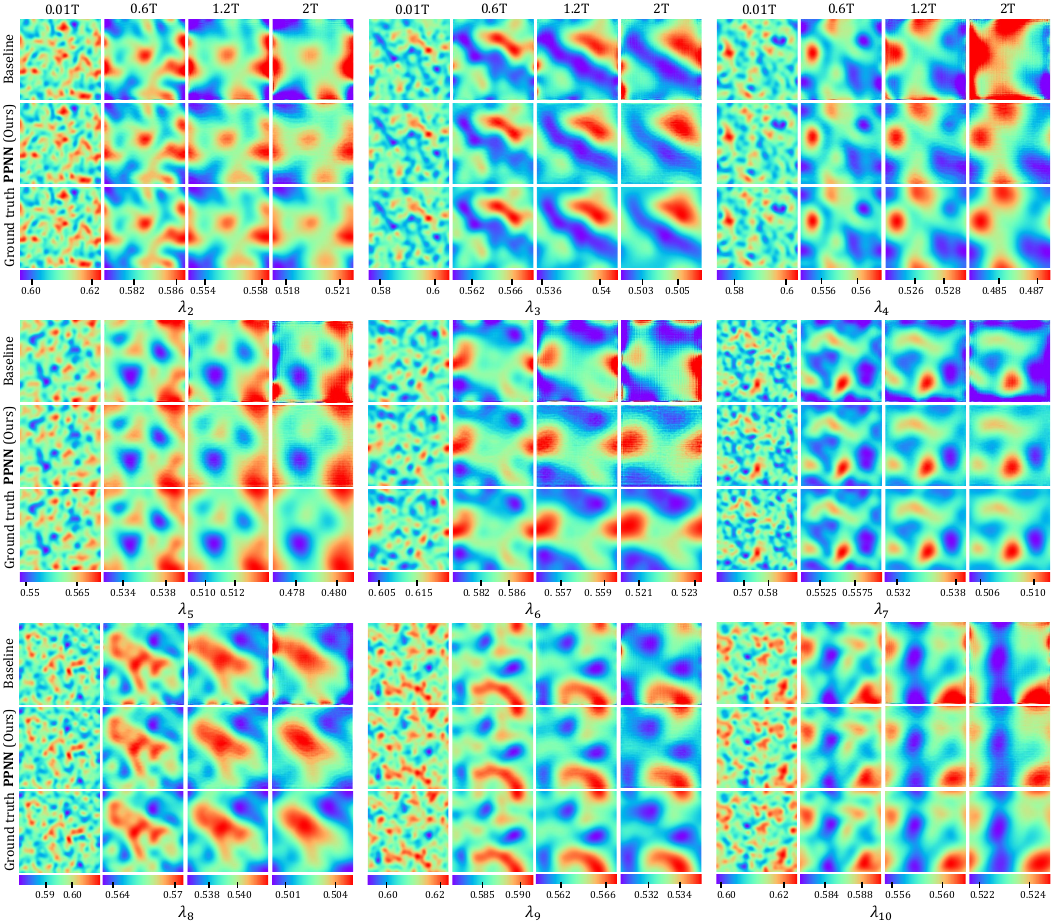}
    \caption{Predicted solution snapshots of reactant $u$ for the reaction-diffusion (RD) equations, obtained by black-box ConvResNet (baseline), and PPNN (ours); compared against ground truth. where $\bm{\lambda}_2 - \bm{\lambda}_{10}$ are testing parameters, which are not in the training set. }
    \label{fig:rdmore}
\end{figure}

\begin{figure}[!h]
    \centering
    \includegraphics[width=\textwidth]{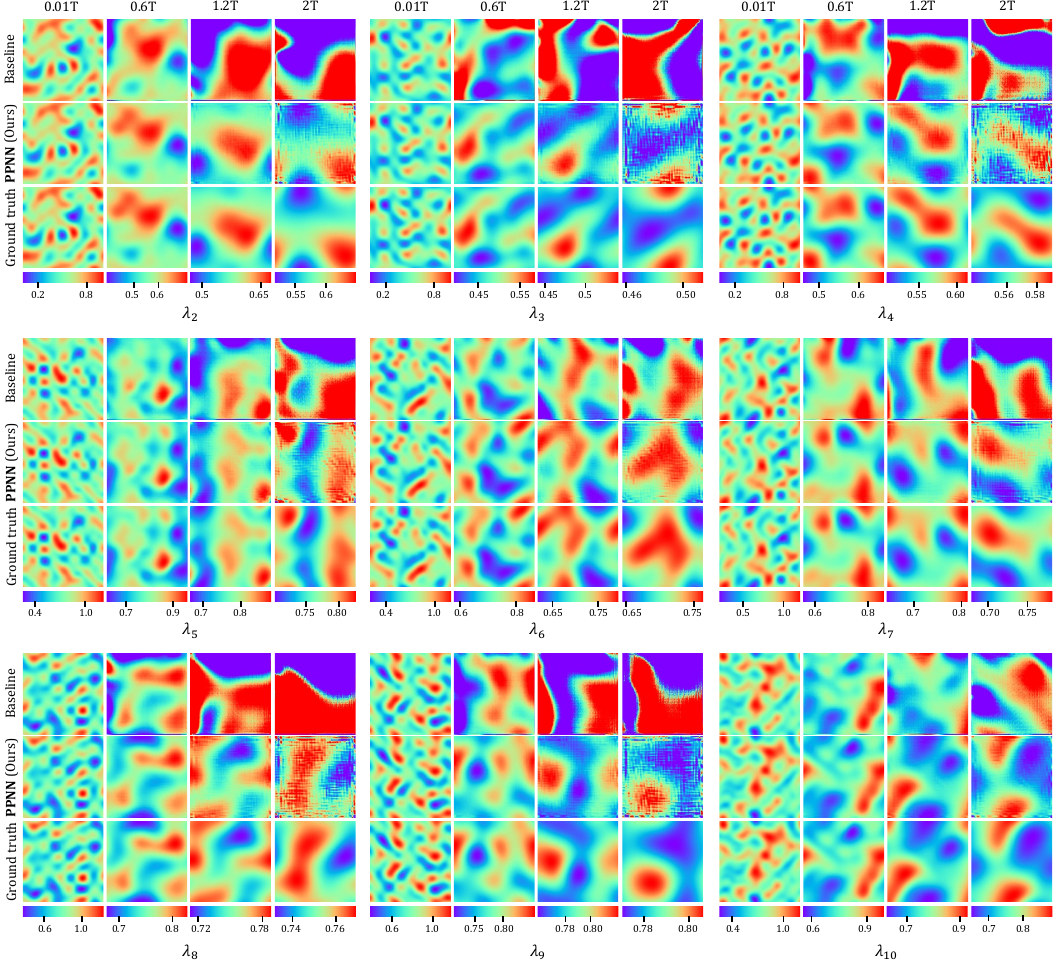}
    \caption{Predicted solution snapshots of reactant $u$ for the Burgers' equations, obtained by black-box ConvResNet (baseline), and PPNN (ours); compared against ground truth. where $\bm{\lambda}_2 - \bm{\lambda}_{10}$ are testing parameters, which are not in the training set. }
    \label{fig:bgmore}
\end{figure}

This section provides more contour comparison among PPNN, black-box ConvResNet and label data on the testing (unseen) dataset, for the reaction-diffusion case (Fig.~\ref{fig:rdmore}) and viscous Burgers' case (Fig.~\ref{fig:bgmore}), respectively.

\subsection{Extra comparison with neural operators}

\begin{figure}[ht!]
    \centering
    \includegraphics[width=0.8\textwidth]{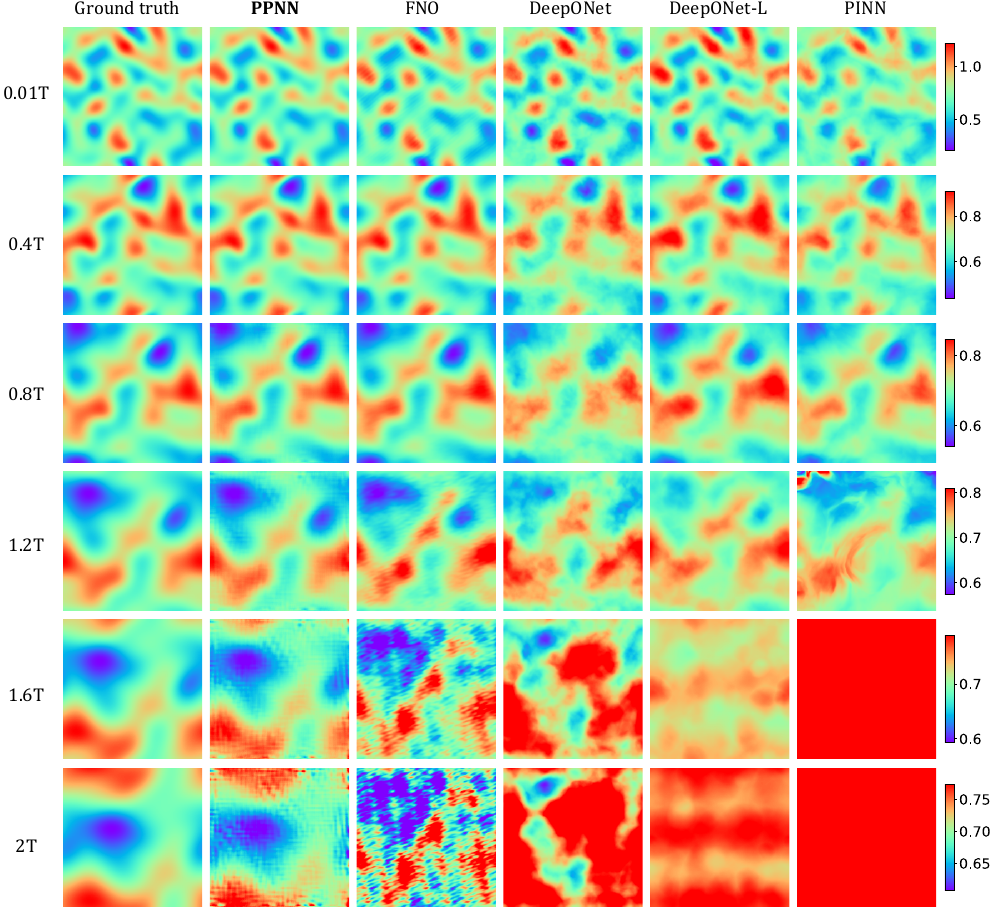}
    \caption{Predicted solution for the velocity magnitude $\lVert\bm{u}\rVert_2$ of the Burgers' equations at different time steps and training parameters $\lambda_9$. Each row represents the predicted solution at a certain time step, while each column shows the results predicted by (from left to right) numerical solver (ground truth), \textbf{PPNN}, FNO, DeepONet, DeepONet-L, and PINN respectively.}
    \label{fig:train1}
\end{figure}

\begin{figure}
    \centering
    \includegraphics[width=0.8\textwidth]{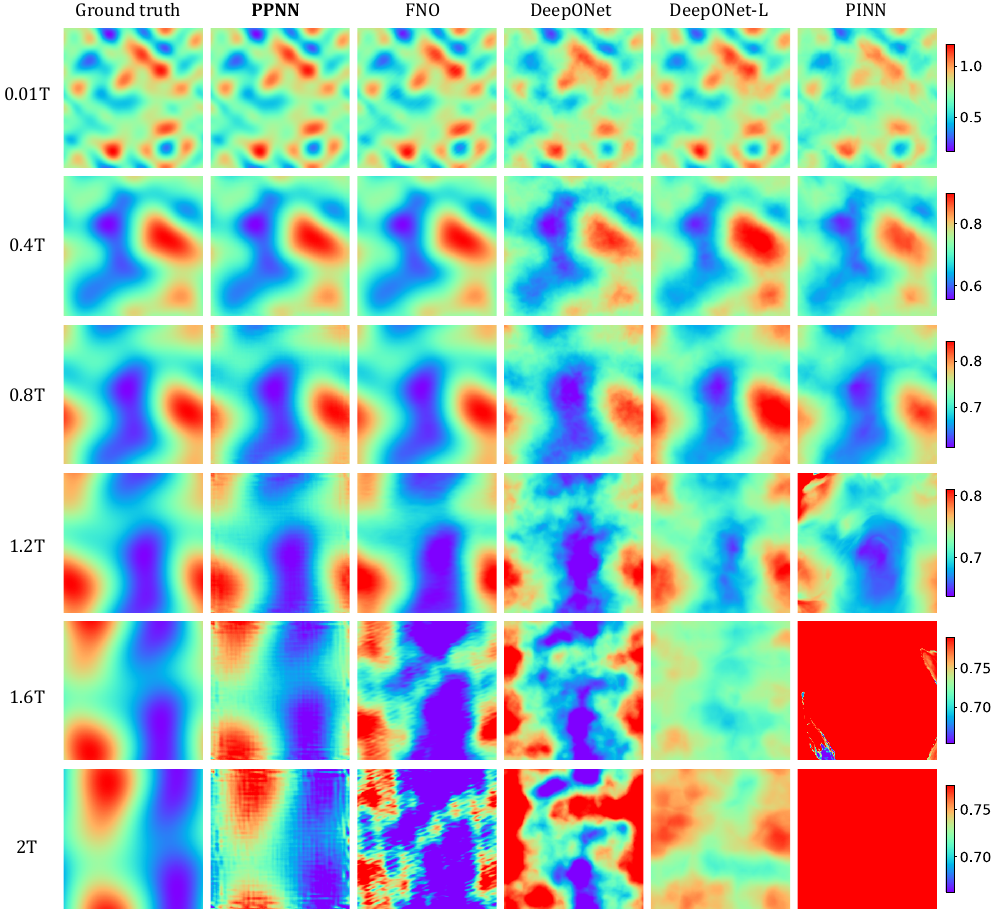}
    \caption{Predicted solution for the velocity magnitude $\lVert\bm{u}\rVert_2$ of the Burgers' equations at different time steps and training parameters $\lambda_{10}$. Each row represents the predicted solution at a certain time step, while each column shows the results predicted by (from left to right) numerical solver (ground truth), \textbf{PPNN}, FNO, DeepONet, DeepONet-L, and PINN respectively.}
    \label{fig:train2}
\end{figure}

\begin{figure}[ht!]
    \centering
    \includegraphics[width=0.8\textwidth]{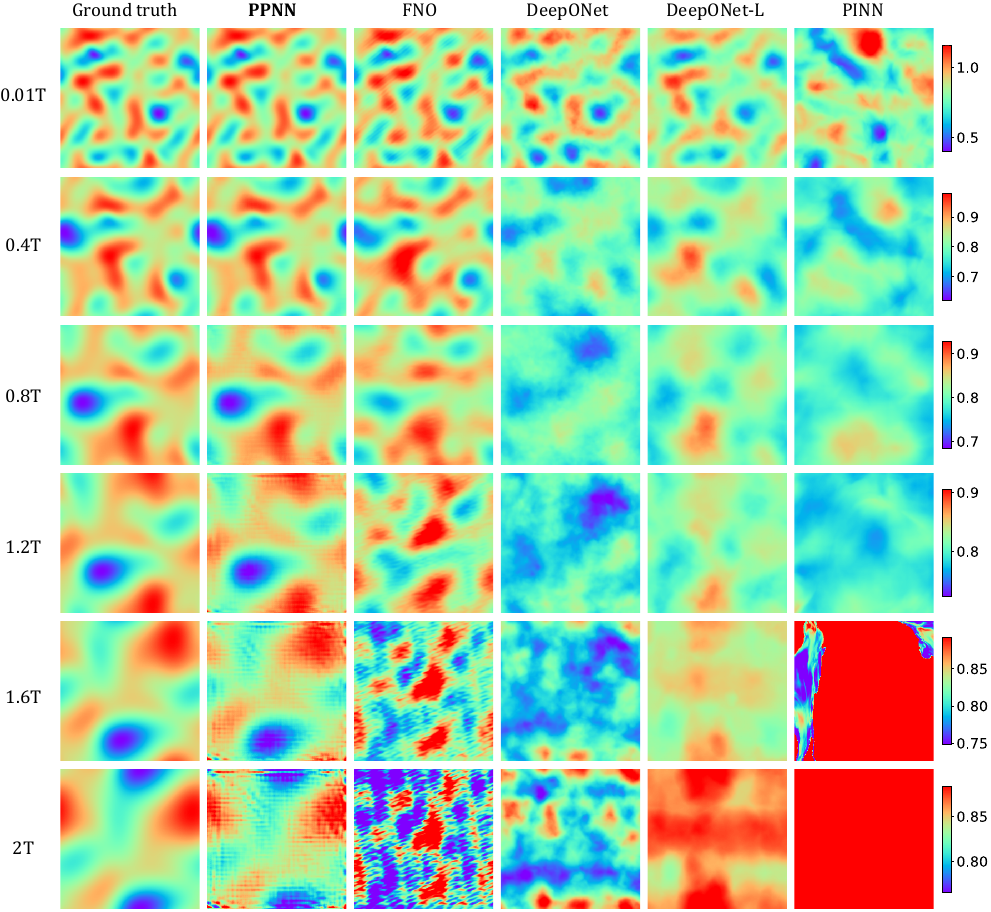}
    \caption{Predicted solution for the velocity magnitude $\lVert\bm{u}\rVert_2$ of the Burgers' equations at different time steps and unseen parameters $\lambda_{11}$. Each row represents the predicted solution at a certain time step, while each column shows the results predicted by (from left to right) numerical solver (ground truth), \textbf{PPNN}, FNO, DeepONet, DeepONet-L, and PINN respectively.}
    \label{fig:test1}
\end{figure}

\begin{figure}
    \centering
    \includegraphics[width=0.8\textwidth]{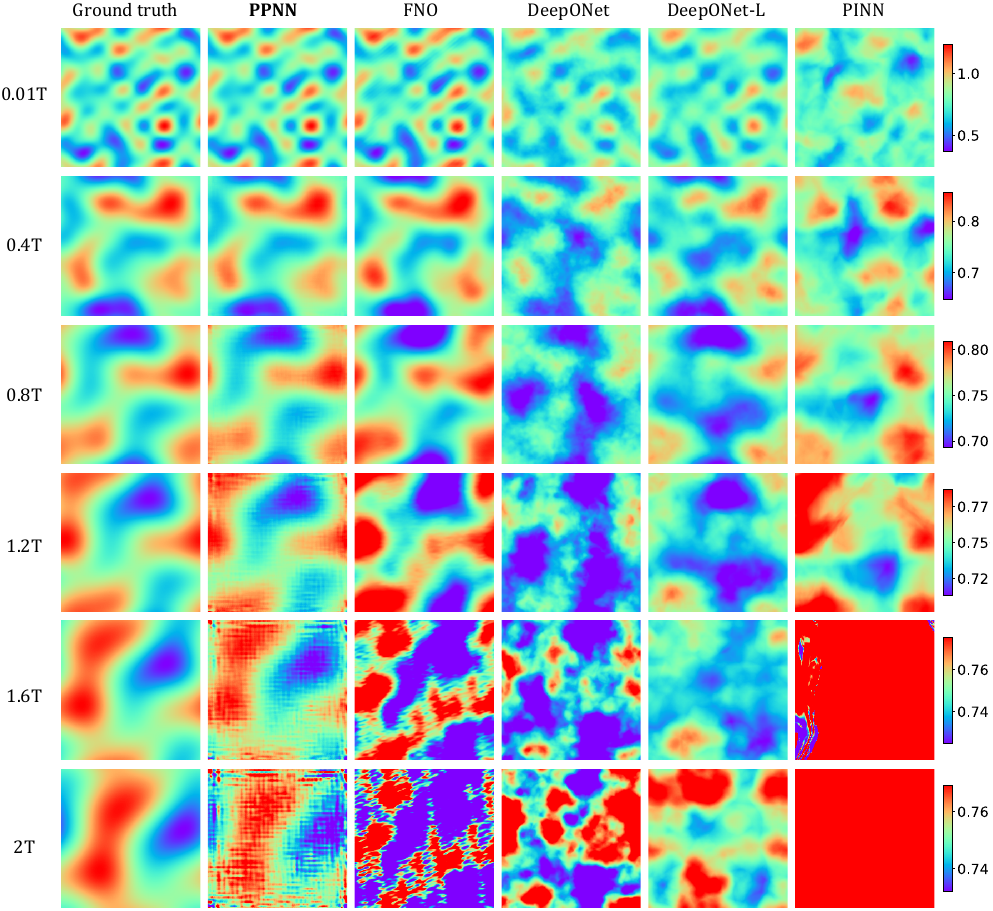}
    \caption{Predicted solution for the velocity magnitude $\lVert\bm{u}\rVert_2$ of the Burgers' equations at different time steps and unseen parameters $\lambda_{12}$. Each row represents the predicted solution at a certain time step, while each column shows the results predicted by (from left to right) numerical solver (ground truth), \textbf{PPNN}, FNO, DeepONet, DeepONet-L, and PINN respectively.}
    \label{fig:test2}
\end{figure}

In this section we provide contours comparison between PPNN and continuous function/operator learning methods for the viscous Burgers' case. Fig.~\ref{fig:train1} and Fig.~\ref{fig:train2} show the predicted results with training parameters at different time steps, while Fig.~\ref{fig:test1} and Fig.~\ref{fig:test2} show the predicted solution snapshots with testing (unseen) parameters. With training parameters, most methods have relatively accurate predictions within the interpolation range while once step into the extrapolation range in time ($t\geq T$), PPNN shows its great generalizability over the other methods though some noise could be observed. The advantage of PPNN is even more obvious with testing (unseen) parameters. Most methods fail to give an acceptable prediction even at the first time step. While FNO gives comparable results at the first few steps, with time matching forward, PPNN show much less discrepancy from the ground truth. Besides, PPNN also keeps a very consistent performance with testing and training parameters, which indicates a great generalizability among parameters.

\begin{thebibliography}{10}

    \bibitem{lui2019construction}
    Hugo~FS Lui and William~R Wolf.
    \newblock Construction of reduced-order models for fluid flows using deep feedforward neural networks.
    \newblock {\em Journal of Fluid Mechanics}, 872:963--994, 2019.
    
    \bibitem{san2019artificial}
    Omer San, Romit Maulik, and Mansoor Ahmed.
    \newblock An artificial neural network framework for reduced order modeling of transient flows.
    \newblock {\em Communications in Nonlinear Science and Numerical Simulation}, 77:271--287, 2019.
    
    \bibitem{gao2020non}
    Han Gao, Jian-Xun Wang, and Matthew~J Zahr.
    \newblock Non-intrusive model reduction of large-scale, nonlinear dynamical systems using deep learning.
    \newblock {\em Physica D: Nonlinear Phenomena}, 412:132614, 2020.
    
    \bibitem{fresca2022pod}
    Stefania Fresca and Andrea Manzoni.
    \newblock Pod-dl-rom: enhancing deep learning-based reduced order models for nonlinear parametrized pdes by proper orthogonal decomposition.
    \newblock {\em Computer Methods in Applied Mechanics and Engineering}, 388:114181, 2022.
    
    \bibitem{murata2020nonlinear}
    Takaaki Murata, Kai Fukami, and Koji Fukagata.
    \newblock Nonlinear mode decomposition with convolutional neural networks for fluid dynamics.
    \newblock {\em Journal of Fluid Mechanics}, 882, 2020.
    
    \bibitem{mohan2020spatio}
    Arvind~T Mohan, Dima Tretiak, Misha Chertkov, and Daniel Livescu.
    \newblock Spatio-temporal deep learning models of 3d turbulence with physics informed diagnostics.
    \newblock {\em Journal of Turbulence}, 21(9-10):484--524, 2020.
    
    \bibitem{maulik2021reduced}
    Romit Maulik, Bethany Lusch, and Prasanna Balaprakash.
    \newblock Reduced-order modeling of advection-dominated systems with recurrent neural networks and convolutional autoencoders.
    \newblock {\em Physics of Fluids}, 33(3):037106, 2021.
    
    \bibitem{fukami2021model}
    Kai Fukami, Kazuto Hasegawa, Taichi Nakamura, Masaki Morimoto, and Koji Fukagata.
    \newblock Model order reduction with neural networks: Application to laminar and turbulent flows.
    \newblock {\em SN Computer Science}, 2(6):1--16, 2021.
    
    \bibitem{pfaff2020learning}
    Tobias Pfaff, Meire Fortunato, Alvaro Sanchez-Gonzalez, and Peter Battaglia.
    \newblock Learning mesh-based simulation with graph networks.
    \newblock In {\em International Conference on Learning Representations}, 2020.
    
    \bibitem{han2022predicting}
    Xu~Han, Han Gao, Tobias Pfaff, Jian-Xun Wang, and Liping Liu.
    \newblock Predicting physics in mesh-reduced space with temporal attention.
    \newblock In {\em International Conference on Learning Representations}, 2022.
    
    \bibitem{baker2019workshop}
    Nathan Baker, Frank Alexander, Timo Bremer, Aric Hagberg, Yannis Kevrekidis, Habib Najm, Manish Parashar, Abani Patra, James Sethian, Stefan Wild, et~al.
    \newblock Workshop report on basic research needs for scientific machine learning: Core technologies for artificial intelligence.
    \newblock Technical report, USDOE Office of Science (SC), Washington, DC (United States), 2019.
    
    \bibitem{raissi2019physics}
    Maziar Raissi, Paris Perdikaris, and George~E Karniadakis.
    \newblock Physics-informed neural networks: A deep learning framework for solving forward and inverse problems involving nonlinear partial differential equations.
    \newblock {\em Journal of Computational physics}, 378:686--707, 2019.
    
    \bibitem{sun2020surrogate}
    Luning Sun, Han Gao, Shaowu Pan, and Jian-Xun Wang.
    \newblock Surrogate modeling for fluid flows based on physics-constrained deep learning without simulation data.
    \newblock {\em Computer Methods in Applied Mechanics and Engineering}, 361:112732, 2020.
    
    \bibitem{zhang2020physics}
    Ruiyang Zhang, Yang Liu, and Hao Sun.
    \newblock Physics-informed multi-lstm networks for metamodeling of nonlinear structures.
    \newblock {\em Computer Methods in Applied Mechanics and Engineering}, 369:113226, 2020.
    
    \bibitem{haghighat2021physics}
    Ehsan Haghighat, Maziar Raissi, Adrian Moure, Hector Gomez, and Ruben Juanes.
    \newblock A physics-informed deep learning framework for inversion and surrogate modeling in solid mechanics.
    \newblock {\em Computer Methods in Applied Mechanics and Engineering}, 379:113741, 2021.
    
    \bibitem{sun2020physics}
    Luning Sun and Jian-Xun Wang.
    \newblock Physics-constrained bayesian neural network for fluid flow reconstruction with sparse and noisy data.
    \newblock {\em Theoretical and Applied Mechanics Letters}, 10(3):161--169, 2020.
    
    \bibitem{arzani2021uncovering}
    Amirhossein Arzani, Jian-Xun Wang, and Roshan~M. D'Souza.
    \newblock Uncovering near-wall blood flow from sparse data with physics-informed neural networks.
    \newblock {\em Physics of Fluids}, 33(7):071905, 2021.
    
    \bibitem{lu2021physics}
    Lu~Lu, Raphael Pestourie, Wenjie Yao, Zhicheng Wang, Francesc Verdugo, and Steven~G Johnson.
    \newblock Physics-informed neural networks with hard constraints for inverse design.
    \newblock {\em SIAM Journal on Scientific Computing}, 43(6):B1105--B1132, 2021.
    
    \bibitem{zhang2022analyses}
    Enrui Zhang, Ming Dao, George~Em Karniadakis, and Subra Suresh.
    \newblock Analyses of internal structures and defects in materials using physics-informed neural networks.
    \newblock {\em Science advances}, 8(7):eabk0644, 2022.
    
    \bibitem{han2018solving}
    Jiequn Han, Arnulf Jentzen, and E~Weinan.
    \newblock Solving high-dimensional partial differential equations using deep learning.
    \newblock {\em Proceedings of the National Academy of Sciences}, 115(34):8505--8510, 2018.
    
    \bibitem{zhang2019quantifying}
    Dongkun Zhang, Lu~Lu, Ling Guo, and George~Em Karniadakis.
    \newblock Quantifying total uncertainty in physics-informed neural networks for solving forward and inverse stochastic problems.
    \newblock {\em Journal of Computational Physics}, 397:108850, 2019.
    
    \bibitem{yang2019adversarial}
    Yibo Yang and Paris Perdikaris.
    \newblock Adversarial uncertainty quantification in physics-informed neural networks.
    \newblock {\em Journal of Computational Physics}, 394:136--152, 2019.
    
    \bibitem{kharazmi2021hp}
    Ehsan Kharazmi, Zhongqiang Zhang, and George~Em Karniadakis.
    \newblock hp-vpinns: Variational physics-informed neural networks with domain decomposition.
    \newblock {\em Computer Methods in Applied Mechanics and Engineering}, 374:113547, 2021.
    
    \bibitem{jagtap2020conservative}
    Ameya~D Jagtap, Ehsan Kharazmi, and George~Em Karniadakis.
    \newblock Conservative physics-informed neural networks on discrete domains for conservation laws: Applications to forward and inverse problems.
    \newblock {\em Computer Methods in Applied Mechanics and Engineering}, 365:113028, 2020.
    
    \bibitem{lu2021learning}
    Lu~Lu, Pengzhan Jin, Guofei Pang, Zhongqiang Zhang, and George~Em Karniadakis.
    \newblock Learning nonlinear operators via deeponet based on the universal approximation theorem of operators.
    \newblock {\em Nature Machine Intelligence}, 3(3):218--229, 2021.
    
    \bibitem{li2020fourier}
    Zongyi Li, Nikola~Borislavov Kovachki, Kamyar Azizzadenesheli, Kaushik Bhattacharya, Andrew Stuart, Anima Anandkumar, et~al.
    \newblock Fourier neural operator for parametric partial differential equations.
    \newblock In {\em International Conference on Learning Representations}, 2020.
    
    \bibitem{wang2021learning}
    Sifan Wang, Hanwen Wang, and Paris Perdikaris.
    \newblock Learning the solution operator of parametric partial differential equations with physics-informed deeponets.
    \newblock {\em Science advances}, 7(40):eabi8605, 2021.
    
    \bibitem{goswami2022physics}
    Somdatta Goswami, Minglang Yin, Yue Yu, and George~Em Karniadakis.
    \newblock A physics-informed variational deeponet for predicting crack path in quasi-brittle materials.
    \newblock {\em Computer Methods in Applied Mechanics and Engineering}, 391:114587, 2022.
    
    \bibitem{jagtap2020adaptive}
    Ameya~D Jagtap, Kenji Kawaguchi, and George~Em Karniadakis.
    \newblock Adaptive activation functions accelerate convergence in deep and physics-informed neural networks.
    \newblock {\em Journal of Computational Physics}, 404:109136, 2020.
    
    \bibitem{wang2022and}
    Sifan Wang, Xinling Yu, and Paris Perdikaris.
    \newblock When and why pinns fail to train: A neural tangent kernel perspective.
    \newblock {\em Journal of Computational Physics}, 449:110768, 2022.
    
    \bibitem{wang2022respecting}
    Sifan Wang, Shyam Sankaran, and Paris Perdikaris.
    \newblock Respecting causality is all you need for training physics-informed neural networks.
    \newblock {\em arXiv preprint arXiv:2203.07404}, 2022.
    
    \bibitem{gao2021phygeonet}
    Han Gao, Luning Sun, and Jian-Xun Wang.
    \newblock {PhyGeoNet:} physics-informed geometry-adaptive convolutional neural networks for solving parameterized steady-state {PDEs} on irregular domain.
    \newblock {\em Journal of Computational Physics}, 428:110079, 2021.
    
    \bibitem{ren2022phycrnet}
    Pu~Ren, Chengping Rao, Yang Liu, Jian-Xun Wang, and Hao Sun.
    \newblock Phycrnet: Physics-informed convolutional-recurrent network for solving spatiotemporal pdes.
    \newblock {\em Computer Methods in Applied Mechanics and Engineering}, 389:114399, 2022.
    
    \bibitem{geneva2020modeling}
    Nicholas Geneva and Nicholas Zabaras.
    \newblock Modeling the dynamics of pde systems with physics-constrained deep auto-regressive networks.
    \newblock {\em Journal of Computational Physics}, 403:109056, 2020.
    
    \bibitem{gao2021super}
    Han Gao, Luning Sun, and Jian-Xun Wang.
    \newblock Super-resolution and denoising of fluid flow using physics-informed convolutional neural networks without high-resolution labels.
    \newblock {\em Physics of Fluids}, 33(7):073603, 2021.
    
    \bibitem{wandel2021teaching}
    Nils Wandel, Michael Weinmann, and Reinhard Klein.
    \newblock Teaching the incompressible navier--stokes equations to fast neural surrogate models in three dimensions.
    \newblock {\em Physics of Fluids}, 33(4):047117, 2021.
    
    \bibitem{ranade2021discretizationnet}
    Rishikesh Ranade, Chris Hill, and Jay Pathak.
    \newblock Discretizationnet: A machine-learning based solver for navier--stokes equations using finite volume discretization.
    \newblock {\em Computer Methods in Applied Mechanics and Engineering}, 378:113722, 2021.
    
    \bibitem{yao2020fea}
    Houpu Yao, Yi~Gao, and Yongming Liu.
    \newblock Fea-net: A physics-guided data-driven model for efficient mechanical response prediction.
    \newblock {\em Computer Methods in Applied Mechanics and Engineering}, 363:112892, 2020.
    
    \bibitem{mitusch2021hybrid}
    Sebastian~K Mitusch, Simon~W Funke, and Miroslav Kuchta.
    \newblock Hybrid fem-nn models: Combining artificial neural networks with the finite element method.
    \newblock {\em Journal of Computational Physics}, 446:110651, 2021.
    
    \bibitem{wang2021variational}
    Zhenlin Wang, Xun Huan, and Krishna Garikipati.
    \newblock Variational system identification of the partial differential equations governing microstructure evolution in materials: Inference over sparse and spatially unrelated data.
    \newblock {\em Computer Methods in Applied Mechanics and Engineering}, 377:113706, 2021.
    
    \bibitem{yin2022interfacing}
    Minglang Yin, Enrui Zhang, Yue Yu, and George~Em Karniadakis.
    \newblock Interfacing finite elements with deep neural operators for fast multiscale modeling of mechanics problems.
    \newblock {\em Computer Methods in Applied Mechanics and Engineering}, page 115027, 2022.
    
    \bibitem{gao2022physics}
    Han Gao, Matthew~J Zahr, and Jian-Xun Wang.
    \newblock Physics-informed graph neural galerkin networks: A unified framework for solving pde-governed forward and inverse problems.
    \newblock {\em Computer Methods in Applied Mechanics and Engineering}, 390:114502, 2022.
    
    \bibitem{liu2021physics}
    Xin-Yang Liu and Jian-Xun Wang.
    \newblock Physics-informed dyna-style model-based deep reinforcement learning for dynamic control.
    \newblock {\em Proceedings of the Royal Society A: Mathematical, Physical and Engineering Sciences}, 477(2255):20210618, 2021.
    
    \bibitem{haber2017stable}
    Eldad Haber and Lars Ruthotto.
    \newblock Stable architectures for deep neural networks.
    \newblock {\em Inverse problems}, 34(1):014004, 2017.
    
    \bibitem{lu2018beyond}
    Yiping Lu, Aoxiao Zhong, Quanzheng Li, and Bin Dong.
    \newblock Beyond finite layer neural networks: Bridging deep architectures and numerical differential equations.
    \newblock In {\em International Conference on Machine Learning}, pages 3276--3285. PMLR, 2018.
    
    \bibitem{rousseau2020residual}
    Fran{\c{c}}ois Rousseau, Lucas Drumetz, and Ronan Fablet.
    \newblock Residual networks as flows of diffeomorphisms.
    \newblock {\em Journal of Mathematical Imaging and Vision}, 62(3):365--375, 2020.
    
    \bibitem{ruthotto2020deep}
    Lars Ruthotto and Eldad Haber.
    \newblock Deep neural networks motivated by partial differential equations.
    \newblock {\em Journal of Mathematical Imaging and Vision}, 62(3):352--364, 2020.
    
    \bibitem{chamberlain2021grand}
    Ben Chamberlain, James Rowbottom, Maria~I Gorinova, Michael Bronstein, Stefan Webb, and Emanuele Rossi.
    \newblock Grand: Graph neural diffusion.
    \newblock In {\em International Conference on Machine Learning}, pages 1407--1418. PMLR, 2021.
    
    \bibitem{eliasof2021pde}
    Moshe Eliasof, Eldad Haber, and Eran Treister.
    \newblock {PDE-GCN}: Novel architectures for graph neural networks motivated by partial differential equations.
    \newblock {\em Advances in Neural Information Processing Systems}, 34, 2021.
    
    \bibitem{chen2018neural}
    Ricky~TQ Chen, Yulia Rubanova, Jesse Bettencourt, and David~K Duvenaud.
    \newblock Neural ordinary differential equations.
    \newblock {\em Advances in neural information processing systems}, 31, 2018.
    
    \bibitem{he2016deep}
    Kaiming He, Xiangyu Zhang, Shaoqing Ren, and Jian Sun.
    \newblock Deep residual learning for image recognition.
    \newblock In {\em Proceedings of the IEEE conference on computer vision and pattern recognition}, pages 770--778, 2016.
    
    \bibitem{gholami2019anode}
    Amir Gholami, Kurt Keutzer, and George Biros.
    \newblock Anode: Unconditionally accurate memory-efficient gradients for neural odes.
    \newblock {\em arXiv preprint arXiv:1902.10298}, 2019.
    
    \bibitem{shi2020finite}
    Zheng Shi, Nur~Sila Gulgec, Albert~S Berahas, Shamim~N Pakzad, and Martin Tak{\'a}{\v{c}}.
    \newblock Finite difference neural networks: Fast prediction of partial differential equations.
    \newblock In {\em 2020 19th IEEE International Conference on Machine Learning and Applications (ICMLA)}, pages 130--135. IEEE, 2020.
    
    \bibitem{innes2019differentiable}
    Mike Innes, Alan Edelman, Keno Fischer, Chris Rackauckas, Elliot Saba, Viral~B Shah, and Will Tebbutt.
    \newblock A differentiable programming system to bridge machine learning and scientific computing.
    \newblock {\em arXiv preprint arXiv:1907.07587}, 2019.
    
    \bibitem{rackauckas2020universal}
    Christopher Rackauckas, Yingbo Ma, Julius Martensen, Collin Warner, Kirill Zubov, Rohit Supekar, Dominic Skinner, Ali Ramadhan, and Alan Edelman.
    \newblock Universal differential equations for scientific machine learning.
    \newblock {\em arXiv preprint arXiv:2001.04385}, 2020.
    
    \bibitem{sun2020neupde}
    Yifan Sun, Linan Zhang, and Hayden Schaeffer.
    \newblock Neupde: Neural network based ordinary and partial differential equations for modeling time-dependent data.
    \newblock In {\em Mathematical and Scientific Machine Learning}, pages 352--372. PMLR, 2020.
    
    \bibitem{hochlehnert2021learning}
    Andreas Hochlehnert, Alexander Terenin, Steind{\'o}r S{\ae}mundsson, and Marc Deisenroth.
    \newblock Learning contact dynamics using physically structured neural networks.
    \newblock In {\em International Conference on Artificial Intelligence and Statistics}, pages 2152--2160. PMLR, 2021.
    
    \bibitem{heiden2021neuralsim}
    Eric Heiden, David Millard, Erwin Coumans, Yizhou Sheng, and Gaurav~S Sukhatme.
    \newblock Neuralsim: Augmenting differentiable simulators with neural networks.
    \newblock In {\em 2021 IEEE International Conference on Robotics and Automation (ICRA)}, pages 9474--9481. IEEE, 2021.
    
    \bibitem{hackenberg2021using}
    Maren Hackenberg, Marlon Grodd, Clemens Kreutz, Martina Fischer, Janina Esins, Linus Grabenhenrich, Christian Karagiannidis, and Harald Binder.
    \newblock Using differentiable programming for flexible statistical modeling.
    \newblock {\em The American Statistician}, pages 1--10, 2021.
    
    \bibitem{kochkov2021machine}
    Dmitrii Kochkov, Jamie~A Smith, Ayya Alieva, Qing Wang, Michael~P Brenner, and Stephan Hoyer.
    \newblock Machine learning--accelerated computational fluid dynamics.
    \newblock {\em Proceedings of the National Academy of Sciences}, 118(21), 2021.
    
    \bibitem{belbute2020combining}
    Filipe De~Avila Belbute-Peres, Thomas Economon, and Zico Kolter.
    \newblock Combining differentiable pde solvers and graph neural networks for fluid flow prediction.
    \newblock In {\em International Conference on Machine Learning}, pages 2402--2411. PMLR, 2020.
    
    \bibitem{um2020solver}
    Kiwon Um, Robert Brand, Yun~Raymond Fei, Philipp Holl, and Nils Thuerey.
    \newblock Solver-in-the-loop: Learning from differentiable physics to interact with iterative pde-solvers.
    \newblock {\em Advances in Neural Information Processing Systems}, 33:6111--6122, 2020.
    
    \bibitem{bar2019learning}
    Yohai Bar-Sinai, Stephan Hoyer, Jason Hickey, and Michael~P Brenner.
    \newblock Learning data-driven discretizations for partial differential equations.
    \newblock {\em Proceedings of the National Academy of Sciences}, 116(31):15344--15349, 2019.
    
    \bibitem{san2018neural}
    Omer San and Romit Maulik.
    \newblock Neural network closures for nonlinear model order reduction.
    \newblock {\em Advances in Computational Mathematics}, 44:1717--1750, 2018.
    
    \bibitem{beck2019deep}
    Andrea Beck, David Flad, and Claus-Dieter Munz.
    \newblock Deep neural networks for data-driven les closure models.
    \newblock {\em Journal of Computational Physics}, 398:108910, 2019.
    
    \bibitem{ronneberger2015u}
    Olaf Ronneberger, Philipp Fischer, and Thomas Brox.
    \newblock U-net: Convolutional networks for biomedical image segmentation.
    \newblock In {\em Medical Image Computing and Computer-Assisted Intervention--MICCAI 2015: 18th International Conference, Munich, Germany, October 5-9, 2015, Proceedings, Part III 18}, pages 234--241. Springer, 2015.
    
    \bibitem{dosovitskiy2020image}
    Alexey Dosovitskiy, Lucas Beyer, Alexander Kolesnikov, Dirk Weissenborn, Xiaohua Zhai, Thomas Unterthiner, Mostafa Dehghani, Matthias Minderer, Georg Heigold, Sylvain Gelly, et~al.
    \newblock An image is worth 16x16 words: Transformers for image recognition at scale.
    \newblock {\em arXiv preprint arXiv:2010.11929}, 2020.
    
    \bibitem{lu2022comprehensive}
    Lu~Lu, Xuhui Meng, Shengze Cai, Zhiping Mao, Somdatta Goswami, Zhongqiang Zhang, and George~Em Karniadakis.
    \newblock A comprehensive and fair comparison of two neural operators (with practical extensions) based on {FAIR} data.
    \newblock {\em Computer Methods in Applied Mechanics and Engineering}, 393:114778, 2022.
    
    \bibitem{li2021physics}
    Zongyi Li, Hongkai Zheng, Nikola Kovachki, David Jin, Haoxuan Chen, Burigede Liu, Kamyar Azizzadenesheli, and Anima Anandkumar.
    \newblock Physics-informed neural operator for learning partial differential equations.
    \newblock {\em arXiv preprint arXiv:2111.03794}, 2021.
    
    \bibitem{chen2018gradnorm}
    Zhao Chen, Vijay Badrinarayanan, Chen-Yu Lee, and Andrew Rabinovich.
    \newblock Gradnorm: Gradient normalization for adaptive loss balancing in deep multitask networks.
    \newblock In {\em International conference on machine learning}, pages 794--803. PMLR, 2018.
    
    \bibitem{mcclenny2020self}
    Levi McClenny and Ulisses Braga-Neto.
    \newblock Self-adaptive physics-informed neural networks using a soft attention mechanism.
    \newblock {\em arXiv preprint arXiv:2009.04544}, 2020.
    
    \bibitem{kucukelbir2017automatic}
    Alp Kucukelbir, Dustin Tran, Rajesh Ranganath, Andrew Gelman, and David~M Blei.
    \newblock Automatic differentiation variational inference.
    \newblock {\em Journal of machine learning research}, 2017.
    
    \bibitem{graves2011practical}
    Alex Graves.
    \newblock Practical variational inference for neural networks.
    \newblock {\em Advances in neural information processing systems}, 24, 2011.
    
    \bibitem{hoffman2013stochastic}
    Matthew~D Hoffman, David~M Blei, Chong Wang, and John Paisley.
    \newblock Stochastic variational inference.
    \newblock {\em Journal of Machine Learning Research}, 2013.
    
    \bibitem{gal2016dropout}
    Yarin Gal and Zoubin Ghahramani.
    \newblock Dropout as a bayesian approximation: Representing model uncertainty in deep learning.
    \newblock In {\em international conference on machine learning}, pages 1050--1059. PMLR, 2016.
    
    \bibitem{lakshminarayanan2017simple}
    Balaji Lakshminarayanan, Alexander Pritzel, and Charles Blundell.
    \newblock Simple and scalable predictive uncertainty estimation using deep ensembles.
    \newblock {\em Advances in neural information processing systems}, 30, 2017.
    
    \bibitem{ovadia2019can}
    Yaniv Ovadia, Emily Fertig, Jie Ren, Zachary Nado, David Sculley, Sebastian Nowozin, Joshua Dillon, Balaji Lakshminarayanan, and Jasper Snoek.
    \newblock Can you trust your model's uncertainty? evaluating predictive uncertainty under dataset shift.
    \newblock {\em Advances in neural information processing systems}, 32, 2019.
    
    \bibitem{rahaman2021uncertainty}
    Rahul Rahaman et~al.
    \newblock Uncertainty quantification and deep ensembles.
    \newblock {\em Advances in Neural Information Processing Systems}, 34:20063--20075, 2021.
    
    \bibitem{dong2017image}
    Bin Dong, Qingtang Jiang, and Zuowei Shen.
    \newblock Image restoration: Wavelet frame shrinkage, nonlinear evolution pdes, and beyond.
    \newblock {\em Multiscale Modeling \& Simulation}, 15(1):606--660, 2017.
    
    \bibitem{long2018pde}
    Zichao Long, Yiping Lu, Xianzhong Ma, and Bin Dong.
    \newblock Pde-net: Learning pdes from data.
    \newblock In {\em International Conference on Machine Learning}, pages 3208--3216. PMLR, 2018.
    
    \bibitem{long2019pde}
    Zichao Long, Yiping Lu, and Bin Dong.
    \newblock Pde-net 2.0: Learning pdes from data with a numeric-symbolic hybrid deep network.
    \newblock {\em Journal of Computational Physics}, 399:108925, 2019.
    
    \bibitem{sanchez2020learning}
    Alvaro Sanchez-Gonzalez, Jonathan Godwin, Tobias Pfaff, Rex Ying, Jure Leskovec, and Peter Battaglia.
    \newblock Learning to simulate complex physics with graph networks.
    \newblock In {\em International Conference on Machine Learning}, pages 8459--8468. PMLR, 2020.
    
\end{thebibliography}

\clearpage
\begin{thebibliography}{1}

\bibitem{NEURIPS2019_9015}
Adam Paszke, Sam Gross, Francisco Massa, Adam Lerer, James Bradbury, Gregory Chanan, Trevor Killeen, Zeming Lin, Natalia Gimelshein, Luca Antiga, Alban Desmaison, Andreas Kopf, Edward Yang, Zachary DeVito, Martin Raison, Alykhan Tejani, Sasank Chilamkurthy, Benoit Steiner, Lu~Fang, Junjie Bai, and Soumith Chintala.
\newblock Pytorch: An imperative style, high-performance deep learning library.
\newblock In H.~Wallach, H.~Larochelle, A.~Beygelzimer, F.~d\textquotesingle Alch\'{e}-Buc, E.~Fox, and R.~Garnett, editors, {\em Advances in Neural Information Processing Systems 32}, pages 8024--8035. Curran Associates, Inc., 2019.

\bibitem{wang2022improved}
Sifan Wang, Hanwen Wang, and Paris Perdikaris.
\newblock Improved architectures and training algorithms for deep operator networks.
\newblock {\em Journal of Scientific Computing}, 92(2):1--42, 2022.

\bibitem{lu2022comprehensive}
Lu~Lu, Xuhui Meng, Shengze Cai, Zhiping Mao, Somdatta Goswami, Zhongqiang Zhang, and George~Em Karniadakis.
\newblock A comprehensive and fair comparison of two neural operators (with practical extensions) based on {FAIR} data.
\newblock {\em Computer Methods in Applied Mechanics and Engineering}, 393:114778, 2022.

\bibitem{lu2021learning}
Lu~Lu, Pengzhan Jin, Guofei Pang, Zhongqiang Zhang, and George~Em Karniadakis.
\newblock Learning nonlinear operators via deeponet based on the universal approximation theorem of operators.
\newblock {\em Nature Machine Intelligence}, 3(3):218--229, 2021.

\bibitem{li2020fourier}
Zongyi Li, Nikola~Borislavov Kovachki, Kamyar Azizzadenesheli, Kaushik Bhattacharya, Andrew Stuart, Anima Anandkumar, et~al.
\newblock Fourier neural operator for parametric partial differential equations.
\newblock In {\em International Conference on Learning Representations}, 2020.

\bibitem{hendrycks2016gaussian}
Dan Hendrycks and Kevin Gimpel.
\newblock Gaussian error linear units (gelus).
\newblock {\em arXiv preprint arXiv:1606.08415}, 2016.

\bibitem{wang2021learning}
Sifan Wang, Hanwen Wang, and Paris Perdikaris.
\newblock Learning the solution operator of parametric partial differential equations with physics-informed deeponets.
\newblock {\em Science advances}, 7(40):eabi8605, 2021.

\bibitem{li2021physics}
Zongyi Li, Hongkai Zheng, Nikola Kovachki, David Jin, Haoxuan Chen, Burigede Liu, Kamyar Azizzadenesheli, and Anima Anandkumar.
\newblock Physics-informed neural operator for learning partial differential equations.
\newblock {\em arXiv preprint arXiv:2111.03794}, 2021.

\end{thebibliography}
\end{document}